\documentclass{report}

\usepackage{graphicx} 
\usepackage[a4paper, total={7.5in, 10.5in}]{geometry}
\usepackage{amsmath}
\usepackage{cite} 
\usepackage{tabularx}
\usepackage{float}
\usepackage{enumitem}
\usepackage{url}
\usepackage{booktabs}
\usepackage{hyperref} 
\usepackage{xcolor}
\usepackage{adjustbox}
\usepackage{lscape}
\usepackage{array}
\usepackage[numbers]{natbib} 
\usepackage{authblk}

\title{\vspace{-3cm}  
\includegraphics[width=0.4\textwidth]{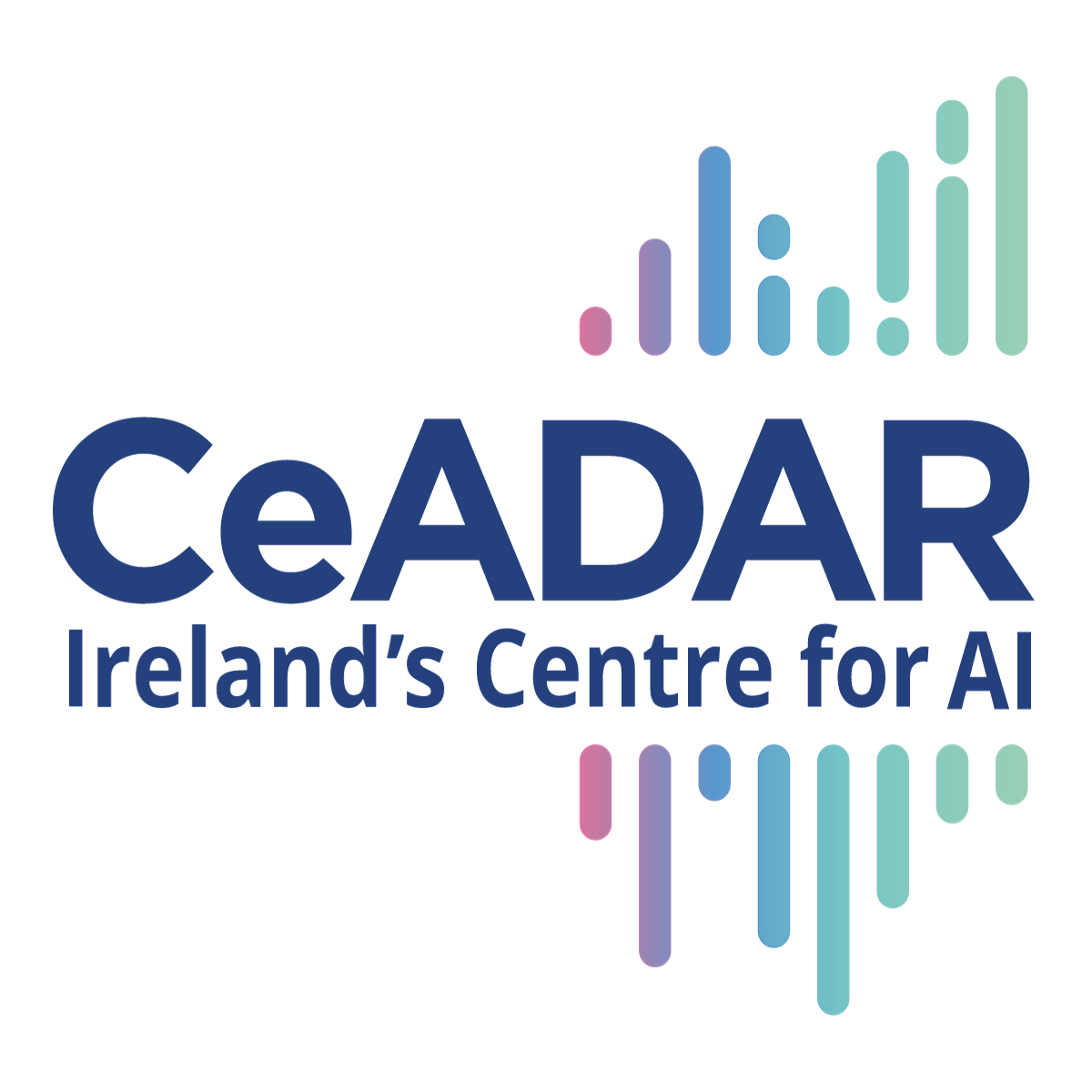}\\ 
    \vspace{1em}\textbf{The Ultimate Guide to Fine-Tuning LLMs from Basics to Breakthroughs: An Exhaustive Review of Technologies, Research, Best Practices, Applied Research Challenges and Opportunities} \\ \vspace{0.5em} \large{(Version 1.1)} \vspace{3.5em}}

\author{\Large{Venkatesh Balavadhani Parthasarathy, Ahtsham Zafar, Aafaq Khan, and Arsalan Shahid} \vspace{1.5em}  \\ \large{@ CeADAR Connect Group} \vspace{1.5em}}

\affil{\large{CeADAR: Ireland's Centre for AI, University College Dublin, Belfield, Dublin, Ireland} \\ \vspace{1.5em} \{\large{ {venkatesh.parthasarathy, ahtsham.zafar, aafaq.khan, arsalan.shahid \} @ ucd.ie}}}

\date{\vspace{11.5em} \textbf{October 2024}}

\begin{document}

\maketitle

\begin{abstract}
This technical report thoroughly examines the process of fine-tuning Large Language Models (LLMs), integrating theoretical insights and practical applications. It begins by tracing the historical development of LLMs, emphasising their evolution from traditional Natural Language Processing (NLP) models and their pivotal role in modern AI systems. The analysis differentiates between various fine-tuning methodologies, including supervised, unsupervised, and instruction-based approaches, underscoring their respective implications for specific tasks.

A structured seven-stage pipeline for LLM fine-tuning is introduced, covering the complete lifecycle from data preparation to model deployment. Key considerations include data collection strategies, handling of imbalanced datasets, model initialisation, and optimisation techniques, with a particular focus on hyperparameter tuning. The report also highlights parameter-efficient fine-tuning methods such as Low-Rank Adaptation (LoRA) and Half Fine-Tuning, which balance resource constraints with optimal model performance.

The exploration extends to advanced fine-tuning techniques and configurations like memory fine-tuning, Mixture of Experts (MoE) and Mixture of Agents (MoA), demonstrating how these methods harness specialised networks and multi-agent collaboration for improved outcomes. Proximal Policy Optimisation (PPO) and Direct Preference Optimisation (DPO) are discussed as innovative approaches to aligning models with human preferences, while the benefits of pruning and routing optimisations are examined for enhancing efficiency.

In the latter sections, the report delves into validation frameworks, post-deployment monitoring, and optimisation techniques for inference. It also addresses the deployment of LLMs on distributed and cloud-based platforms. Additionally, cutting-edge topics such as multimodal LLMs and fine-tuning for audio and speech processing are covered, alongside emerging challenges related to scalability, privacy, and accountability.

This report aims to serve as a comprehensive guide for researchers and practitioners, offering actionable insights into fine-tuning LLMs while navigating the challenges and opportunities inherent in this rapidly evolving field.
\end{abstract}

\tableofcontents

\pagebreak

\chapter{Introduction}

\section{Background of Large Language Models (LLMs)}

Large Language Models (LLMs) represent a significant leap in computational systems capable of understanding and generating human language. Building on traditional language models (LMs) like N-gram models \cite{ngramLMStanford}, LLMs address limitations such as rare word handling, overfitting, and capturing complex linguistic patterns. Notable examples, such as GPT-3 and GPT-4 \cite{gptSeriesComparison}, leverage the self-attention mechanism within Transformer architectures to efficiently manage sequential data and understand long-range dependencies. Key advancements include in-context learning for generating coherent text from prompts and Reinforcement Learning from Human Feedback (RLHF) \cite{surveyRLHF} for refining models using human responses. Techniques like prompt engineering, question-answering, and conversational interactions have significantly advanced the field of natural language processing (NLP) \cite{surveyOfLLMs}.

\section{Historical Development and Key Milestones}

Language models are fundamental to natural language processing (NLP), leveraging mathematical techniques to generalise linguistic rules and knowledge for tasks involving prediction and generation. Over several decades, language modelling has evolved from early statistical language models (SLMs) to today's advanced large language models (LLMs). This rapid advancement has enabled LLMs to process, comprehend, and generate text at a level comparable to human capabilities \cite{zafar2024building, historyOfLLMsSurvey}.

\begin{figure}[h]
    \centering
    \includegraphics[width=\textwidth]{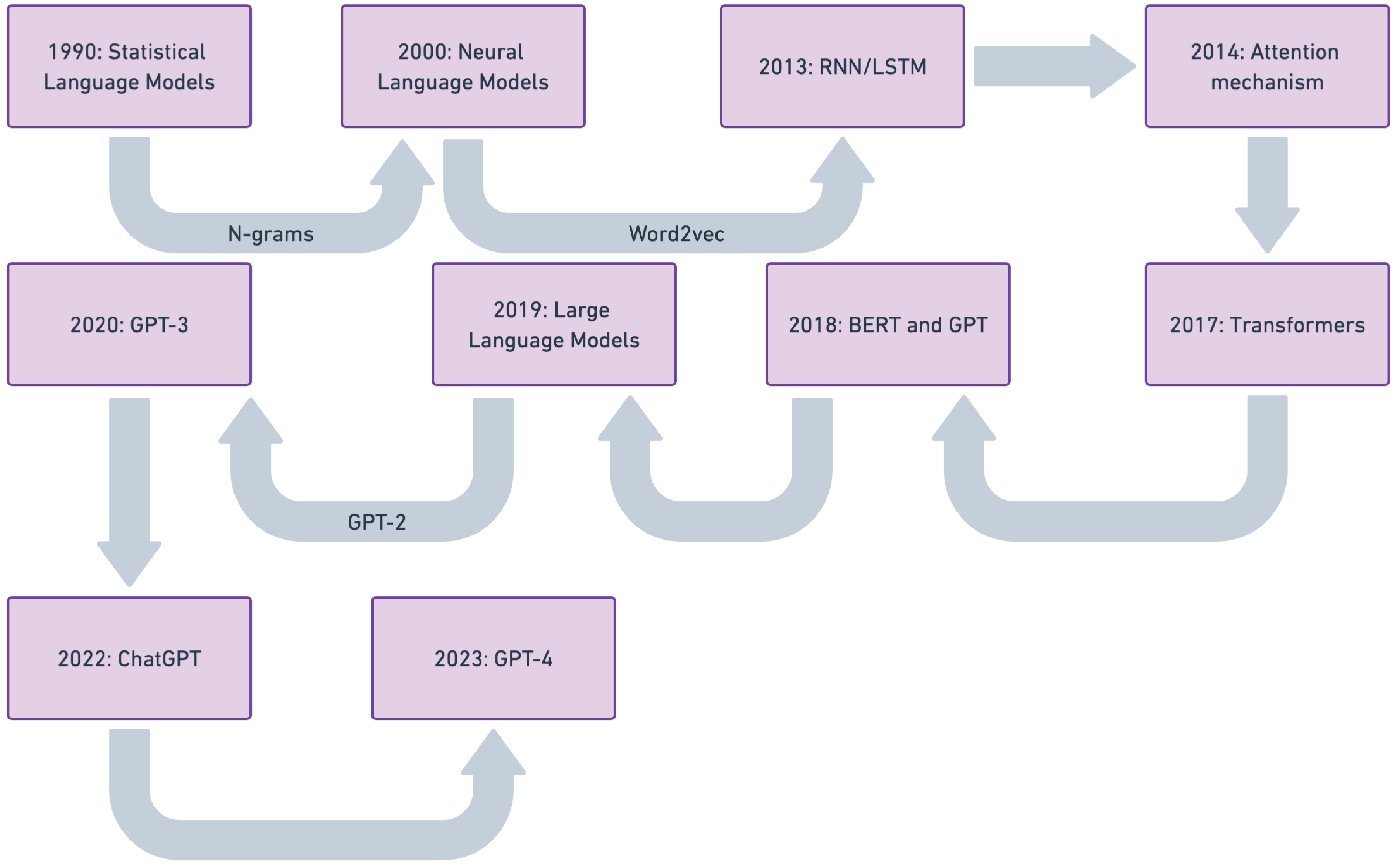} 
    \caption{A chronological timeline showcasing the evolution of Large Language Models (LLMs) from 1990 to 2023. This progression begins with early statistical models such as N-grams, transitions through neural language models like Word2Vec and RNN/LSTM, and advances into the era of pre-trained models with the introduction of transformers and attention mechanisms. The figure highlights significant milestones, including the development of BERT, GPT series, and recent innovations such as GPT-4 and ChatGPT, demonstrating the rapid advancements in LLM technology over time. (adapted from \cite{historyOfLLMsSurvey})}
    \label{evolutionOfLLMs}
\end{figure}

\noindent Figure \ref{evolutionOfLLMs} shows the evolution of large language models from early statistical approaches to current advanced models.

\section{Evolution from Traditional NLP Models to State-of-the-Art LLMs}

Understanding LLMs requires tracing the development of language models through stages such as Statistical Language Models (SLMs), Neural Language Models (NLMs), Pre-trained Language Models (PLMs), and LLMs.

\subsection{Statistical Language Models (SLMs)}

Emerging in the 1990s, SLMs analyse natural language using probabilistic methods to determine the likelihood of sentences within texts. For instance, the probability \( P(S) \) of the sentence “I am very happy” is given by:

\begin{equation}
P(S) = P(\omega_1, \omega_2, \omega_3, \omega_4) = P(\text{I}, \text{am}, \text{very}, \text{happy})
\end{equation}

This probability can be calculated using conditional probabilities:

\begin{equation}
P(\text{I}, \text{am}, \text{very}, \text{happy}) = P(\text{I}) \cdot P(\text{am} \mid \text{I}) \cdot P(\text{very} \mid \text{I}, \text{am}) \cdot P(\text{happy} \mid \text{I}, \text{am}, \text{very})
\end{equation}

Conditional probabilities are estimated using Maximum Likelihood Estimation (MLE):

\begin{equation}
P(\omega_i \mid \omega_1 \omega_2 \cdots \omega_{i-1}) = \frac{C(\omega_1 \omega_2 \cdots \omega_i)}{C(\omega_1 \omega_2 \cdots \omega_{i-1})}
\end{equation}

\subsection{Neural Language Models (NLMs)}

NLMs leverage neural networks to predict word sequences, overcoming SLM limitations. Word vectors enable computers to understand word meanings. Tools like Word2Vec \cite{word2vecPaper} represent words in a vector space where semantic relationships are reflected in vector angles. NLMs consist of interconnected neurons organised into layers, resembling the human brain's structure.
The input layer concatenates word vectors, the hidden layer applies a non-linear activation function, and the output layer predicts subsequent words using the Softmax function to transform values into a probability distribution.

\begin{figure}[h]
    \centering
    \includegraphics[width=0.75\textwidth]{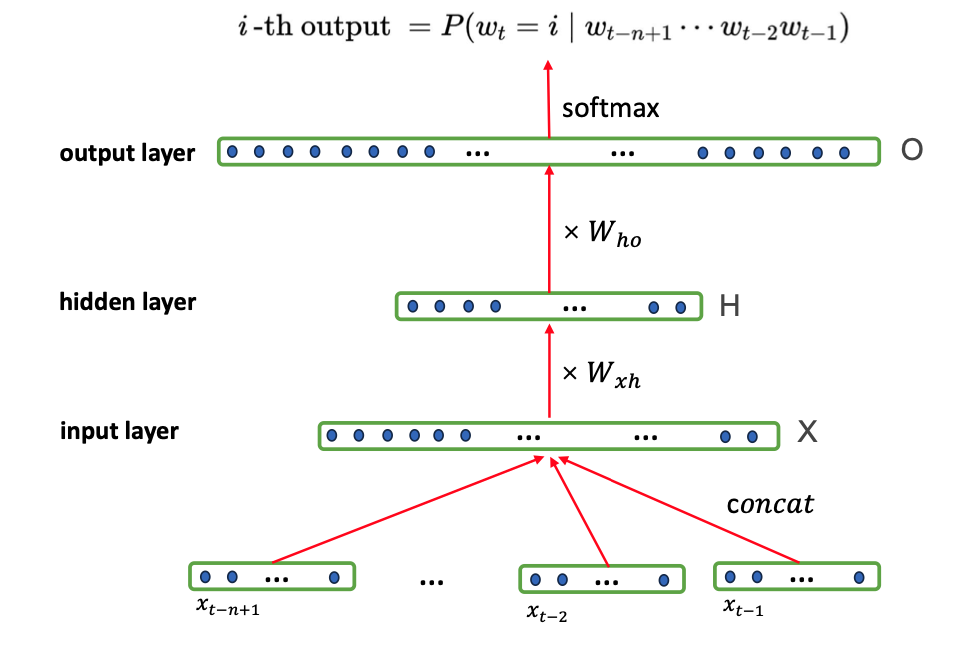} 
    \caption{A schematic representation of Neural Language Models, showcasing the layered architecture where the input layer processes sequential data, the hidden layer captures dependencies, and the output layer generates predictions. The figure emphasises the flow of information through concatenation and matrix multiplications, culminating in a probability distribution via the softmax function. (adopted from \cite{historyOfLLMsSurvey})}
    \label{nlms}
\end{figure}

\noindent Figure \ref{nlms} illustrates the structure of Neural Language Models, highlighting the layers and connections used to predict subsequent words.

\subsection{Pre-trained Language Models (PLMs)}

PLMs are initially trained on extensive volumes of unlabelled text to understand fundamental language structures (pre-training). They are then fine-tuned on a smaller, task-specific dataset. This "pre-training and fine-tuning" paradigm, exemplified by GPT-2 \cite{gpt2ModelPaper} and BERT \cite{bertLanguageModel}, has led to diverse and effective model architectures.

\subsection{Large Language Models (LLMs)}

LLMs like GPT-3, GPT-4, PaLM \cite{palmPaper}, and LLaMA \cite{llamaPaper} are trained on massive text corpora with tens of billions of parameters. LLMs undergo a two-stage process: initial pre-training on a vast corpus followed by alignment with human values. This approach enables LLMs to understand human commands and values better.

\section{Overview of Current Leading LLMs}

LLMs are powerful tools in NLP, capable of performing tasks such as translation, summarisation, and conversational interaction. Advances in transformer architectures, computational power, and extensive datasets have driven their success. These models approximate human-level performance, making them invaluable for research and practical implementations.
LLMs' rapid development has spurred research into architectural innovations, training strategies, extending context lengths, fine-tuning techniques, and integrating multi-modal data. Their applications extend beyond NLP, aiding in human-robot interactions and creating intuitive AI systems. This highlights the importance of comprehensive reviews consolidating the latest developments \cite{artOfFineTuningLLMs}.

\begin{figure}[h]
    \centering
    \includegraphics[width=\textwidth]{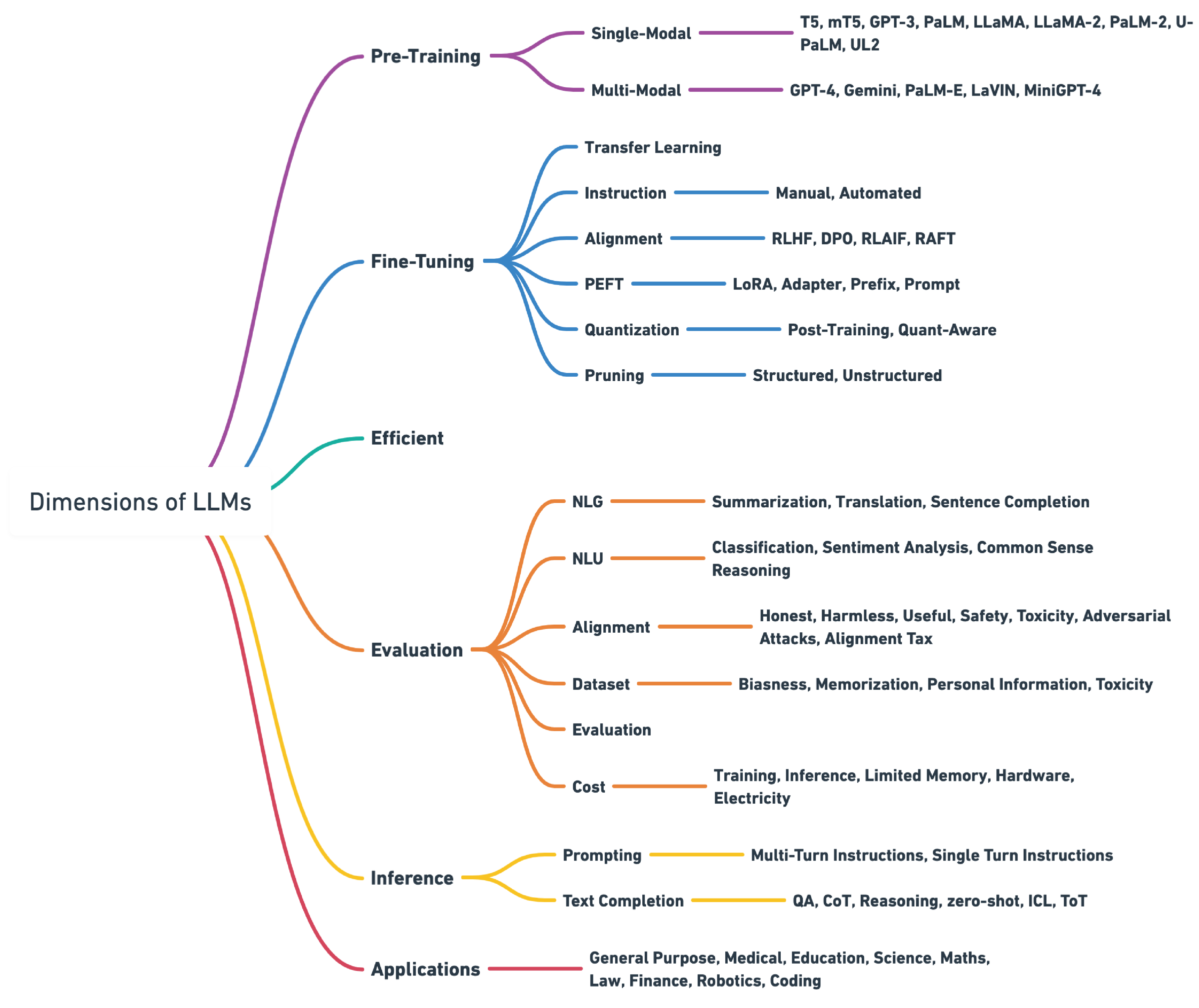} 
    \caption{Mind map depicting various dimensions of Large Language Models (LLMs), covering aspects from pre-training and fine-tuning methodologies to efficiency, evaluation, inference, and application domains. Each dimension is linked to specific techniques, challenges, and examples of models that exemplify the discussed characteristics. This diagram serves as an overview of the multifaceted considerations in the development and deployment of LLMs. (adapted from \cite{overviewOfLLMs})}
    \label{overViewOfLLMs}
\end{figure}

\noindent Figure \ref{overViewOfLLMs} provides an overview of current leading LLMs, highlighting their capabilities and applications.

\section{What is Fine-Tuning?}

Fine-tuning uses a pre-trained model, such as OpenAI's GPT series, as a foundation. The process involves further training on a smaller, domain-specific dataset. This approach builds upon the model's pre-existing knowledge, enhancing performance on specific tasks with reduced data and computational requirements.

\noindent Fine-tuning transfers the pre-trained model's learned patterns and features to new tasks, improving performance and reducing training data needs. It has become popular in NLP for tasks like text classification, sentiment analysis, and question-answering.

\section{Types of LLM Fine-Tuning}

\subsection{Unsupervised Fine-Tuning}

This method does not require labelled data. Instead, the LLM is exposed to a large corpus of unlabelled text from the target domain, refining its understanding of language. This approach is useful for new domains like legal or medical fields but is less precise for specific tasks such as classification or summarisation.

\subsection{Supervised Fine-Tuning (SFT)}

SFT involves providing the LLM with labelled data tailored to the target task. For example, fine-tuning an LLM for text classification in a business context uses a dataset of text snippets with class labels. While effective, this method requires substantial labelled data, which can be costly and time-consuming to obtain.

\subsection{Instruction Fine-Tuning via Prompt Engineering}

This method relies on providing the LLM with natural language instructions, useful for creating specialised assistants. It reduces the need for vast amounts of labelled data but depends heavily on the quality of the prompts.

\section{Pre-training vs Fine-tuning}

Table \ref{tab:pretraining_vs_finetuning} provides a comparison between pre-training and fine-tuning, highlighting their respective characteristics and processes.

\begin{table}[!ht]
\centering
\begin{tabularx}{\textwidth}{|X|X|X|}
\hline
\textbf{Aspect} & \textbf{Pre-training} & \textbf{Fine-tuning} \\ \hline
Definition & Training on a vast amount of unlabelled text data & Adapting a pre-trained model to specific tasks \\ \hline
Data Requirement & Extensive and diverse unlabelled text data & Smaller, task-specific labelled data \\ \hline
Objective & Build general linguistic knowledge & Specialise model for specific tasks \\ \hline
Process & Data collection, training on large dataset, predict next word/sequence & Task-specific data collection, modify last layer for task, train on new dataset, generate output based on tasks \\ \hline
Model Modification & Entire model trained & Last layers adapted for new task \\ \hline
Computational Cost & High (large dataset, complex model) & Lower (smaller dataset, fine-tuning layers) \\ \hline
Training Duration & Weeks to months & Days to weeks \\ \hline
Purpose & General language understanding & Task-specific performance improvement \\ \hline
Examples & GPT, LLaMA 3 & Fine-tuning LLaMA 3 for summarisation \\ \hline
\end{tabularx}
\caption{A Comparative Overview of Pre-training and Fine-tuning in Large Language Models (LLMs). The table outlines key differences between the pre-training and fine-tuning phases across various aspects such as definition, data requirements, objectives, processes, model modification, computational costs, training duration, and their respective purposes, with examples highlighting specific models and tasks. Pre-training involves extensive training on vast amounts of unlabelled data to build general linguistic knowledge, while fine-tuning adapts the pre-trained models to specialised tasks using smaller, labelled datasets, focusing on task-specific performance improvements.}
\label{tab:pretraining_vs_finetuning}
\end{table}

\section{Importance of Fine-Tuning LLMs}

\begin{enumerate}
    \item \textbf{Transfer Learning:} Fine-tuning leverages the knowledge acquired during pre-training, adapting it to specific tasks with reduced computation time and resources.
    \item \textbf{Reduced Data Requirements:} Fine-tuning requires less labelled data, focusing on tailoring pre-trained features to the target task.
    \item \textbf{Improved Generalisation:} Fine-tuning enhances the model's ability to generalise to specific tasks or domains, capturing general language features and customising them.
    \item \textbf{Efficient Model Deployment:} Fine-tuned models are more efficient for real-world applications, being computationally efficient and well-suited for specific tasks.
    \item \textbf{Adaptability to Various Tasks:} Fine-tuned LLMs can adapt to a broad range of tasks, performing well across various applications without task-specific architectures.
    \item \textbf{Domain-Specific Performance:} Fine-tuning allows models to excel in domain-specific tasks by adjusting to the nuances and vocabulary of the target domain.
    \item \textbf{Faster Convergence:} Fine-tuning usually achieves faster convergence, starting with weights that already capture general language features.
\end{enumerate}

\section{Retrieval Augmented Generation (RAG)}

\begin{figure}[h]
\centering
\includegraphics[width=0.6\textwidth]{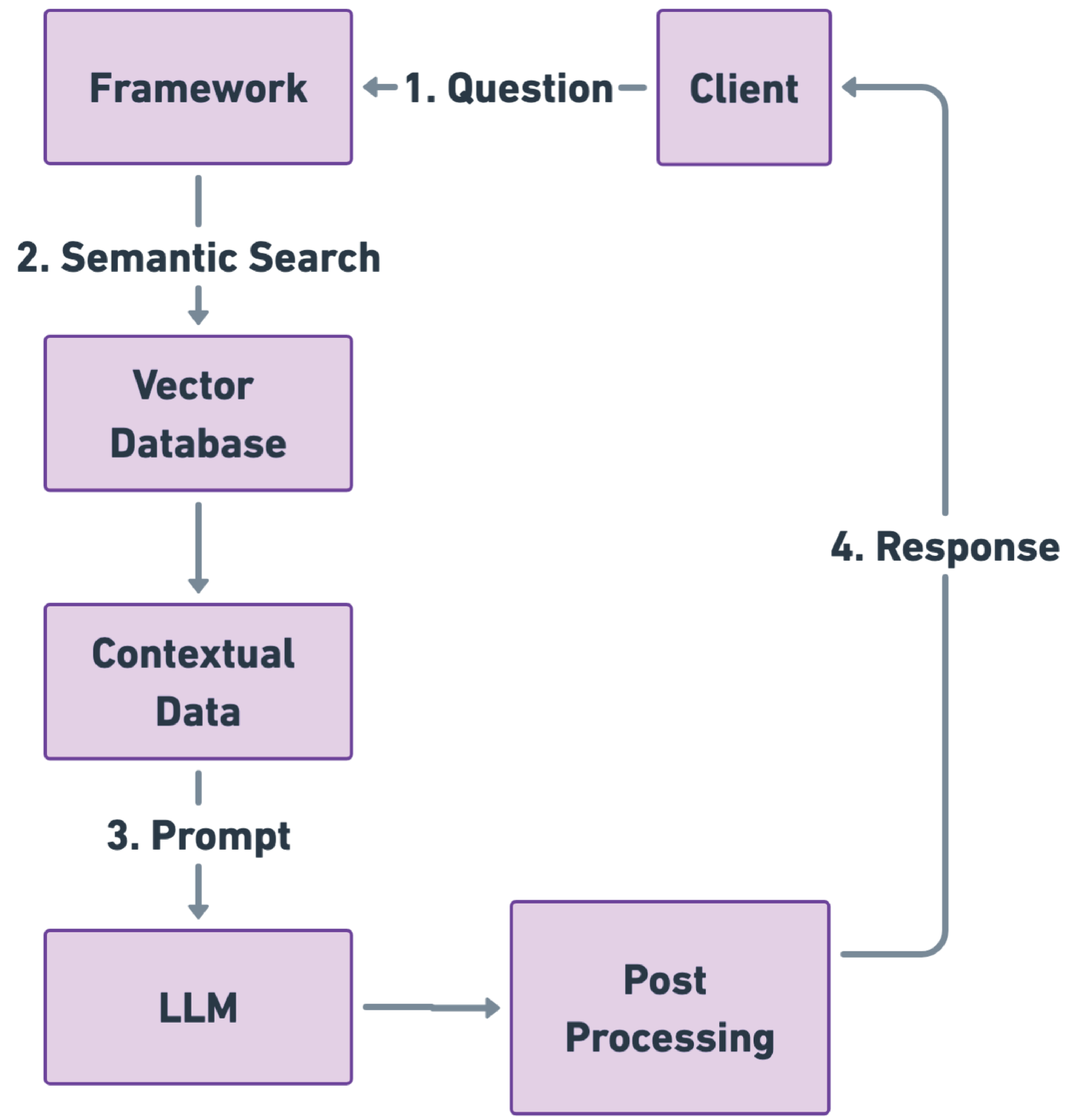}
\caption{An illustration of the Traditional Retrieval-Augmented Generation (RAG) pipeline steps, depicting the sequential process from client query to response generation. The pipeline starts with the client’s question, followed by semantic search in a vector database, contextually enriching the data before generating a prompt for the large language model (LLM). The final response is post-processed and returned to the client.}
\label{ragPipeline}
\end{figure}

A popular method to utilise your own data is by incorporating it into the prompt when querying the LLM model. This approach, known as Retrieval-Augmented Generation (RAG), involves retrieving relevant data and using it as additional context for the LLM. Instead of depending solely on knowledge from the training data, a RAG workflow pulls pertinent information, connecting static LLMs with real-time data retrieval. With RAG architecture, organisations can deploy any LLM model and enhance it to return relevant results by providing a small amount of their own data (see Figure\ref{ragPipeline} for visual workflow). This process avoids the costs and time associated with fine-tuning or pre-training the model.

\subsection{Traditional RAG Pipeline and Steps}

\begin{enumerate}
    \item \textbf{Data Indexing:} Organise data efficiently for quick retrieval. This involves processing, chunking, and storing data in a vector database using indexing strategies like search indexing, vector indexing, and hybrid indexing.
    \item \textbf{Input Query Processing:} Refine user queries to improve compatibility with indexed data. This can include simplification or vector transformation of queries for enhanced search efficiency.
    \item \textbf{Searching and Ranking:} Retrieve and rank data based on relevance using search algorithms such as TF-IDF, BM25, and deep learning models like BERT to interpret the query’s intent and context.
    \item \textbf{Prompt Augmentation:} Incorporate relevant information from the search results into the original query to provide the LLM with additional context, enhancing response accuracy and relevance.
    \item \textbf{Response Generation:} Use the augmented prompt to generate responses that combine the LLM’s knowledge with current, specific data, ensuring high-quality, contextually grounded answers.
\end{enumerate}

\subsection{Benefits of Using RAG}

\begin{itemize}
    \item \textbf{Up-to-Date and Accurate Responses:} Enhances the LLM’s responses with current external data, improving accuracy and relevance.
    \item \textbf{Reducing Inaccurate Responses:} Grounds the LLM’s output in relevant knowledge, reducing the risk of generating incorrect information.
    \item \textbf{Domain-Specific Responses:} Delivers contextually relevant responses tailored to an organisation's proprietary data.
    \item \textbf{Efficiency and Cost-Effectiveness:} Offers a cost-effective method for customising LLMs without extensive model fine-tuning.
\end{itemize}

\subsection{Challenges and Considerations in Serving RAG}

\begin{enumerate}
    \item \textbf{User Experience:} Ensuring rapid response times suitable for real-time applications.
    \item \textbf{Cost Efficiency:} Managing the costs associated with serving millions of responses.
    \item \textbf{Accuracy:} Ensuring outputs are accurate to avoid misinformation.
    \item \textbf{Recency and Relevance:} Keeping responses and content current with the latest data.
    \item \textbf{Business Context Awareness:} Aligning LLM responses with specific business contexts.
    \item \textbf{Service Scalability:} Managing increased capacity while controlling costs.
    \item \textbf{Security and Governance:} Implementing protocols for data security, privacy, and governance.
\end{enumerate}

\subsection{Use Cases and Examples}

\begin{enumerate}
    \item \textbf{Question and Answer Chatbots:} Integrate LLMs with chatbots to generate accurate answers from company documents, enhancing customer support.
    \item \textbf{Search Augmentation:} Enhance search engines with LLM-generated answers for more accurate informational queries.
    \item \textbf{Knowledge Engine:} Use LLMs to answer questions related to internal functions, such as HR and compliance, using company data.
\end{enumerate}

\subsection{Considerations for Choosing Between RAG and Fine-Tuning}
\begin{figure}[h]
\centering
\includegraphics[width=0.7\textwidth]{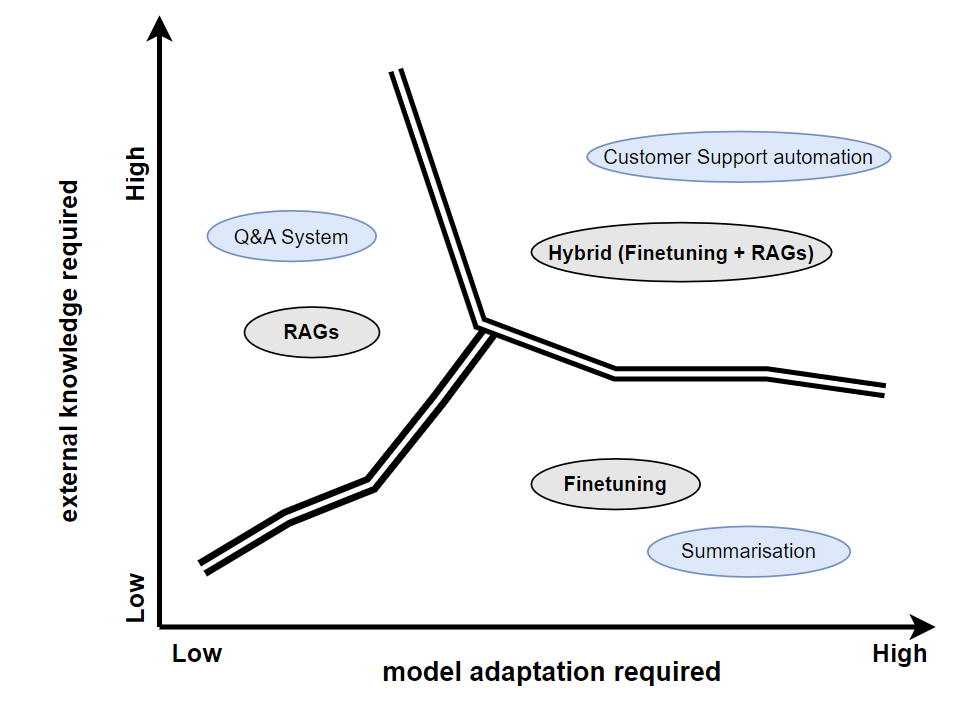}
\caption{Graph comparing the model adaptation required versus the level of external knowledge needed across different scenarios, highlighting the roles of Retrieval-Augmented Generation (RAG), Fine-Tuning, and their hybrid applications in various contexts such as Q\&A systems, customer support automation, and summarisation tasks. (adapted from \cite{ragVsFtGraph})}
\label{ragFineTuningGraph}
\end{figure}

When considering external data access, RAG is likely a superior option for applications needing to access external data sources. Fine-tuning, on the other hand, is more suitable if you require the model to adjust its behaviour, and writing style, or incorporate domain-specific knowledge. In terms of suppressing hallucinations and ensuring accuracy, RAG systems tend to perform better as they are less prone to generating incorrect information. If you have ample domain-specific, labelled training data, fine-tuning can result in a more tailored model behaviour, whereas RAG systems are robust alternatives when such data is scarce. RAG systems provide an advantage with dynamic data retrieval capabilities for environments where data frequently updates or changes. Additionally, it is crucial to ensure the transparency and interpret ability of the model’s decision-making process. In that case, RAG systems offer insight that is typically not available in models that are solely fine-tuned. Figure\ref{ragFineTuningGraph} illustrates the visual representation alongside example use cases. 

\section{Objectives of the Report}

\subsection{Goals and Scope}

The primary goal of this report is to conduct a comprehensive analysis of fine-tuning techniques for LLMs. This involves exploring theoretical foundations, practical implementation strategies, and challenges. The report examines various fine-tuning methodologies, their applications, and recent advancements.

\subsection{Key Questions and Issues Addressed}

This report addresses critical questions surrounding fine-tuning LLMs, starting with foundational insights into LLMs, their evolution, and significance in NLP. It defines fine-tuning, distinguishes it from pre-training, and emphasises its role in adapting models for specific tasks. Key objectives include enhancing model performance for targeted applications and domains. \\

\noindent The report outlines a structured fine-tuning process, featuring a high-level pipeline with visual representations and detailed stage explanations. It covers practical implementation strategies, including model initialisation, hyperparameter definition, and fine-tuning techniques such as Parameter-Efficient Fine-Tuning (PEFT) and Retrieval-Augmented Generation (RAG). Industry applications, evaluation methods, deployment challenges, and recent advancements are also explored.

\subsection{Overview of the Report Structure}

The rest of the report provides a comprehensive understanding of fine-tuning LLMs. The main chapters include an in-depth look at the fine-tuning pipeline, practical applications, model alignment, evaluation metrics, and challenges. The concluding sections discuss the evolution of fine-tuning techniques, highlight ongoing research challenges, and provide insights for researchers and practitioners.

\chapter{Seven Stage Fine-Tuning Pipeline for LLM}

Fine-tuning a Large Language Model (LLM) is a comprehensive process divided into seven distinct stages, each essential for adapting the pre-trained model to specific tasks and ensuring optimal performance. These stages encompass everything from initial dataset preparation to the final deployment and maintenance of the fine-tuned model. By following these stages systematically, the model is refined and tailored to meet precise requirements, ultimately enhancing its ability to generate accurate and contextually appropriate responses. The seven stages include Dataset Preparation, Model Initialisation, Training Environment Setup, Fine-Tuning, Evaluation and Validation, Deployment, and Monitoring and Maintenance.

\begin{figure}[h]
\centering
\includegraphics[width=1\textwidth]{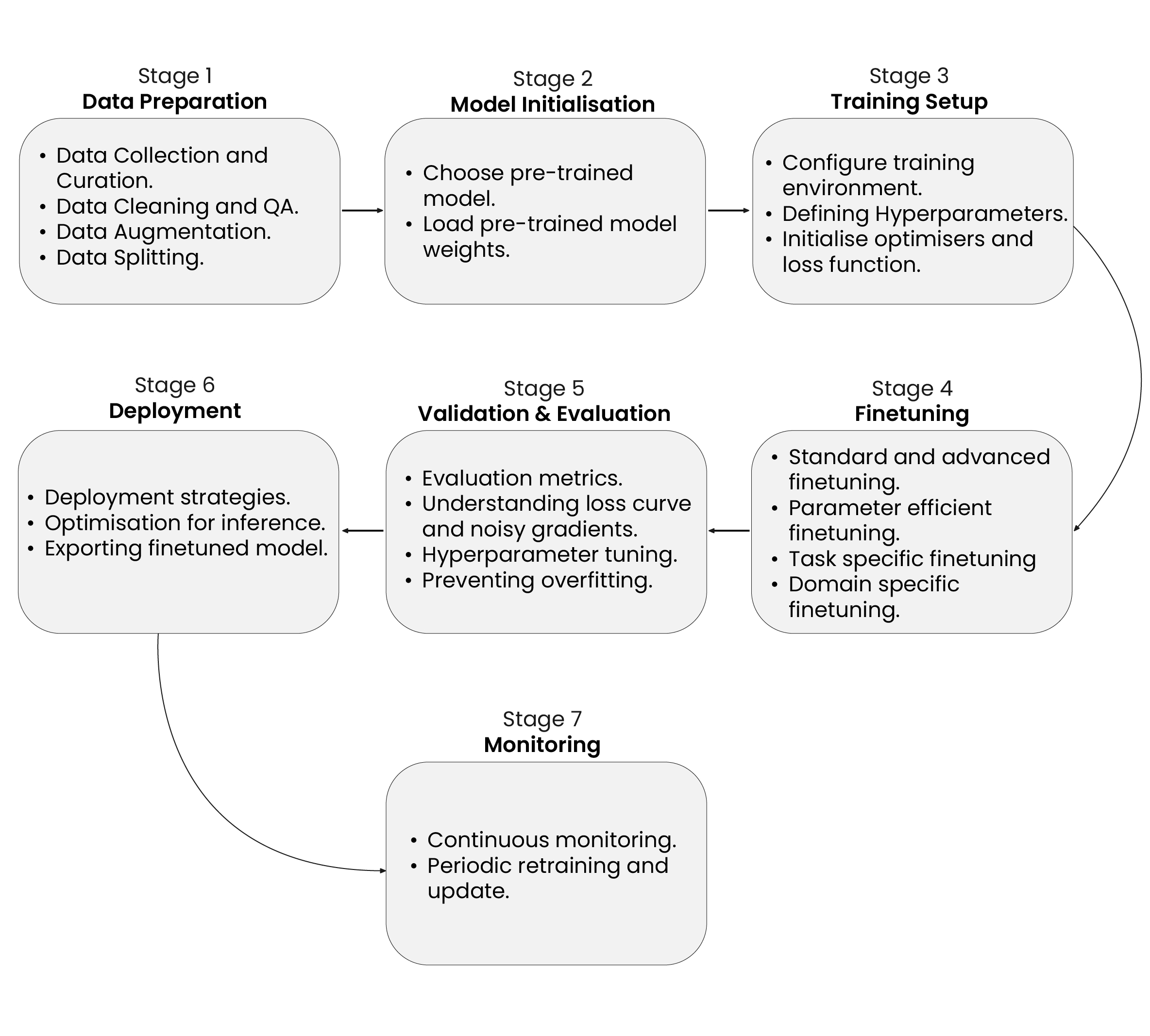}
\caption{A comprehensive pipeline for fine-tuning Large Language Models (LLMs), illustrating the seven essential stages: Dataset Preparation, Model Initialisation, Training Environment Setup, Fine-Tuning, Evaluation and Validation, Deployment, and Monitoring and Maintenance. Each stage plays a crucial role in adapting the pre-trained model to specific tasks and ensuring optimal performance throughout its lifecycle.}
\label{fineTuningLLMs}
\end{figure}

\noindent Figure \ref{fineTuningLLMs} illustrates the comprehensive pipeline for fine-tuning LLMs, encompassing all necessary stages from dataset preparation to monitoring and maintenance.

\section{Stage 1: Dataset Preparation}

Fine-tuning a Large Language Model (LLM) starts with adapting the pre-trained model for specific tasks by updating its parameters using a new dataset. This involves cleaning and formatting the dataset to match the target task, such as instruction tuning, sentiment analysis, or topic mapping. The dataset is composed of \(<\text{input}, \text{output}>\) pairs, demonstrating the desired behaviour for the model.

\noindent For example, in instruction tuning, the dataset may look like:

\begin{verbatim}
###Human: $<Input Query>$
###Assistant: $<Generated Output>$
\end{verbatim}

\noindent Here, the 'Input Query' is what the user asks, and the 'Generated Output' is the model's response. The structure and style of these pairs can be adjusted based on the specific needs of the task.

\section{Stage 2: Model Initialisation}

Model initialisation is the process of setting up the initial parameters and configurations of the LLM before training or deploying it. This step is crucial for ensuring the model performs optimally, trains efficiently, and avoids issues such as vanishing or exploding gradients.

\section{Stage 3: Training Environment Setup}

Setting up the training environment for LLM fine-tuning involves configuring the necessary infrastructure to adapt a pre-existing model for specific tasks. This includes selecting relevant training data, defining the model's architecture and hyperparameters, and running training iterations to adjust the model's weights and biases. The aim is to enhance the LLM's performance in generating accurate and contextually appropriate outputs tailored to specific applications, like content creation, translation, or sentiment analysis. Successful fine-tuning relies on careful preparation and rigorous experimentation.

\section{Stage 4: Partial or Full Fine-Tuning}

This stage involves updating the parameters of the LLM using a task-specific dataset. Full fine-tuning updates all parameters of the model, ensuring comprehensive adaptation to the new task. Alternatively, Half fine-tuning (HFT) \cite{hui2024hft} or Parameter-Efficient Fine-Tuning (PEFT) approaches, such as using adapter layers, can be employed to partially fine-tune the model. This method attaches additional layers to the pre-trained model, allowing for efficient fine-tuning with fewer parameters, which can address challenges related to computational efficiency, overfitting, and optimisation.

\section{Stage 5: Evaluation and Validation}

Evaluation and validation involve assessing the fine-tuned LLM's performance on unseen data to ensure it generalises well and meets the desired objectives. Evaluation metrics, such as cross-entropy, measure prediction errors, while validation monitors loss curves and other performance indicators to detect issues like overfitting or underfitting. This stage helps guide further fine-tuning to achieve optimal model performance.

\section{Stage 6: Deployment}

Deploying an LLM means making it operational and accessible for specific applications. This involves configuring the model to run efficiently on designated hardware or software platforms, ensuring it can handle tasks like natural language processing, text generation, or user query understanding. Deployment also includes setting up integration, security measures, and monitoring systems to ensure reliable and secure performance in real-world applications.

\section{Stage 7: Monitoring and Maintenance}

Monitoring and maintaining an LLM after deployment is crucial to ensure ongoing performance and reliability. This involves continuously tracking the model's performance, addressing any issues that arise, and updating the model as needed to adapt to new data or changing requirements. Effective monitoring and maintenance help sustain the model's accuracy and effectiveness over time.

\chapter{Stage 1: Data Preparation}

\section{Steps Involved in Data Preparation}

\subsection{Data Collection}

The first step in data preparation is to collect data from various sources. These sources can be in any format such as CSV, web pages, SQL databases, S3 storage, etc. Python provides several libraries to gather the data efficiently and accurately. Table \ref{tab:data_collection_libraries} presents a selection of commonly used data formats along with the corresponding Python libraries used for data collection.

\begin{table}[!ht]
\centering
\begin{tabularx}{\textwidth}{|p{2.5cm}|p{2.5cm}|X|p{3cm}|}
\hline
\textbf{Data Format} & \textbf{Python Library} & \textbf{Description} & \textbf{Library Link} \\ \hline
CSV Files & pandas & pandas is a powerful library for data manipulation and analysis. It provides the \texttt{read\_csv} function for easy and efficient reading of CSV files into DataFrame objects. It also supports reading data in Excel, JSON, and more. & \href{https://pandas.pydata.org/pandas-docs/stable/}{\color{blue}pandas documentation} \\ \hline
Web Pages & BeautifulSoup and requests & BeautifulSoup is a library for parsing HTML and XML documents. Combined with \texttt{requests} for sending HTTP requests, it enables data extraction from web pages, essential for web scraping tasks. & \href{https://www.crummy.com/software/BeautifulSoup/bs4/doc/}{\color{blue}BeautifulSoup documentation}, \href{https://requests.readthedocs.io/en/latest/}{\color{blue}requests documentation} \\ \hline
SQL Databases & SQLAlchemy & SQLAlchemy is a SQL toolkit and Object-Relational Mapping (ORM) library for Python, providing a full suite of enterprise-level persistence patterns. & \href{https://www.sqlalchemy.org/}{\color{blue}SQLAlchemy documentation} \\ \hline
S3 Storage & boto3 & boto3 is the Amazon Web Services (AWS) SDK for Python, allowing developers to use services like Amazon S3 and EC2. It enables interaction with AWS services, including uploading, downloading, and managing S3 bucket files. & \href{https://boto3.amazonaws.com/v1/documentation/api/latest/index.html}{\color{blue}boto3 documentation} \\ \hline
Data Integration & RapidMiner & RapidMiner is a comprehensive environment for data preparation, machine learning, and predictive analytics, allowing efficient processing and transformation of raw data into actionable insights. & \href{https://rapidminer.com/}{\color{blue}RapidMiner documentation} \\ \hline
Data Cleaning & Trifacta Wrangler & Trifacta Wrangler focuses on simplifying and automating data wrangling processes, transforming raw data into clean and structured formats. & \href{https://www.trifacta.com/}{\color{blue}Trifacta Wrangler documentation} \\ \hline
\end{tabularx}
\caption{Python libraries and tools for data collection and integration in various formats, providing an overview of commonly used libraries, their functions, and links to their official documentation for efficient data management and processing.}
\label{tab:data_collection_libraries}
\end{table}

\subsection{Data Preprocessing and Formatting}

Data preprocessing and formatting are crucial for ensuring high-quality data for fine-tuning. This step involves tasks such as cleaning the data, handling missing values, and formatting the data to match the specific requirements of the task. Several libraries assist with text data processing and Table \ref{tab:text_preprocessing_libraries} contains some of the most commonly used data preprocessing libraries in python.

\begin{table}[!ht]
\centering
\begin{tabularx}{\textwidth}{|l|X|l|}
\hline
\textbf{Library Name} & \textbf{Data Preprocessing Options} & \textbf{Link} \\ \hline
spaCy & spaCy provides robust capabilities for text preprocessing, including tokenization, lemmatization, and efficient sentence boundary detection. & \href{https://spacy.io/}{\color{blue}spaCy documentation} \\ \hline
NLTK & NLTK offers a comprehensive set of tools for data preprocessing, such as tokenization, stemming, and stop word removal. & \href{https://www.nltk.org/}{\color{blue}NLTK documentation} \\ \hline
HuggingFace & HuggingFace provides extensive capabilities for text preprocessing through its transformers library, including functionalities for tokenization and support for various pre-trained models. & \href{https://huggingface.co/}{\color{blue}HuggingFace documentation} \\ \hline
KNIME & KNIME Analytics Platform allows visual workflow design for data integration, preprocessing, and advanced manipulations like text mining and image analysis. & \href{https://www.knime.com/}{\color{blue}KNIME documentation} \\ \hline
\end{tabularx}
\caption{Outline of Python libraries commonly used for text data preprocessing, including spaCy, NLTK, HuggingFace, and KNIME. It details the specific preprocessing options offered by each library and provides links to their official documentation for users seeking more in-depth guidance on their use.}
\label{tab:text_preprocessing_libraries}
\end{table}

\subsection{Handling Data Imbalance}

Handling imbalanced datasets is crucial for ensuring balanced performance across all classes. Several techniques and strategies are employed:

\begin{enumerate}
    \item \textbf{Over-sampling and Under-sampling:} Techniques like SMOTE (Synthetic Minority Over-sampling Technique) generate synthetic examples to achieve balance. \\
    \textbf{Python Library:} \href{https://imbalanced-learn.org/stable/references/index.html}{\color{blue}imbalanced-learn} \\
    \textbf{Description:} imbalanced-learn provides various methods to deal with imbalanced datasets, including oversampling techniques like SMOTE.

    \item \textbf{Adjusting Loss Function:} Modify the loss function to give more weight to the minority class, setting class weights inversely proportional to the class frequencies.

    \item \textbf{Focal Loss:} A variant of cross-entropy loss that adds a factor to down-weight easy examples and focus training on hard negatives. \\
    \textbf{Python Library:} \href{https://pypi.org/project/focal-loss/}{\color{blue} focal\_loss} \\
    \textbf{Description:} The focal\_loss package provides robust implementations of various focal loss functions, including BinaryFocalLoss and SparseCategoricalFocalLoss.

    \item \textbf{Cost-sensitive Learning:} Incorporating the cost of misclassifications directly into the learning algorithm, assigning a higher cost to misclassifying minority class samples.

    \item \textbf{Ensemble Methods:} Using techniques like bagging and boosting to combine multiple models and handle class imbalance. \\
    \textbf{Python Library:} \href{https://scikit-learn.org/stable/modules/ensemble.html}{\color{blue}sklearn.ensemble} \\
    \textbf{Description:} scikit-learn provides robust implementations of various ensemble methods, including bagging and boosting.

    \item \textbf{Stratified Sampling:} Ensuring that each mini-batch during training contains an equal or proportional representation of each class. \\
    \textbf{Python Library:} \href{https://scikit-learn.org/stable/modules/generated/sklearn.model_selection.StratifiedShuffleSplit.html}{\color{blue}sklearn.model\_selection.StratifiedShuffleSplit} \\
    \textbf{Description:} scikit-learn offers tools for stratified sampling, ensuring balanced representation across classes.

    \item \textbf{Data Cleaning:} Removing noisy and mislabelled data, which can disproportionately affect the minority class. \\
    \textbf{Python Library:} \href{https://pandas.pydata.org/docs/reference/api/pandas.DataFrame.sample.html}{\color{blue}pandas.DataFrame.sample} \\
    \textbf{Description:} pandas provides methods for sampling data from DataFrames, useful for data cleaning and preprocessing.

    \item \textbf{Using Appropriate Metrics:} Metrics like Precision-Recall AUC, F1-score, and Cohen's Kappa are more informative than accuracy when dealing with imbalanced datasets. \\
    \textbf{Python Library:} \href{https://scikit-learn.org/stable/modules/model_evaluation.html}{\color{blue}sklearn.metrics} \\
    \textbf{Description:} scikit-learn offers a comprehensive set of tools for evaluating the performance of classification models, particularly with imbalanced datasets.
\end{enumerate}

\subsection{Splitting Dataset}

Splitting the dataset for fine-tuning involves dividing it into training and validation sets, typically using an 80:20 ratio. Different techniques include:
\begin{enumerate}
    \item \textbf{Random Sampling:} Selecting a subset of data randomly to create a representative sample. \\
    \textbf{Python Library:} \href{https://scikit-learn.org/stable/modules/generated/sklearn.model_selection.train_test_split.html}{\textcolor{blue}{sklearn.model\_selection.train\_test\_split}}

    \item \textbf{Stratified Sampling:} Dividing the dataset into subgroups and sampling from each to maintain class balance. \\
    \textbf{Python Library:} \href{https://scikit-learn.org/stable/modules/generated/sklearn.model_selection.StratifiedShuffleSplit.html}{\textcolor{blue}{sklearn.model\_selection.StratifiedShuffleSplit}}

    \item \textbf{K-Fold Cross Validation:} Splitting the dataset into K folds and performing training and validation K times. \\
    \textbf{Python Library:} \href{https://scikit-learn.org/stable/modules/generated/sklearn.model_selection.KFold.html}{\textcolor{blue}{sklearn.model\_selection.KFold}}

    \item \textbf{Leave-One-Out Cross Validation:} Using a single data point as the validation set and the rest for training, repeated for each data point. \\
    \textbf{Python Library:} \href{https://scikit-learn.org/stable/modules/generated/sklearn.model_selection.LeaveOneOut.html}{\textcolor{blue}{sklearn.model\_selection.LeaveOneOut}}
\end{enumerate}

Further details can be found in \href{https://scikit-learn.org/stable/api/sklearn.model_selection.html}{\textcolor{blue}{scikit-learn's documentation on model selection}}.

\section{Existing and Potential Research Methodologies}

\subsection{Data Annotation}

Data annotation involves labelling or tagging textual data with specific attributes relevant to the model's training objectives. This process is crucial for supervised learning tasks and greatly influences the performance of the fine-tuned model. Recent research highlights various approaches to data annotation:

\begin{itemize}
    \item \textbf{Human Annotation:} Manual annotation by human experts remains a gold standard due to its accuracy and context understanding. However, it is time-consuming and costly for large datasets \cite{snow2008cheap}. Tools like \textbf{Excel}, \textbf{Prodigy\footnote{\url{https://prodi.gy}}}, and \textbf{Innodata\footnote{\url{https://innodata.com/}}} facilitate this process.

    \item \textbf{Semi-automatic Annotation:} Combining machine learning algorithms with human review to create labelled datasets more efficiently. This approach balances efficiency and accuracy. Tools like \textbf{Snorkel\footnote{\url{https://snorkel.ai/}}} use weak supervision to generate initial labels, which are then refined by human annotators \cite{ratner2017snorkel}. 

    \item \textbf{Automatic Annotation:} Fully automated annotation leverages machine learning algorithms to label data without human intervention, offering scalability and cost-effectiveness. Services like \textbf{Amazon SageMaker Ground Truth\footnote{\url{https://aws.amazon.com/sagemaker/groundtruth/}}} utilise machine learning to automate data labelling, although the accuracy may vary depending on the complexity of the task \cite{ding2019automatic}.
\end{itemize}

\subsection{Data Augmentation}

Data Augmentation (DA) techniques expand training datasets artificially to address data scarcity and improve model performance. Advanced techniques often used in NLP include:

\begin{itemize}
    \item \textbf{Word Embeddings:} Using word embeddings like Word2Vec and GloVe to replace words with their semantic equivalents, thereby generating new data instances \cite{mikolov2013efficient, pennington2014glove}.
    
    \item \textbf{Back Translation:} Translating text to another language and then back to the original language to create paraphrased data. This technique helps in generating diverse training samples \cite{sennrich2016improving}. Tools like \textbf{Google Translate API\footnote{\url{https://translate.google.com/?sl=auto&tl=en&op=translate}}} are commonly used for this purpose.

    \item \textbf{Adversarial Attacks:} Generating augmented data through adversarial examples that slightly modify the original text to create new training samples while preserving the original meaning \cite{ebrahimi2017hotflip}. Libraries like \textbf{TextAttack\footnote{\url{https://github.com/QData/TextAttack}}} provide frameworks for such augmentations.
    
    \item \textbf{NLP-AUG\footnote{\url{https://github.com/makcedward/nlpaug}}:} This library offers a variety of augmenters for character, word, sentence, audio, and spectrogram augmentation, enhancing dataset diversity.
\end{itemize}

\subsection{Synthetic Data Generation using LLMs}

Large Language Models (LLMs) can generate synthetic data through innovative techniques such as:

\begin{itemize}
    \item \textbf{Prompt Engineering:} Crafting specific prompts to guide LLMs like GPT-3 in generating relevant and high-quality synthetic data \cite{brown2020language}.
    
    \item \textbf{Multi-Step Generation:} Employing iterative generation processes where LLMs generate initial data that is refined through subsequent steps \cite{gao2021making}. This method can produce high-quality synthetic data for various tasks, including summarising and bias detection.
    
    It is crucial to verify the accuracy and relevance of synthetic data generated by LLMs before using them for fine-tuning processes \cite{feng2021survey}.
\end{itemize}

\section{Challenges in Data Preparation for Fine-Tuning LLMs}

Key challenges in data preparation include:

\begin{enumerate}
    \item \textbf{Domain Relevance:} Ensuring that the data is relevant to the specific domain for accurate model performance. Mismatched domain data can lead to poor generalisation and inaccurate outputs \cite{gururangan2020don}.

    \item \textbf{Data Diversity:} Including diverse and well-balanced data to prevent model biases and improve generalisation. A lack of diversity can cause the model to perform poorly on underrepresented scenarios \cite{bender2021dangers}.

    \item \textbf{Data Size:} Managing and processing large datasets, with at least 1000 samples recommended for effective fine-tuning. However, large datasets pose challenges in terms of storage, computational requirements, and processing time.

    \item \textbf{Data Cleaning and Preprocessing:} Removing noise, errors, and inconsistencies are critical for providing clean inputs to the model. Poorly preprocessed data can degrade model performance significantly.

    \item \textbf{Data Annotation:} Ensuring precise and consistent labelling is essential for tasks requiring labelled data. Inconsistent annotation can lead to unreliable model predictions.

    \item \textbf{Handling Rare Cases:} Adequately representing rare but important instances in the dataset to ensure the model can generalise to less frequent but critical scenarios.

    \item \textbf{Ethical Considerations:} Scrutinising data for harmful or biased content to prevent unintended consequences. Ethical data handling includes removing biases and ensuring privacy \cite{binns2018fairness}.
\end{enumerate}

\section{Available LLM Fine-Tuning Datasets}

For a comprehensive list of datasets suitable for fine-tuning LLMs, refer to resources like \href{https://forms.gle/TNUbqHiCBsinD4Bu8}{\color{blue}LLMXplorer}, which provides domain and task-specific datasets.

\section{Best Practices}

\subsection{High-Quality Data Collection}
Ensuring high-quality, diverse, and representative data is critical. Leveraging curated sources and ensuring comprehensive coverage across different scenarios enhances model robustness \cite{ruder2021stanford}. Tools like \textbf{DataRobot Paxata\footnote{\url{https://www.datarobot.com/platform/preparation/}}} and \textbf{KNIME Analytics Platform\footnote{\url{https://www.knime.com/}}} offer robust data profiling and transformation capabilities.

\subsection{Effective Data Preprocessing}
Proper data preprocessing is essential for model performance. Utilising libraries like \textbf{spaCy}, \textbf{NLTK}, and \textbf{HuggingFace Transformers} can streamline preprocessing tasks. Platforms like \textbf{Trifacta Wrangler} and \textbf{RapidMiner} automate data cleaning tasks, improving efficiency and ensuring consistency \cite{rajan2019machine}.

\subsection{Managing Data Imbalance}
Addressing data imbalance is crucial. Techniques like over-sampling, under-sampling, and SMOTE help balance datasets. Libraries like \textbf{imbalanced-learn} and ensemble methods in \textbf{scikit-learn} provide robust tools for managing imbalanced datasets \cite{chawla2002smote}.

\subsection{Augmenting and Annotating Data}
Data augmentation and annotation improve model robustness. Tools like \textbf{NLP-AUG}, \textbf{TextAttack}, and \textbf{Snorkel} offer sophisticated capabilities for creating diverse and well-labelled datasets \cite{shorten2019survey, ratner2020snorkel}.

\subsection{Ethical Data Handling}
Ensuring ethical data handling involves thorough scrutiny for biases and privacy concerns. Implementing privacy-preserving techniques and filtering harmful content is critical. Services like \textbf{Amazon SageMaker Ground Truth} ensure scalable and secure data annotation \cite{barocas2017fairness}.

\subsection{Regular Evaluation and Iteration}
Continuous evaluation and iteration of the data preparation pipeline help maintain data quality and relevance. Leveraging feedback loops and performance metrics ensures ongoing improvements and adaptation to new data requirements.

\noindent By integrating these best practices, researchers and practitioners can enhance the effectiveness of LLM fine-tuning, ensuring robust and reliable model performance.

\chapter{Stage 2: Model Initialisation}

\section{Steps Involved in Model Initialisation}
\begin{figure}[h]
    \centering
    \includegraphics[width=1\textwidth]{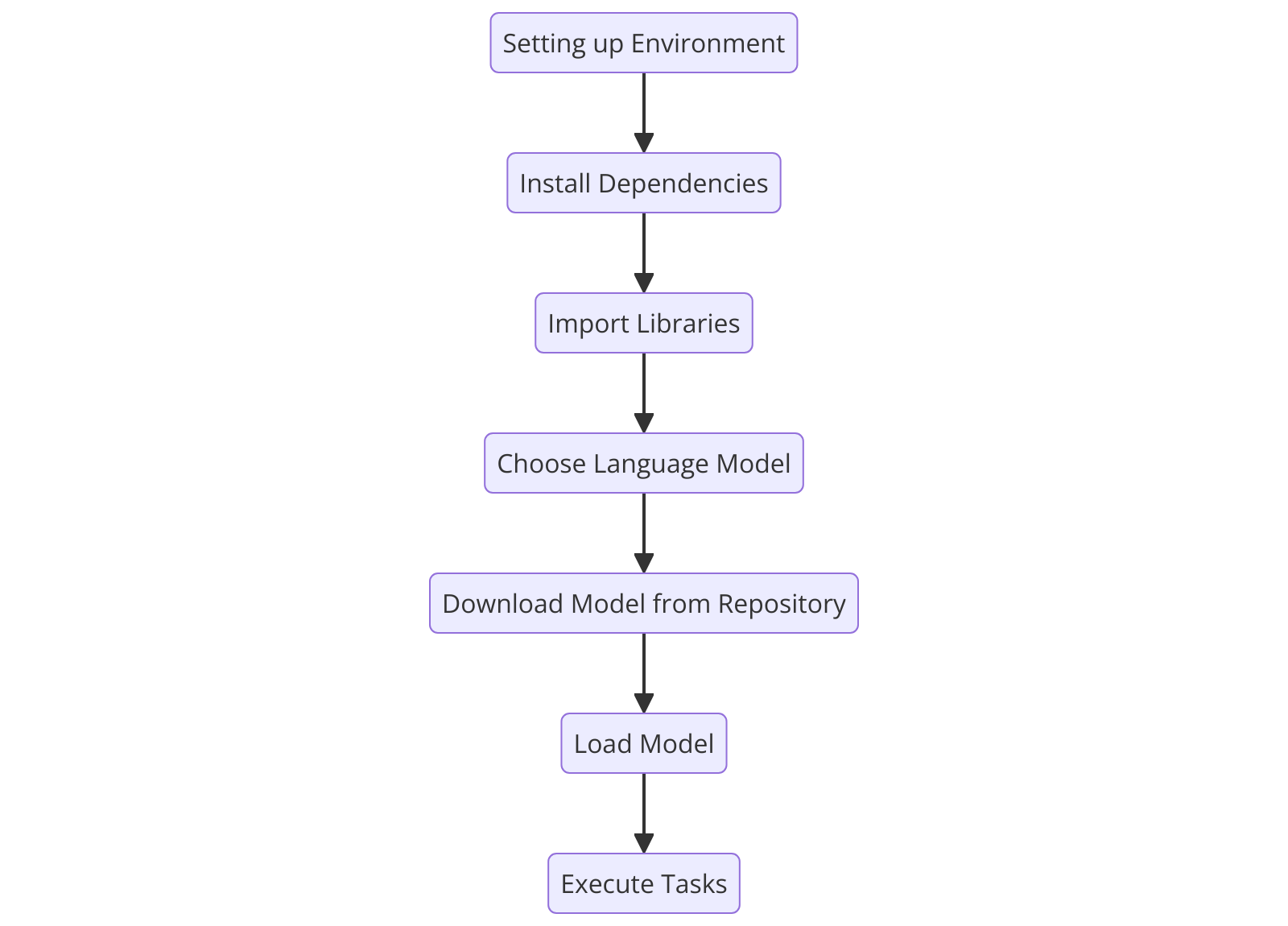} 
    \caption{Sequential steps involved in Initialising a Large Language Model (LLM), illustrating the process from setting up the environment to executing tasks. Each step is critical for ensuring that the LLM is correctly configured and ready for operation. This includes installing necessary dependencies, importing libraries, selecting and downloading the appropriate language model from a repository, and finally, loading the model to perform specific tasks.}
    \label{modelInitialisationSteps}
\end{figure}

\begin{enumerate}
    \item \textbf{Set Up the Environment:} Configure your environment, such as setting up GPU/TPU usage if available, which can significantly speed up model loading and inference.

    \item \textbf{Install the Dependencies:} Ensure that all necessary software and libraries are installed. This typically includes package managers like pip and frameworks like PyTorch or TensorFlow.

    \item \textbf{Import the Libraries:} Import the required libraries in your script or notebook. Common libraries include transformers from Hugging Face, torch for PyTorch, and other utility libraries.

    \item \textbf{Choose the Language Model:} Select the appropriate pre-trained language model based on your task requirements. This could be models like BERT, GPT-3, or others available on platforms like Hugging Face's Model Hub.

    \item \textbf{Download the Model from the Repository:} Use the chosen framework's functions to download the pre-trained model from an online repository. For instance, using transformers, you might use AutoModel.from\_pretrained('model\_name').

    \item \textbf{Load the Model in the Memory:} Load the model into memory, ready for inference or further fine-tuning. This step ensures the model weights are initialised and ready for use.

    \item \textbf{Execute Tasks:} Perform the desired tasks using the loaded model. This could involve making predictions, generating text, or fine-tuning the model on a new dataset.
\end{enumerate}

\section{Tools and Libraries for Model Initialisation}
Python offers a wide range of libraries for Initialising large language models, providing access to both open and closed-source models. Here are some notable libraries:

\begin{enumerate}
    \item \textbf{Python Library:} \href{https://huggingface.co/docs/transformers/en/index}{\color{blue} \emph{HuggingFace}}

    \textbf{Description:} HuggingFace is renowned for its support of numerous pre-trained large language models, ranging from Phi-3 mini to Llama-3 70B. The transformers library, part of HuggingFace, enables users to access these models via classes such as AutoModelForCausalLM. This library supports loading fine-tuned models as well as 4-bit quantised models. Additionally, the transformers library includes the "pipeline" feature, making it easy to use pre-trained models for various tasks \cite{wolf2020transformers}.

    \item \textbf{Python Framework:} \href{https://pytorch.org/docs/stable/index.html}{\color{blue} \emph{PyTorch}}

    \textbf{Description:} PyTorch offers comprehensive tools and libraries for Initialising and fine-tuning large language models. It provides a flexible and efficient platform for building and deploying deep learning models. HuggingFace’s transformers library bridges the gap between PyTorch and other frameworks, enhancing its usability for state-of-the-art language models \cite{paszke2019pytorch}.

    \item \textbf{Python Framework:} \href{https://www.tensorflow.org/tutorials}{\color{blue} \emph{TensorFlow}}

    \textbf{Description:} TensorFlow also provides extensive tools and libraries for Initialising and fine-tuning large language models. Similar to PyTorch, it benefits from the HuggingFace transformers library, which provides a versatile and user-friendly API and interface for working with the latest advancements in large language models \cite{tensorflow2015-whitepaper}.
\end{enumerate}

\section{Challenges in Model Initialisation}

\begin{table}[H]
\begin{tabularx}{1\textwidth}{|>{\raggedright\arraybackslash}p{4cm}|X|}
\hline
\multicolumn{1}{|c|}{\textbf{Challenge}} &
  \multicolumn{1}{c|}{\textbf{Description}} \\ \hline
Alignment with the Target Task &
  It's essential that the pre-trained model closely aligns with your specific task or domain. This initial alignment serves as a solid foundation for further fine-tuning efforts, leading to improved efficiency and results \cite{devlin2018bert}. \\ \hline
Understanding the Pre-trained Model &
  Before making a selection, it's crucial to thoroughly comprehend the architecture, capabilities, limitations, and the tasks the model was originally trained on. Without this understanding, fine-tuning efforts may not yield the desired outcomes \cite{brown2020language}. \\ \hline
Availability and Compatibility &
  Careful consideration of a model's documentation, license, maintenance, and update frequency is necessary to avoid potential issues and ensure smooth integration into your application. \\ \hline
Model Architecture &
  Not all models excel at every task. Each model architecture has its strengths and weaknesses, so selecting one aligned with your specific task is essential for favourable outcomes \cite{liu2019roberta}. \\ \hline
Resource Constraints &
  Loading pre-trained LLMs is resource-heavy and requires more computation. These models need high-performance CPUs and GPUs and a significant amount of disk space. For instance, the Llama 3 8B model requires a minimum of 16GB of memory to load and run the inference. \\ \hline
Privacy &
  Privacy and confidentiality are crucial factors when selecting a large language model (LLM). Many businesses prefer not to share their data with external LLM providers. In such instances, hosting an LLM on local servers or using pre-trained LLMs available through private cloud providers can be viable solutions. These approaches ensure that data remains within the company’s premises, thereby preserving privacy and confidentiality. \\ \hline
Cost and Maintenance &
  Hosting LLMs on local servers entails significant time and expense for setup and ongoing maintenance. Conversely, utilising cloud vendors alleviates concerns about resource maintenance but incurs monthly billing costs. These charges are typically based on factors such as model size and the volume of requests per minute. \\ \hline
Model Size and Quantisation &
  utilising a pre-trained model with high memory consumption can still be viable by employing its quantised version. Through quantisation, pre-trained weights can be loaded with reduced precision, typically 4-bit or 8-bit floating point, substantially diminishing parameter volume while maintaining considerable accuracy \cite{shen2020q}. \\ \hline
Pre-training Datasets &
  Examine the datasets used for pre-training to gauge the model's understanding of language. These are important as there are models available specifically for performing code generation, and we do not want to use those models for finance text classification \cite{radford2019language}. \\ \hline
Bias Awareness &
  Be vigilant regarding potential biases in pre-trained models, especially if unbiased predictions are required. The bias awareness can be evaluated by testing different models and backtracking the datasets used for pre-training \cite{gebru2021datasheets}. \\ \hline
\end{tabularx}
\caption{Comprehensive Overview of Challenges in Initialising a Large Language Model (LLM). This table highlights critical considerations, such as the importance of aligning pre-trained models with specific tasks, understanding model architecture and compatibility, managing resource constraints, and ensuring data privacy. Additionally, it discusses the challenges related to cost, maintenance, and the complexities of model size, quantisation, and bias awareness. Each challenge is associated with specific references to ensure thorough understanding and proper model deployment.}
\label{tab:modelInitChallenges}
\end{table}

\section{Tutorials}
\begin{enumerate}
    \item \href{https://medium.com/@manuelescobar-dev/implementing-and-running-llama-3-with-hugging-faces-transformers-library-40e9754d8c80}{\color{blue} Summarisation using Llama 3}

    \item \href{https://huggingface.co/docs/transformers/en/llm_tutorial}{\color{blue} HuggingFace tutorial for getting started with LLMs}

    \item \href{https://pytorch.org/tutorials/beginner/finetuning_torchvision_models_tutorial.html}{\color{blue} PyTorch tutorial for fine-tuning models}

    \item \href{https://www.tensorflow.org/tutorials/text/transformer}{\color{blue} TensorFlow tutorial for transformer models}
\end{enumerate}

\chapter{Stage 3: Training Setup}

\section{Steps Involved in Training Setup}

\begin{enumerate}
    \item \textbf{Setting up the training environment:} When setting up the environment for training an LLM, it is crucial to configure high-performance hardware, such as GPUs or TPUs, and ensure proper installation of necessary software components like CUDA, cuDNN, and deep learning frameworks such as PyTorch or TensorFlow. Verify hardware recognition and compatibility with the software to leverage computational power effectively, reducing training time and improving model performance.

    \item \textbf{Defining the Hyper-parameters:} When defining hyperparameters for fine-tuning an LLM, it is essential to carefully tune key parameters such as learning rate, batch size, and epochs to optimise the model's performance.

    \item \textbf{Initialising Optimisers and Loss Functions:} When initialising optimisers and loss functions for fine-tuning an LLM, it is crucial to select the appropriate optimiser to efficiently update the model's weights and the correct loss function to measure model performance \cite{kingma2014adam}.
\end{enumerate}

\section{Setting up Training Environment}

When fine-tuning a large language model (LLM), the computational environment plays a crucial role in ensuring efficient training. To achieve optimal performance, it's essential to configure the environment with high-performance hardware such as GPUs (Graphics Processing Units) or TPUs (Tensor Processing Units). GPUs, such as the NVIDIA A100 or V100, are widely used for training deep learning models due to their parallel processing capabilities. For larger-scale operations, TPUs offered by Google Cloud can provide even greater acceleration \cite{jouppi2017datacenter}. \\

\noindent First, ensure that your system or cloud environment has the necessary hardware installed. For GPUs, this involves setting up CUDA\footnote{\url{https://developer.nvidia.com/cuda-toolkit}} (Compute Unified Device Architecture) and cuDNN\footnote{\url{https://developer.nvidia.com/cudnn}} (CUDA Deep Neural Network library) from NVIDIA, which are essential for enabling GPU acceleration. For TPU usage, you would typically set up a Google Cloud environment with TPU instances, which includes configuring the TPU runtime in your training scripts. \\

\noindent Verify that your hardware is correctly recognised and utilised by your deep learning frameworks. In PyTorch, for instance, you can check GPU availability with torch.cuda.is\_available(). Properly setting up and testing the hardware ensures that the training process can leverage the computational power effectively, reducing training time and improving model performance \cite{paszke2019pytorch}. \\

\noindent When fine-tuning an LLM, both software and hardware considerations are paramount to ensure a smooth and efficient training process. On the software side, you need a compatible deep learning framework like PyTorch or TensorFlow. These frameworks have extensive support for LLMs and provide utilities for efficient model training and evaluation. Installing the latest versions of these frameworks, along with any necessary dependencies, is crucial for leveraging the latest features and performance improvements \cite{abadi2016tensorflow}. \\

\noindent Additionally, use libraries like Hugging Face’s transformers to simplify the process of loading pre-trained models and tokenizers. This library is particularly well-suited for working with various LLMs and offers a user-friendly interface for model fine-tuning. Ensure that all software components, including libraries and dependencies, are compatible with your chosen framework and hardware setup \cite{wolf2020transformers}. \\

\noindent On the hardware side, consider the memory requirements of the model and your dataset. LLMs typically require substantial GPU memory, so opting for GPUs with higher VRAM (e.g., 16GB or more) can be beneficial. If your model is exceptionally large or if you are training with very large datasets, distributed training across multiple GPUs or TPUs might be necessary. This requires a careful setup of data parallelism or model parallelism techniques to efficiently utilise the available hardware \cite{shoeybi2019megatron}. \\

\noindent Lastly, ensure robust cooling and power supply for your hardware, as training LLMs can be resource-intensive, generating significant heat and requiring consistent power. Proper hardware setup not only enhances training performance but also prolongs the lifespan of your equipment \cite{you2019large}.

\section{Defining Hyperparameters}

Key hyperparameters like learning rate, batch size, epochs are crucial for enhancing the model’s performance and obtaining superior outcomes. This process entails adjusting hyperparameters and training settings to align with your particular use case. Below are the key hyperparameters:

\begin{enumerate}
    \item \textbf{Learning Rate:} Fine-tuning an LLM involves using optimisation algorithms like stochastic gradient descent (SGD). This technique estimates the error gradient for the model's current state using samples from the training dataset and subsequently updates the model's weights via the backpropagation of errors algorithm. The learning rate dictates the speed at which the model adapts to the problem. Smaller learning rates necessitate more training due to the minimal weight adjustments per update, while larger learning rates lead to quicker changes to weights \cite{goodfellow2016deep}.

    \item \textbf{Batch Size:} A batch refers to a subset of the training data used to update a model’s weights during the training process. Batch training involves dividing the entire training set into smaller groups, updating the model after processing each batch. The batch size is a hyperparameter that determines the number of samples processed before the model parameters are updated.

    \item \textbf{Epochs:} Epoch refers to a full pass through the entire training dataset. This involves a complete forward and backward pass through the dataset. The dataset can be processed as a single batch or divided into multiple smaller batches. An epoch is considered complete once the model has processed all batches and updated its parameters based on the calculated loss.
\end{enumerate}

\subsection{Methods for Hyperparameter Tuning}

LLM hyperparameter tuning involves adjusting various hyperparameters during the training process to identify the optimal combination that yields the best output. This process often entails significant trial and error, meticulously tracking each hyperparameter adjustment, and recording the resulting performance. Conducting this manually can be highly time-consuming. To address this, automated hyperparameter tuning methods have been developed to streamline the process. The three most common methods of automated hyperparameter tuning are random search, grid search, and Bayesian optimisation:

\begin{enumerate}
    \item \textbf{Random Search:} This method randomly selects and evaluates combinations of hyperparameters from a specified range. It is a straightforward and efficient approach capable of exploring a large parameter space. However, it may not always find the optimal combination of hyperparameters and can be computationally expensive \cite{bergstra2012random}.

    \item \textbf{Grid Search:} Unlike random search, grid search exhaustively evaluates every possible combination of hyperparameters from a given range. Although resource-intensive, this systematic approach ensures that the optimal set of hyperparameters is found \cite{hutter2019automated}.

    \item \textbf{Bayesian Optimisation:} This method uses a probabilistic model to predict the performance of different hyperparameters and selects the best ones accordingly. It is an efficient method that can handle large parameter spaces better and is less resource-intensive than grid search. However, it is more complex to set up and may be less reliable in identifying the optimal set of hyperparameters compared to grid search.

    \item \textbf{Automated hyperparameter tuning:} This facilitates the development of multiple language models, each with a unique combination of hyperparameters. By training these models on the same dataset, it becomes possible to compare their outputs and determine which configuration is best suited for the desired use case. Additionally, models tuned with different sets of hyperparameters can be tailored to various specific applications.
\end{enumerate}

\section{Initialising Optimisers and Loss Functions}

Choosing the right optimiser and loss function is crucial for training and fine-tuning LLMs. Below are descriptions of some commonly used optimisation algorithms, their advantages, disadvantages, and appropriate use cases:

\subsection{Gradient Descent}

Gradient Descent is a fundamental optimisation algorithm used to minimise cost functions in machine learning models. It aims to find the optimal parameters for a neural network.

\textbf{How it Works:}
Gradient Descent iteratively updates model parameters in the direction of the negative gradient of the cost function. It calculates gradients for each parameter and applies updates across all data points until convergence. This method utilises the entire dataset to calculate gradients, often requiring a fixed learning rate and being sensitive to the scale of data and learning rate choice.

\textbf{Pros:}
\begin{itemize}
    \item Simple and easy to implement.
    \item Intuitive and easy to understand.
    \item Converges to the global minimum for convex functions.
    \item Suitable for small-scale problems.
\end{itemize}

\textbf{Cons:}
\begin{itemize}
    \item Computationally expensive on large datasets.
    \item May get stuck in local minima.
    \item Requires a large number of iterations.
    \item Sensitive to the choice of learning rate.
\end{itemize}

\textbf{When to Use:}
Gradient Descent is best used for small datasets where gradient computation is cheap and simplicity and clarity are preferred.

\subsection{Stochastic Gradient Descent (SGD)}

Stochastic Gradient Descent (SGD) is a variant of Gradient Descent that focuses on reducing computation per iteration.

\textbf{How it Works:}
SGD updates parameters using a single or few data points at each iteration, introducing randomness in updates. It reduces the computational burden per iteration and often converges faster than batch Gradient Descent. However, it requires a smaller learning rate due to higher variance and benefits from momentum to stabilise updates.

\textbf{Pros:}
\begin{itemize}
    \item Fast and handles large datasets well.
    \item Efficient memory usage.
    \item Simple and easy to implement.
    \item Can escape local minima due to noise.
\end{itemize}

\textbf{Cons:}
\begin{itemize}
    \item High variance in updates can lead to instability.
    \item Can overshoot the minimum.
    \item Sensitive to the choice of learning rate.
    \item Can be slower to converge compared to batch methods.
\end{itemize}

\textbf{When to Use:}
SGD is ideal for large datasets, incremental learning scenarios, and real-time learning environments where computational resources are limited.

\subsection{Mini-batch Gradient Descent}

Mini-batch Gradient Descent combines the efficiency of SGD and the stability of batch Gradient Descent, offering a compromise between batch and stochastic approaches.

\textbf{How it Works:}
It splits data into small batches and updates parameters using gradients averaged over each mini-batch. This reduces variance compared to SGD and is more efficient than batch Gradient Descent, helping in generalising the updates.

\textbf{Pros:}
\begin{itemize}
    \item Balances between efficiency and stability.
    \item More generalisable updates.
    \item Reduces the variance of parameter updates.
    \item Provides a compromise between SGD and batch.
\end{itemize}

\textbf{Cons:}
\begin{itemize}
    \item Requires tuning of batch size.
    \item Can still be computationally expensive for very large datasets.
    \item More complex implementation.
    \item Can require more iterations than full-batch Gradient Descent.
\end{itemize}

\textbf{When to Use:}
Mini-batch Gradient Descent is suitable for most deep learning tasks, especially when working with moderate to large datasets.

\subsection{AdaGrad}

Adaptive Gradient Algorithm (AdaGrad) is designed for sparse data and high-dimensional models, adjusting learning rates to improve performance on sparse data.

\textbf{How it Works:}
AdaGrad adapts the learning rate for each parameter based on historical gradient information, accumulating squared gradients. This approach prevents large updates for frequent parameters and helps in dealing with sparse features.

\textbf{Pros:}
\begin{itemize}
    \item Adapts learning rate for each parameter.
    \item Good for sparse data.
    \item No need to manually tune learning rates.
    \item Works well with high-dimensional data.
\end{itemize}

\textbf{Cons:}
\begin{itemize}
    \item Learning rate can diminish to zero, stopping learning.
    \item May require more tuning for convergence.
    \item Accumulation of squared gradients can lead to overly small learning rates.
    \item Can slow down significantly.
\end{itemize}

\textbf{When to Use:}
AdaGrad is useful for sparse datasets like text and images where learning rates need to adapt to feature frequency.

\subsection{RMSprop}

Root Mean Square Propagation (RMSprop) is an adaptive learning rate method designed to perform better on non-stationary and online problems.

\textbf{How it Works:}
RMSprop modifies AdaGrad by using a moving average of squared gradients to adapt learning rates based on recent gradient magnitudes. It maintains a running average of squared gradients to help in maintaining steady learning rates.

\textbf{Pros:}
\begin{itemize}
    \item Addresses the diminishing learning rate problem of AdaGrad.
    \item Adapts learning rate based on recent gradients.
    \item Effective for recurrent neural networks.
    \item More robust against non-stationary targets.
\end{itemize}

\textbf{Cons:}
\begin{itemize}
    \item Can still get stuck in local minima on non-convex problems.
    \item Requires hyperparameter tuning.
    \item Requires careful tuning of the decay rate.
    \item Can be sensitive to the initial learning rate.
\end{itemize}

\textbf{When to Use:}
RMSprop is best for non-convex optimisation problems, training RNNs and LSTMs, and dealing with noisy or non-stationary objectives.

\subsection{AdaDelta}

Adaptive Delta (AdaDelta) improves on AdaGrad and RMSprop, focusing on adaptive learning rates without diminishing too quickly.

\textbf{How it Works:}
AdaDelta eliminates the need for a default learning rate by using a moving window of gradient updates. It adapts learning rates based on recent gradient magnitudes to ensure consistent updates even with sparse gradients.

\textbf{Pros:}
\begin{itemize}
    \item Eliminates the need to set a default learning rate.
    \item Addresses the diminishing learning rate issue.
    \item Does not require manual tuning of the learning rate.
    \item Handles gradient sparsity well.
\end{itemize}

\textbf{Cons:}
\begin{itemize}
    \item More complex than RMSprop and AdaGrad.
    \item Can have slower convergence initially.
    \item Can require more iterations to converge.
    \item Implementation can be more complex.
\end{itemize}

\textbf{When to Use:}
AdaDelta is suitable for scenarios similar to RMSprop but is preferred when avoiding manual learning rate setting.

\subsection{Adam}

Adaptive Moment Estimation (Adam) combines the advantages of AdaGrad and RMSprop, making it suitable for problems with large datasets and high-dimensional spaces.

\textbf{How it Works:}
Adam uses running averages of both gradients and their squared values to compute adaptive learning rates for each parameter. It includes bias correction and often achieves faster convergence than other methods.

\textbf{Pros:}
\begin{itemize}
    \item Combines advantages of AdaGrad and RMSprop.
    \item Adaptive learning rates.
    \item Includes bias correction.
    \item Fast convergence.
    \item Works well with large datasets and high-dimensional spaces.
\end{itemize}

\textbf{Cons:}
\begin{itemize}
    \item Requires tuning of hyperparameters (though it often works well with defaults).
    \item Computationally intensive.
    \item Can lead to overfitting if not regularised properly.
    \item Requires more memory.
\end{itemize}

\textbf{When to Use:}
Adam is widely used in most deep learning applications due to its efficiency and effectiveness, particularly in complex neural network architectures.

\subsection{AdamW}

AdamW is an extension of Adam that includes weight decay regularisation to address overfitting issues present in Adam.

\textbf{How it Works:}
AdamW integrates L2 regularisation directly into the parameter updates, decoupling weight decay from the learning rate. This improves generalisation and is suitable for fine-tuning large models.

\textbf{Pros:}
\begin{itemize}
    \item Includes weight decay for better regularisation.
    \item Combines Adam’s adaptive learning rate with L2 regularisation.
    \item Improves generalisation.
    \item Reduces overfitting compared to Adam.
\end{itemize}

\textbf{Cons:}
\begin{itemize}
    \item Slightly more complex than Adam.
    \item Requires careful tuning of the weight decay parameter.
    \item Slightly slower than Adam due to additional computations.
    \item Requires more memory.
\end{itemize}

\textbf{When to Use:}
AdamW is ideal for scenarios where regularisation is needed, such as preventing overfitting in large models and fine-tuning pre-trained models.

A comprehensive collection of optimisation algorithms implemented within the PyTorch library can be found in \href{https://pytorch.org/docs/stable/optim.html}{\color{blue}here}. The Hugging Face Transformers package also offers a variety of optimisers for initialising and fine-tuning language models, available \href{https://huggingface.co/docs/transformers/en/main_classes/optimiser_schedules}{\color{blue}here}.

\section{Challenges in Training Setup}

\begin{enumerate}
    \item Ensuring compatibility and proper configuration of high-performance hardware like GPUs or TPUs can be complex and time-consuming.
    
    \item Managing dependencies and versions of deep learning frameworks and libraries to avoid conflicts and leverage the latest features.
    
    \item Selecting an appropriate learning rate is critical, as too high a rate can cause suboptimal convergence, while too low a rate can make the training process excessively slow.
    
    \item Determining the optimal batch size that balances memory constraints and training efficiency, especially given the large memory requirements of LLMs.
    
    \item Choosing the right number of epochs to avoid underfitting or overfitting the model, requiring careful monitoring and validation.
    
    \item Selecting the most suitable optimiser for the specific training task to efficiently update the model's weights.
    
    \item Choosing the correct loss function to accurately measure model performance and guide the optimisation process.
\end{enumerate}

\section{Best Practices}

\begin{itemize}
    \item \textbf{Optimal Learning Rate:} Use a lower learning rate, typically between 1e-4 to 2e-4, to ensure stable convergence. A learning rate schedule, such as learning rate warm-up followed by a linear decay, can also be beneficial. This helps in initially stabilising the training and then allowing the model to converge more accurately.

    \item \textbf{Batch Size Considerations:} Opt for a batch size that balances memory constraints and training efficiency. Smaller batch sizes can help in achieving faster convergence but may require more frequent updates. Conversely, larger batch sizes can be more memory-intensive but may lead to more stable updates. Experiment with different batch sizes to find the optimal balance for your specific use case.

    \item \textbf{Save Checkpoints Regularly:} Regularly save model weights at various intervals across 5-8 epochs to capture optimal performance without overfitting. Implement early stopping mechanisms to halt training once the model performance starts to degrade on the validation set, thereby preventing overfitting \cite{prechelt1998early}.

    \item \textbf{Hyperparameter Tuning:} Utilise hyperparameter tuning methods like grid search, random search, and Bayesian optimisation to find the optimal set of hyperparameters. Tools such as Optuna, Hyperopt, and Ray Tune can automate this process and help in efficiently exploring the hyperparameter space \cite{bergstra2012random}.

    \item \textbf{Data Parallelism and Model Parallelism:} For large-scale training, consider using data parallelism or model parallelism techniques to distribute the training workload across multiple GPUs or TPUs. Libraries like Horovod and DeepSpeed can facilitate efficient distributed training, helping to reduce training time and manage memory usage effectively \cite{sergeev2018horovod, rajbhandari2020deepspeed}.

    \item \textbf{Regular Monitoring and Logging:} Implement robust monitoring and logging to track training metrics, resource usage, and potential bottlenecks. Tools like TensorBoard, Weights \& Biases, and MLflow can provide real-time insights into the training process, allowing for timely interventions and adjustments.

    \item \textbf{Handling Overfitting and Underfitting:} Ensure that your model generalises well by implementing techniques to handle overfitting and underfitting. regularisation techniques such as L2 regularisation, dropout, and data augmentation can help prevent overfitting. Conversely, if your model is underfitting, consider increasing the model complexity or training for more epochs.

    \item \textbf{Use Mixed Precision Training:} Mixed precision training involves using both 16-bit and 32-bit floating-point types to reduce memory usage and increase computational efficiency. This technique can significantly speed up training and reduce the required memory footprint, especially when using large models. NVIDIA’s Apex and TensorFlow’s mixed precision API provide support for implementing mixed precision training \cite{micikevicius2018mixed}.

    \item \textbf{Evaluate and Iterate:} Continuously evaluate the model performance using a separate validation set and iterate on the training process based on the results. Regularly update your training data and retrain the model to keep it current with new data trends and patterns.

    \item \textbf{Documentation and Reproducibility:} Maintain thorough documentation of your training setup, including the hardware configuration, software environment, and hyperparameters used. Ensure reproducibility by setting random seeds and providing detailed records of the training process. This practice not only aids in debugging and further development but also facilitates collaboration and sharing of results with the broader research community.
\end{itemize}

\chapter{Stage 4: Selection of Fine-Tuning Techniques and Appropriate Model Configurations}

This chapter focuses on selecting appropriate fine-tuning techniques and model configurations that suit the specific requirements of various tasks. Fine-tuning is a crucial stage where pre-trained models are adapted to specific tasks or domains. 

\section{Steps Involved in Fine-Tuning}

The following steps outline the fine-tuning process, integrating advanced techniques and best practices.

\begin{enumerate}
    \item \textbf{Initialise the Pre-Trained Tokenizer and Model:} 
    Begin by loading the pre-trained tokenizer and model. The tokenizer ensures that the input text is converted into a format the model can process, while the pre-trained model serves as the foundation for further adaptation. Depending on the task, select a model that has been pre-trained on relevant data to provide a strong starting point.

    \item \textbf{Modify the Model's Output Layer:}
    Adjust the model’s output layer to align with the specific requirements of the target task. This may involve modifying existing layers or adding new layers. For instance, tasks like classification may require a softmax layer with the appropriate number of classes, while text generation tasks might involve changes in the decoding mechanism.

    \item \textbf{Choose an Appropriate Fine-Tuning Strategy:}
    Select the fine-tuning strategy that best fits the task and the model architecture. Some Options include:
    \begin{itemize}
        \item \textbf{Task-Specific Fine-Tuning:} For tasks such as text summarisation, code generation, classification, and question answering, adapt the model using relevant datasets.
        \item \textbf{Domain-Specific Fine-Tuning:} Tailor the model to comprehend and generate text relevant to specific domains, such as medical, financial, or legal fields.
        \item \textbf{Parameter-Efficient Fine-Tuning (PEFT):} Techniques like LoRA, QLoRA, and adapters allow for fine-tuning with reduced computational costs by updating a small subset of model parameters.
        \item \textbf{Half Fine-Tuning (HFT):} Balance between retaining pre-trained knowledge and learning new tasks by updating only half of the model’s parameters during each fine-tuning round.
    \end{itemize}

    \item \textbf{Set Up the Training Loop:}
    Establish the training loop, incorporating the selected fine-tuning strategy. The loop should include data loading, loss computation, backpropagation, and parameter updates. When using PEFT methods, ensure that only the relevant parameters are updated to maximise efficiency. Implement techniques like dynamic learning rates and early stopping to enhance the training process.

    \item \textbf{Incorporate Techniques for Handling Multiple Tasks:}
    If fine-tuning for multiple tasks, consider strategies like fine-tuning with multiple adapters or leveraging Mixture of Experts (MoE) architectures. These methods allow a single model to handle various tasks by utilising specialised sub-networks or adapters for each task.

    \item \textbf{Monitor Performance on a Validation Set:}
    Regularly evaluate the model’s performance on a validation set to ensure it generalises well to unseen data. Adjust hyperparameters such as learning rate, batch size, and dropout rates based on the validation performance. Utilise advanced monitoring tools to track metrics like accuracy, loss, and overfitting.

    \item \textbf{Optimise Model Using Advanced Techniques:}
    Employ techniques such as Proximal Policy Optimisation (PPO) for reinforcement learning scenarios, or Direct Preference Optimisation (DPO) for aligning model outputs with human preferences. These techniques are particularly useful in fine-tuning models for tasks requiring nuanced decision-making or human-like responses.

    \item \textbf{Prune and optimise the Model (if necessary):}
    To deploy the model in resource-constrained environments, consider pruning techniques to reduce its size and complexity. This involves removing unnecessary parameters or components without significantly affecting performance. Utilise dynamic pruning methods during inference to optimise the model on-the-fly for different scenarios.

    \item \textbf{Continuous Evaluation and Iteration:}
    Continuously evaluate the model’s performance across various tasks using appropriate benchmarks. Iterate on the fine-tuning process, making adjustments based on performance metrics and real-world testing. This iterative approach helps in refining the model to meet specific performance criteria.

\end{enumerate}

\section{Fine-Tuning Strategies for LLMs}

\subsection{Task-Specific Fine-Tuning}

Task-specific fine-tuning adapts large language models (LLMs) for particular downstream tasks using appropriately formatted and cleaned data. Below is a summary of key tasks suitable for fine-tuning LLMs, including examples of LLMs tailored to these tasks.

\begin{table}[h]
\centering
\begin{tabularx}{\textwidth}{|l|X|l|}
\hline
\textbf{Task} & \textbf{Description} & \textbf{Key Models} \\ \hline
\textbf{Text Summarisation} & Condensing long texts into coherent summaries while retaining key information. Approaches include Extractive (selecting key sentences) and Abstractive summarisation (generating new sentences). & BERTSUM, GPT-3, T5 \\ \hline
\textbf{Code Generation} & Automatically generating programming code based on natural language descriptions, partial code snippets, or structured data inputs. & Codex, GPT-3, CodeBERT \\ \hline
\textbf{Classification} & Categorising text into predefined labels such as Sentiment Analysis, Topic Classification, and Entity Classification. & BERT, RoBERTa, GPT-4 \\ \hline
\textbf{Q\&A} & Understanding and generating accurate, contextually relevant answers to natural language questions. & BERT, GPT-3, T5 \\ \hline
\end{tabularx}
\caption{Overview of tasks such as text summarisation, code generation, classification, and Q\&A, along with their key LLMs and descriptions.}
\label{tab:tasksAndDescriptions}
\end{table}

\subsection{Domain-Specific Fine-Tuning}

Domain-specific fine-tuning focuses on tailoring the model to comprehend and produce text relevant to a specific domain or industry. By fine-tuning the model on a dataset derived from the target domain, it enhances the model's contextual understanding and expertise in domain-specific tasks. Below are examples of domain-specific LLMs.

\subsubsection{Medical Domain}

\textbf{Model Description:} Med-PaLM 2 is trained on meticulously curated medical datasets and is capable of accurately answering medical questions, achieving performance comparable to that of medical professionals \cite{medPalm2Model}. \\
\textbf{Base Model:} PaLM 2 \\
\textbf{Fine-tuned Model Parameters:} Not Known \\
\textbf{Fine-Tuning Techniques Used:} Instruction fine-tuning \\
\textbf{Datasets Used:} 
\begin{itemize}
    \item MedQA 
    \item MedMCQA
    \item LiveQA
    \item MedicationQA
    \item HealthSearchQA
\end{itemize}
\textbf{Results:} Med-PaLM 2 outperformed GPT-4 in several key medical benchmarks, demonstrating superior performance in handling complex medical knowledge and reasoning tasks.

\subsubsection{Finance Domain}

\textbf{Model Description:} FinGPT, an open-source LLM tailored for the financial sector, enhances financial research and cooperation by promoting data accessibility and handling finance-specific issues like data acquisition and quality \cite{finGPTModel}. \\
\textbf{Base Model:} LlaMA, ChatGLM, and other Transformer Models \\
\textbf{Fine-tuned Model Parameters:} Not Known \\
\textbf{Fine-Tuning Techniques Used:} LoRA, Reinforcement Learning on Stock Prices (RLSP) \\
\textbf{Datasets Used:} 
\begin{itemize}
    \item Financial News (Reuters, CNBC, Yahoo Finance)
    \item Social Media (Twitter, Facebook, Reddit, Weibo)
    \item Regulatory Filings (e.g., SEC filings)
    \item Trends (Seeking Alpha, Google Trends)
    \item Academic Datasets
\end{itemize}
\textbf{Results:} Not Applicable

\subsubsection{Legal Domain}

\textbf{Model Description:} LAWGPT, the first open-source model specifically designed for Chinese legal applications, demonstrates superior capability in handling Chinese legal tasks \cite{lawGPTPaper}. \\
\textbf{Base Model:} Chinese Alpaca Plus 7B base model \\
\textbf{Fine-tuned Model Parameters:} Not Known \\
\textbf{Fine-Tuning Techniques Used:} LoRA with Alpaca template \\
\textbf{Datasets Used:} 
\begin{itemize}
    \item Open-source dataset: 200,000 examples containing crime type prediction and crime consultation tasks.
    \item JEC-QA dataset: 20,000 examples containing legal question answering tasks.
    \item Constructed legal dataset: 80,000 examples, refined from open-source and JEC-QA datasets using ChatGPT.
\end{itemize}
\textbf{Results:} LAWGPT demonstrates notable performance improvements over the LLaMA 7B model in various legal tasks, but still trails behind proprietary models like GPT-3.5 Turbo and GPT-4.

\subsubsection{Pharmaceutical Domain}

\textbf{Model Description:} PharmaGPT, a suite of domain-specific large language models tailored to the biopharmaceutical and chemical industries, sets a new benchmark for precision in these fields \cite{pharmGPTPaper}. \\
\textbf{Base Model:} LlaMA series \\
\textbf{Fine-tuned Model Parameters:} 13B and 70B \\
\textbf{Fine-Tuning Techniques Used:} Instruction fine-tuning and RLHF \\
\textbf{Datasets Used:} 
\begin{itemize}
    \item Specific-domain data from academic papers and clinical reports
    \item Text data from NLP dataset formats (e.g., question answering, summarisation, dialogue)
    \item Instruction fine-tuning dataset for multitask learning
    \item RLHF dataset with human preference expert-annotated instructions
\end{itemize}
\textbf{Results:} PharmaGPT models demonstrated impressive performance on various pharmaceutical benchmarks, consistently outperforming GPT-3.5 Turbo.

\subsubsection{Finance Domain}

\textbf{Model Description:} Palmyra-Fin-70B-32K, developed by Writer, is a leading large language model specifically designed for the financial sector. \cite{palmyraFinPaper} \\
\textbf{Base Model:} LlaMA \\
\textbf{Fine-tuned Model Parameters:} 70B \\
\textbf{Fine-Tuning Techniques Used:} Not Known \\
\textbf{Datasets Used:} Not Known \\
\textbf{Results:} Palmyra-Fin-70B-32K exhibits state-of-the-art performance, achieving leading results across various financial datasets and excelling in financial document analysis, market trend prediction, and risk assessment.

\section{Parameter-Efficient Fine-Tuning (PEFT) Techniques}

\begin{figure}[h]
\centering
\includegraphics[width=1\textwidth]{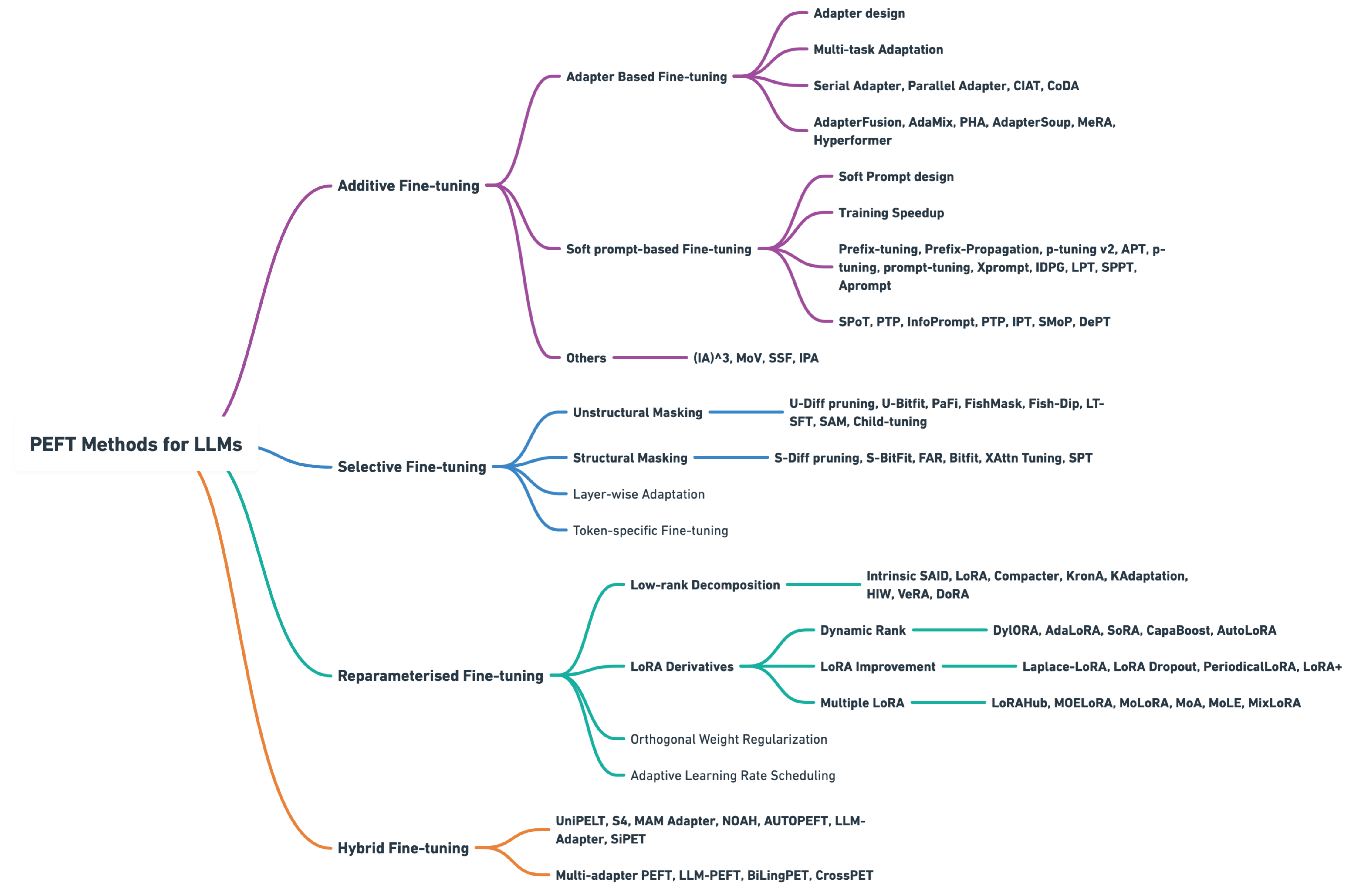}
\caption{Comprehensive Taxonomy of Parameter-Efficient Fine-Tuning (PEFT) Methods for Large Language Models (LLMs). This figure categorises various PEFT techniques, highlighting their distinct approaches, from additive and selective fine-tuning to reparameterised and hybrid methods. It details specific strategies within each category, such as Adapter-Based Fine-Tuning, Soft Prompt-Based Fine-Tuning, and their respective sub-techniques like LoRA and its derivatives, showcasing the diverse and evolving landscape of LLM fine-tuning. (adapted from \cite{surveyOfPEFT})}
\label{surveyOfPEFT}
\end{figure}

Parameter Efficient Fine Tuning \href{https://github.com/huggingface/peft}{\color{blue} (PEFT)} is an impactful NLP technique that adeptly adapts pre-trained language models to various applications with remarkable efficiency. PEFT methods fine-tune only a small subset of (additional) model parameters while keeping most of the pre-trained LLM parameters frozen, thereby significantly reducing computational and storage costs. This approach mitigates the issue of catastrophic forgetting, a phenomenon where neural networks lose previously acquired knowledge and experience a significant performance decline on previously learned tasks when trained on new datasets. PEFT methods have demonstrated superior performance compared to full fine-tuning, particularly in low-data scenarios, and exhibit better generalisation to out-of-domain contexts. This technique is applicable to various modalities, such as financial sentiment classification and machine translation of medical terminologies. A taxonomy of PEFT-based fine-tuning approaches is provided in Figure\ref{surveyOfPEFT}. We will further discuss a few key PEFT-based approaches in the following sections. 

\subsection{Adapters}
Adapter-based methods introduce additional trainable parameters after the attention and fully connected layers of a frozen pre-trained model, aiming to reduce memory usage and accelerate training. The specific approach varies depending on the adapter; it might involve adding an extra layer or representing the weight updates delta (W) as a low-rank decomposition of the weight matrix. Regardless of the method, adapters are generally small yet achieve performance comparable to fully fine-tuned models, allowing for the training of larger models with fewer resources. \\

\begin{figure}[h]
\centering

\includegraphics[width=0.45\textwidth]{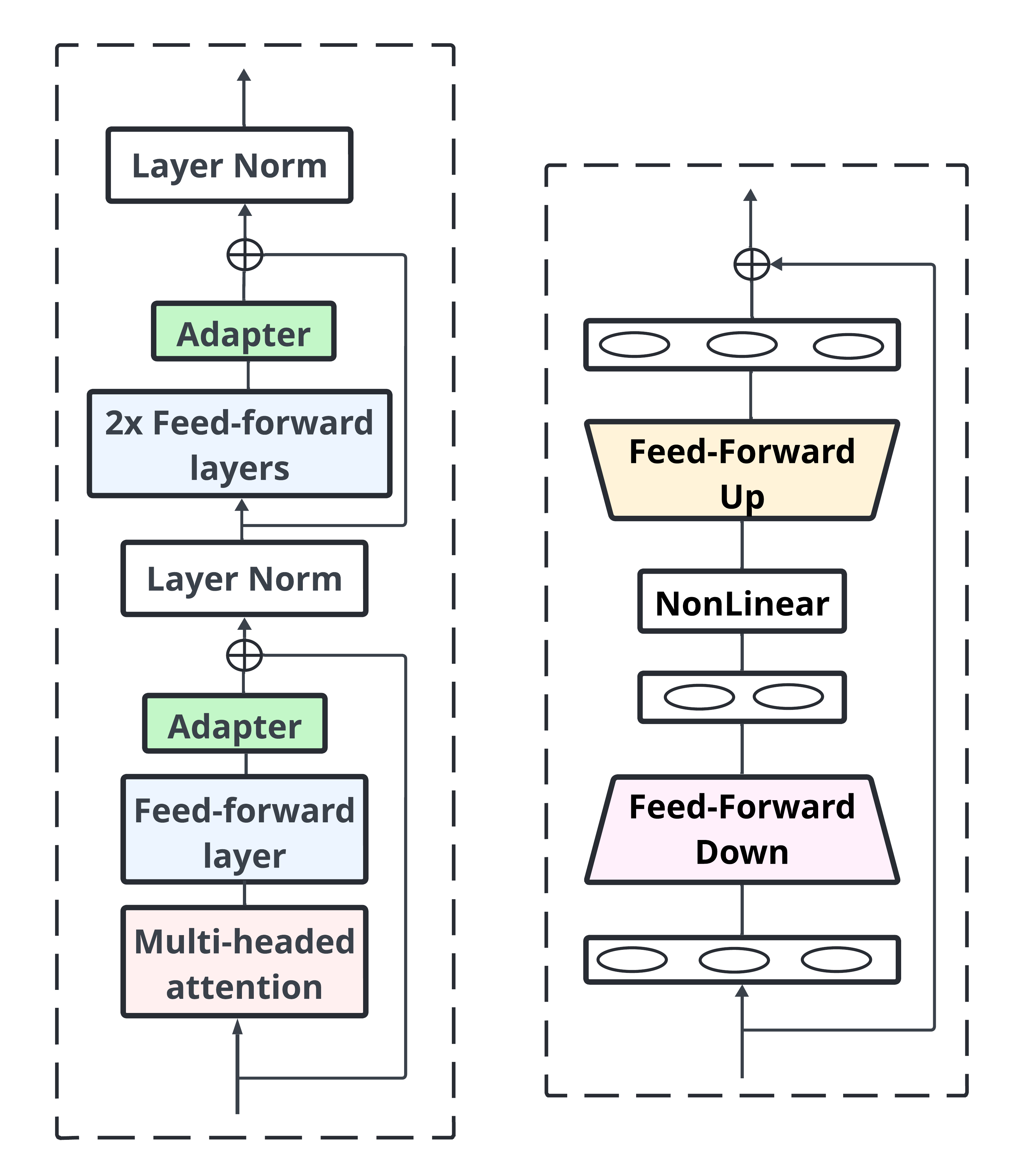}
\caption{Schematic representation of the Adapter Architecture used in LLMs. The diagram showcases the integration of adapters within the Transformer architecture, including the feed-forward up and down layers and their role in enabling efficient model adaptation by inserting additional parameters while maintaining the model's core structure (adapted from \cite{adapterArchitecture})}
\label{adapterArchitecture}
\end{figure}

\noindent HuggingFace supports adapter configurations through the PEFT library. During fine-tuning, new adapters are integrated into the model using LoraConfig \footnote{\url{https://huggingface.co/docs/peft/en/package_reference/lora}}. HuggingFace uses PeftConfig to load existing pre-trained models and apply PEFT techniques. Additionally, HuggingFace provides built-in support to run the fine-tuning process across any distributed configuration using Accelerate\footnote{\url{https://huggingface.co/docs/accelerate/en/index}}, making large-scale training and inference simple, efficient, and adaptable.

\subsection{Low-Rank Adaptation (LoRA)}
Low-Rank Adaptation (LoRA)\cite{lora} is a technique designed for fine-tuning large language models, which modifies the fine-tuning process by freezing the original model weights and applying changes to a separate set of weights, added to the original parameters. LoRA transforms the model parameters into a lower-rank dimension, reducing the number of trainable parameters, speeding up the process, and lowering costs. This method is particularly useful in scenarios where multiple clients require fine-tuned models for different applications, allowing for the creation of specific weights for each use case without the need for separate models. By employing low-rank approximation methods, LoRA effectively reduces computational and resource requirements while preserving the pre-trained model's adaptability to specific tasks or domains.

\begin{figure}[h]
\centering
\includegraphics[width=0.75\textwidth]{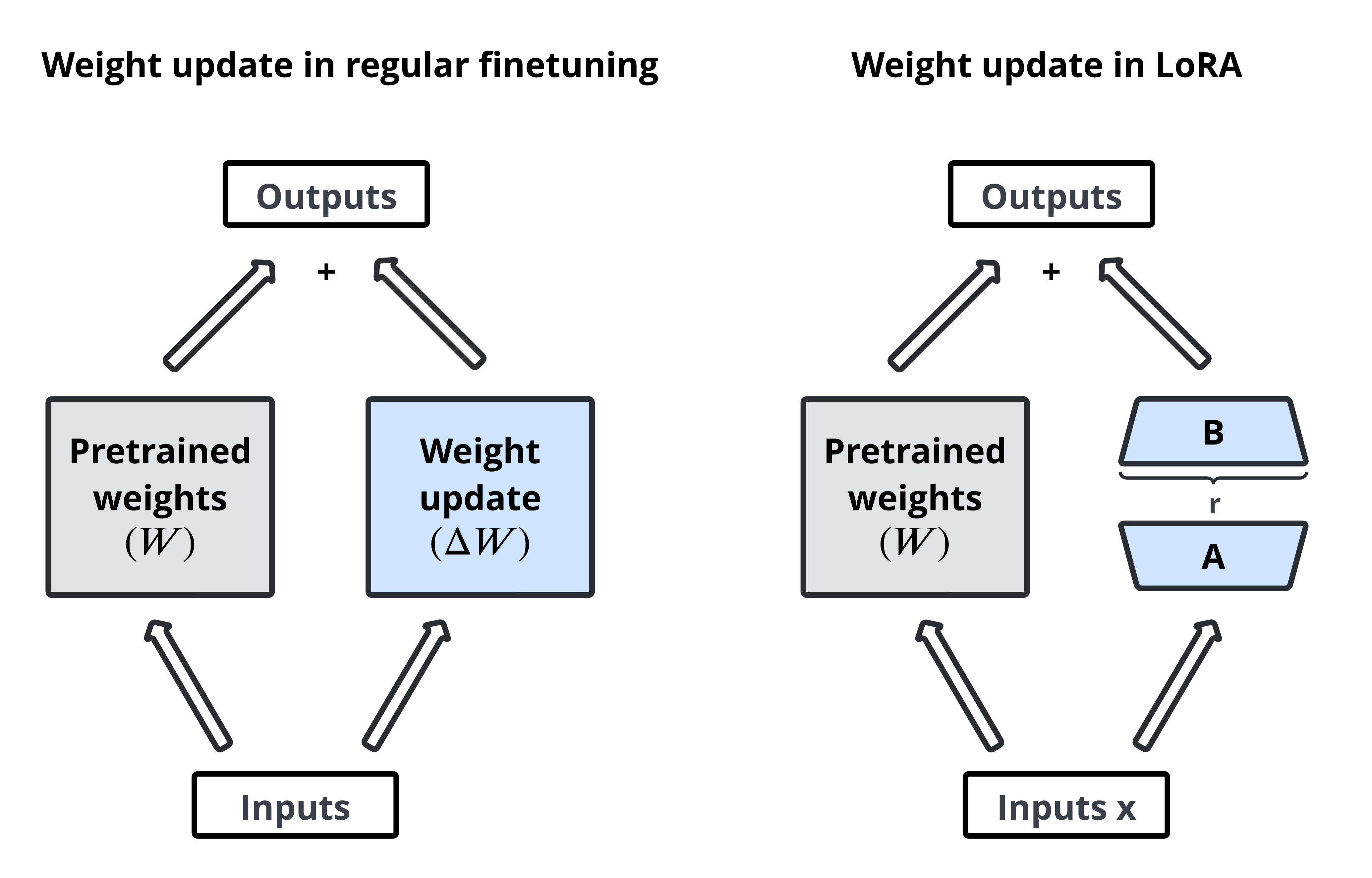}
\caption{A comparison between weight updates in regular fine-tuning and LoRA fine-tuning. In regular fine-tuning, the entire weight update matrix ($\Delta W$) is applied to the pre-trained weights. In contrast, LoRA fine-tuning introduces two low-rank matrices (A and B) that approximate the weight update matrix ($\Delta W$), significantly reducing the number of trainable parameters by leveraging the inner dimension (r), which is a hyperparameter. This method is more efficient in terms of memory and computation, making it ideal for fine-tuning large models. (adapted from \cite{regularFTvsLora})}
\label{regularFineTuningVsLora}
\end{figure}

\subsubsection{Benefits of Using LoRA}
\begin{enumerate}
\item \textbf{Parameter Efficiency:} LoRA significantly reduces the number of parameters that need to be trained by focusing only on the low-rank matrices, resulting in lower memory and storage requirements compared to full fine-tuning.
\item \textbf{Efficient Storage:} The storage of the trained model is more efficient as it only requires storing the low-rank matrices instead of the full model weights.
\item \textbf{Reduced Computational Load:} Training with low-rank matrices requires fewer computational resources, making it faster and more scalable.
\item \textbf{Lower Memory Footprint:} Since fewer parameters are being updated, the memory footprint during training is reduced, enabling the use of larger batch sizes or more complex models within the same hardware constraints.
\item \textbf{Flexibility:} LoRA can be easily integrated with existing pre-trained models without extensive modifications to the model architecture.
\item \textbf{Compatibility:} It can be used alongside other fine-tuning techniques, such as adapter layers or prompt-tuning, to further enhance performance.
\item \textbf{Comparable Results:} Despite the reduction in the number of trainable parameters, LoRA has been shown to achieve performance comparable to full fine-tuning in many tasks.
\item \textbf{Task-Specific Adaptation:} It effectively adapts the pre-trained model to specific tasks, leveraging the knowledge already embedded in the original model.
\item \textbf{Avoiding Overfitting:} By focusing on low-rank updates, LoRA can help in mitigating overfitting, especially when dealing with smaller task-specific datasets.
\end{enumerate}

\subsubsection{Limitations}
While LoRA demonstrates considerable power, it also presents challenges:
\begin{itemize}
\item \textbf{Fine-tuning Scope:} LoRA may face difficulties when applied to tasks demanding substantial alterations to the pre-trained model’s internal representations.
\item \textbf{Hyperparameter Optimisation:} Tuning the rank parameter ‘r’ requires meticulous adjustment for optimal performance.
\item \textbf{Ongoing Research:} Despite its promise, LoRA is still in active research stages, and its long-term implications remain to be fully explored.
\end{itemize}

\noindent Despite these challenges, LoRA stands as a pioneering technique with vast potential to democratise access to the capabilities of LLMs. Continued research and development offer the prospect of overcoming current limitations and unlocking even greater efficiency and adaptability.

\subsubsection{Tutorial for Fine-Tuning LLM Using LoRA}

An open-source template for fine-tuning LLMs using the LoRA method with the Hugging Face library can be found \href{https://gitlab.com/CeADARIreland_Public/llm-resources}{\color{blue} here}. This template is designed specifically for adapting LLMs for instruction fine-tuning processes.

\subsection{QLoRA}

\begin{figure}[h]
\centering
\includegraphics[width=0.8\textwidth]{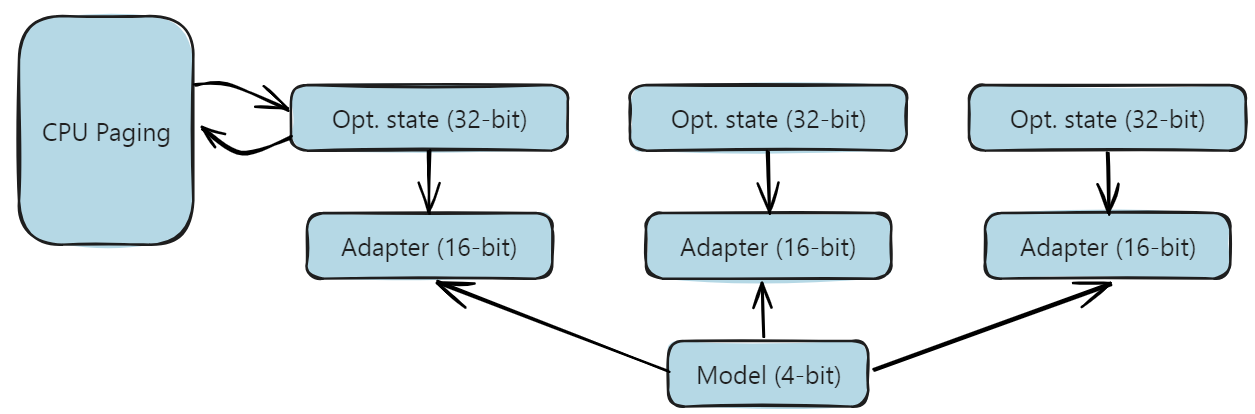}
\caption{Quantised Low-Rank Adaptation (QLoRA) Optimisation Workflow. This figure illustrates the QLoRA optimisation process, showing how the optimisation states, adapters, and the model interact during fine-tuning. It demonstrates the use of different bit-widths (32-bit, 16-bit, and 4-bit) to optimise the memory and computational efficiency during the fine-tuning of large language models (adapted from \cite{qloraArchitecture}).}
\label{qlora}
\end{figure}

QLoRA\cite{qlora} is an extended version of LoRA designed for greater memory efficiency in large language models (LLMs) by quantising weight parameters to 4-bit precision. Typically, LLM parameters are stored in a 32-bit format, but QLoRA compresses them to 4-bit, significantly reducing the memory footprint. This allows fine-tuning on less powerful hardware, including consumer GPUs. QLoRA also quantises the weights of the LoRA adapters from 8-bit to 4-bit, further decreasing memory and storage requirements (see Figure~\ref{qlora}). Despite the reduction in bit precision, QLoRA maintains performance levels comparable to traditional 16-bit fine-tuning. \\

\noindent It achieves this by backpropagating gradients through a frozen, 4-bit quantised pre-trained language model into Low-Rank Adapters, making the fine-tuning process efficient while preserving model effectiveness. The QLoRA configuration is supported by HuggingFace via the PEFT library, utilising LoraConfig and BitsAndBytesConfig for quantisation. Innovations such as an optimal 4-bit data type, double quantisation of constants, and memory spike management enable QLoRA to reduce memory usage from 96 bits per parameter in traditional fine-tuning to 5.2 bits per parameter, an 18-fold reduction.

\noindent Performance-wise, QLoRA outperforms naive 4-bit quantisation and matches 16-bit quantised models on benchmarks. Additionally, QLoRA enabled the fine-tuning of a high-quality 4-bit chatbot using a single GPU in 24 hours, achieving quality comparable to ChatGPT. \\

\noindent This \href{https://dassum.medium.com/fine-tune-large-language-model-llm-on-a-custom-dataset-with-qlora-fb60abdeba07}{\color{blue} tutorial} explains the end-to-end steps of fine-tuning QLoRA on a custom dataset for the Phi-2 model.

\subsection{Weight-Decomposed Low-Rank Adaptation (DoRA)}
In the context of optimising model fine-tuning, the pattern analysis of LoRA and Full Fine-Tuning (FT) reveals significant differences in learning behaviours and updates. LoRA, employing a strategy of incrementally updating pre-trained weights using the product of two low-rank matrices, maintains the original weights largely static during the fine-tuning process, which allows for efficient inference. Despite its computational efficiency, previous studies have suggested that LoRA's limited number of trainable parameters might contribute to its performance discrepancies when compared to FT.

\noindent Weight-Decomposed Low-Rank Adaptation (DoRA) \cite{doraPaper} is a novel fine-tuning methodology designed to optimise pre-trained models by decomposing their weights into magnitude and directional components. This approach leverages the efficiency of Low-Rank Adaptation (LoRA) for directional updates, facilitating substantial parameter updates without altering the entire model architecture. DoRA addresses the computational challenges associated with traditional full fine-tuning (FT) by maintaining model simplicity and inference efficiency, while simultaneously bridging the performance gap typically observed between LoRA and FT. Empirical and theoretical evaluations demonstrate that DoRA not only achieves learning outcomes comparable to FT across diverse tasks—including natural language processing and vision-language applications—but also consistently surpasses LoRA in performance, providing a robust solution for enhancing the adaptability and efficiency of large-scale models. \\

\begin{figure}[h]
\centering
\includegraphics[width=0.8\textwidth]{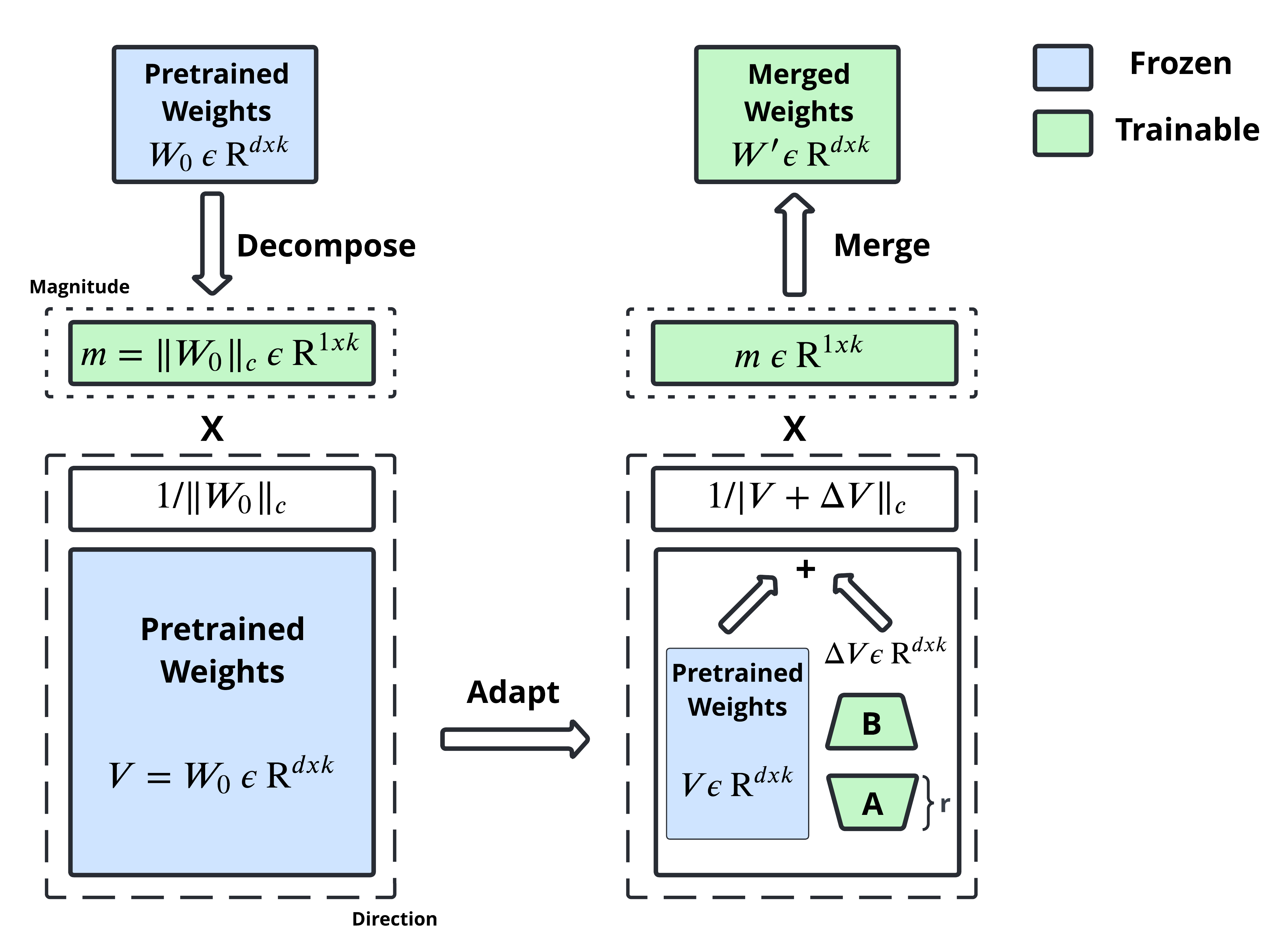}
\caption{An overview of DoRA (Decomposed Representations for Adaptation), which is a method for weight decomposed low-rank adaptation. The figure illustrates how pre-trained weights are decomposed and adapted for fine-tuning. In the left section, pre-trained weights are decomposed into a magnitude and direction. The right section shows how these decomposed weights are merged with trainable parameters during fine-tuning, resulting in updated weights that combine both frozen (blue) and trainable (green) components. The process emphasises efficient adaptation by focusing on the most significant directions in the parameter space, facilitating effective fine-tuning while maintaining the integrity of the original model (adapted from \cite{doraPaper}).}
\label{dora}
\end{figure}

\noindent \textbf{Python Library -} DoRA is facilitated via the HuggingFace LoraConfig package. To incorporate DoRA into the fine-tuning process, it is essential to specify the 'use\_dora = True' parameter during the Lora configuration. Further information on initialisation can be found \href{https://huggingface.co/docs/peft/v0.8.2/en/package_reference/lora}{here}.

\subsubsection{Benefits of DoRA}
\begin{enumerate}
    \item \textbf{Enhanced Learning Capacity:} DoRA achieves a learning capacity closely resembling full fine-tuning (FT) by decomposing pre-trained weights into magnitude and directional components, allowing for more nuanced updates.

    \item \textbf{Efficient Fine-Tuning:} By utilising the structural advantages of Low-Rank Adaptation (LoRA) for directional updates, DoRA enables efficient fine-tuning without altering the entire model architecture.

    \item \textbf{No Additional Inference Latency:} Despite its improved learning capabilities, DoRA does not introduce any additional inference latency over LoRA, maintaining model simplicity and efficiency.

    \item \textbf{Superior Performance:} Experimental results demonstrate that DoRA consistently outperforms LoRA across a wide range of tasks, including natural language processing (NLP), visual instruction tuning, and image/video-text understanding. For example, it shows significant improvements in commonsense reasoning and visual instruction tuning benchmarks.

    \item \textbf{Versatility Across Backbones:} DoRA has been validated across various model backbones, including large language models (LLM) and vision-language models (LVLM), indicating its broad applicability and robustness in different domains.

    \item \textbf{Innovative Analysis:} The introduction of a novel weight decomposition analysis helps uncover fundamental differences in the learning patterns of FT and various parameter-efficient fine-tuning (PEFT) methods, contributing to a deeper understanding of model fine-tuning dynamics.
\end{enumerate}

\subsubsection{Comparison between LoRA and DoRA}

Low-Rank Adaptation (LoRA) and Weight-Decomposed Low-Rank Adaptation (DoRA) are both advanced techniques designed to improve the efficiency and effectiveness of fine-tuning large pre-trained models. While they share the common goal of reducing computational overhead, they employ different strategies to achieve this (see Table\ref{tab:loraVsDora}).

\begin{table}[H]
\centering
\begin{tabularx}{\textwidth}{|>{\raggedright\arraybackslash}p{4cm}|X|X|}
\hline
\textbf{Criteria} & \textbf{LoRA (Low-Rank Adaptation)} & \textbf{DoRA (Weight-Decomposed Low-Rank Adaptation)} \\ \hline
\textbf{Objective} & Provide an efficient method for fine-tuning pre-trained models by using low-rank matrix products to update weights incrementally without increasing inference latency. & Improves learning capacity by closely mimicking the learning patterns of full fine-tuning, optimising magnitude and direction separately. \\ \hline
\textbf{Approach} & Implements a low-rank decomposition where the weight update is modelled as the product of two low-rank matrices (B and A), keeping the original weights static. & Uses weight decomposition analysis to reparameterise the weight matrix into separate magnitude and direction components for distinct updates. \\ \hline
\textbf{Model Architecture} & Keeps the pre-trained weight matrix (W0) unchanged and applies updates using low-rank matrices (B and A). Matrix A is initialised with a uniform Kaiming distribution, while B is set to zero initially. & Restructures the weight matrix into magnitude and directional components, ensuring directional vectors are unit vectors for more detailed adjustments. \\ \hline
\end{tabularx}
\caption{A detailed comparison between LoRA (Low-Rank Adaptation) and DoRA (Weight-Decomposed Low-Rank Adaptation), highlighting their objectives, approaches, and the specific architectural strategies they employ for fine-tuning large language models.}
\label{tab:loraVsDora}
\end{table}

\subsubsection{Tutorial for Fine-Tuning LLM using DoRA}
\noindent This \href{https://www.kaggle.com/code/aisuko/dora-from-scratch}{tutorial} offers an in-depth guide and detailed explanation of the steps involved in implementing DoRA from scratch, as well as insights into the fine-tuning process essential for optimising performance.

\subsection{Fine-Tuning with Multiple Adapters}
\begin{figure}[h]
\centering
\includegraphics[width=1\textwidth]{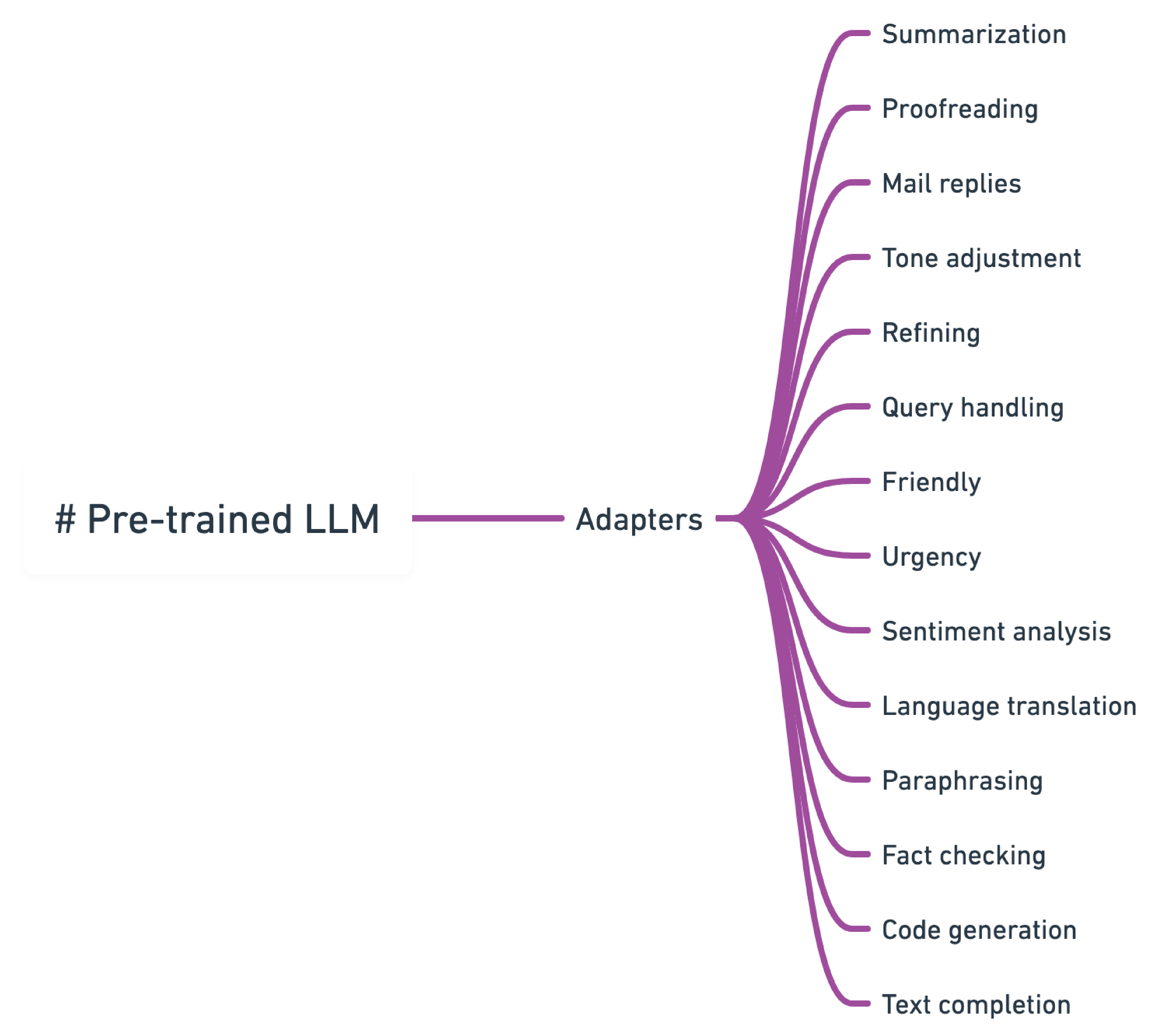}
\caption{Overview of how multiple adapters can be used with a pre-trained LLM to fine-tune it for various specific tasks, such as summarisation, proofreading, sentiment analysis, and more. (adapted from \cite{fineTuningWithMultipleAdapters})}
\label{fineTuningWithMultipleAdapters}
\end{figure}

During fine-tuning, we have explored the method of freezing the parameters of the LLM and focusing solely on fine-tuning a few million trainable parameters using LoRA. For example, fine-tuning an LLM for translation involves training a translation adapter with relevant data. This approach allows us to fine-tune separate adapters for each specific task we want the LLM to perform. However, a key question arises: can we consolidate multiple adapters into a unified multi-task adapter? For instance, if we have separate adapters for translation and summarisation tasks, can we merge them so that the LLM can proficiently handle both tasks? (Illustrated via Figure\ref{fineTuningWithMultipleAdapters}).\\

\noindent The PEFT library simplifies the process of merging adapters with its add\_weighted\_adapter function \footnote{\url{https://huggingface.co/docs/peft/main/en/package_reference/lora\#peft.LoraModel.add_weighted_adapter}}, which offers three distinct methods:

\begin{enumerate}
\item \textbf{Concatenation:} This straightforward method concatenates the parameters of the adapters. For instance, if two adapters each have a rank of 16, the resulting adapter will have a rank of 32. This method is highly efficient.
\item \textbf{Linear Combination:} Although less documented, this method appears to perform a weighted sum of the adapters’ parameters.
\item \textbf{SVD:} The default method employs singular value decomposition through torch.linalg.svd. While versatile, it is notably slower than the other methods, particularly for adapters with high ranks (greater than 100), which can take several hours.
\end{enumerate}

\noindent Each method allows for customising the combination by adjusting weights. For instance, when merging two adapters, X and Y, assigning more weight to X ensures that the resulting adapter prioritises behaviour similar to X over Y.

\noindent This approach is particularly suited for consolidating a single LLM to handle multiple tasks rather than creating separate models for each task domain. By adopting this method, there is no longer a need to individually fine-tune a model for each task. Instead, a single adapter layer can be fine-tuned for each task, allowing queries to yield the desired responses efficiently.

\subsubsection{Steps for Fine-Tuning LLM with LoRA for Multiple Tasks and Adapters}

\begin{enumerate}
    \item \textbf{Adapter Creation:} Create multiple adapters, each fine-tuned for specific tasks using different prompt formats or task-identifying tags (e.g., [translate\_fren], [chat]).
    \item \textbf{LoRA Integration:} Implement LoRA to efficiently integrate these adapters into the pre-trained LLM. Utilise LoRA's methods such as concatenation, linear combination, or singular value decomposition (SVD) to combine adapters while minimising computational overhead and maintaining performance.
    \item \textbf{Task-Specific Adaptation:} Fine-tune each adapter with task-specific data to enhance performance for individual tasks. Ensure adapters are trained with data relevant to their respective tasks, optimising their ability to generate accurate responses.
    \item \textbf{Behaviour Adjustment:} Monitor the behaviour of combined adapters to identify any undesired inherited behaviours from individual adapters (e.g., short response generation from a translation adapter). Adjust the combination weights or types to modify adapter behaviour as needed, ensuring each adapter performs optimally for its intended task.
    \item \textbf{Evaluation and Iteration:} Evaluate the performance of the combined model across multiple tasks using validation datasets. Iterate on the fine-tuning process, making adjustments to adapter combinations and training parameters based on performance metrics and user feedback.
\end{enumerate}

\noindent Therefore, for optimal performance, it is advisable to combine adapters that have been fine-tuned with distinctly varied prompt formats. However, even when using adapters with different prompt formats, the resulting adapter may not exhibit desired behaviour. For example, a newly combined adapter designed for chatting may only generate short responses, inheriting this tendency from an adapter that was originally trained to halt after producing a single sentence. To adjust the behaviour of the combined adapter, one can prioritise the influence of a specific adapter during the combination process and/or modify the method of combination used.

\noindent An illustrative tutorial demonstrating the fine-tuning of large language models (LLMs) using multiple adapter layers for various tasks can be found \href{https://kaitchup.substack.com/p/combine-multiple-lora-adapters-for}{\color{blue} here}.

\section{Half Fine Tuning}
Half Fine-Tuning (HFT)\cite{halfFineTuningPaper} is a technique designed to balance the retention of foundational knowledge with the acquisition of new skills in large language models (LLMs). HFT involves freezing half of the model's parameters during each fine-tuning round while updating the other half, allowing the model to retain pre-trained knowledge and enhance new task performance without altering the model architecture. 

\noindent Each repetitive transformer layer is divided into three blocks: self-attention, feed-forward, and layernorm, with half of the parameters in each block updated and the other half frozen, varying with each round. This strategic parameter update helps maintain knowledge parity across training rounds and enhances scalability in successive training sessions. 

\noindent Research on models like LLAMA 2-7B demonstrated that HFT could significantly restore forgotten basic knowledge while preserving high general ability performance. This method's robustness and efficiency make it applicable to various fine-tuning scenarios, including supervised fine-tuning, direct preference optimisation, and continual learning. Additionally, HFT's ability to maintain the model architecture simplifies its implementation and ensures compatibility with existing systems, further promoting its practical adoption.

\begin{figure}[h]
\centering
\includegraphics[width=0.8\textwidth]{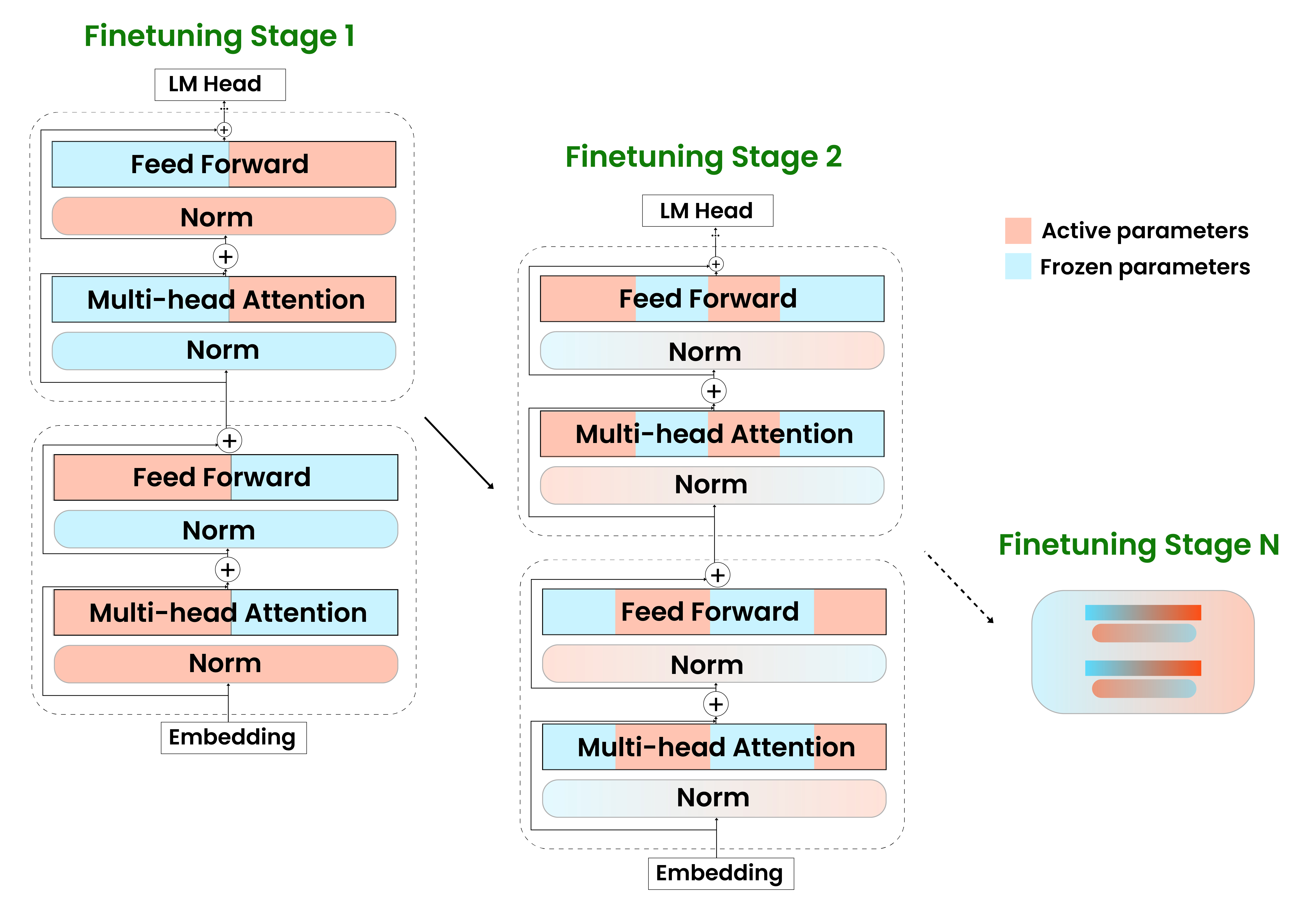}
\caption{Schematic illustration of the Half Fine-Tuning (HFT) method as applied to LLAMA 2’s architecture. The diagram shows multiple stages of fine-tuning, where specific model parameters are selectively activated (orange) while others remain frozen (blue). This approach optimises training by reducing computational requirements while still effectively adapting the model to new tasks or data. (adapted from \cite{halfFineTuningPaper})}
\label{halfFineTuning}
\end{figure}

\subsection{Benefits of using Half Fine tuning}
\begin{enumerate}
    \item \textbf{Recovery of Pre-Trained Knowledge:} By rolling back half of the fine-tuned parameters to their pre-trained state, HFT effectively recovers a portion of the original knowledge, thereby mitigating catastrophic forgetting of previously acquired capabilities.
    
    \item \textbf{Enhanced Performance:} Research experiments shows that HFT maintains or even surpasses the performance of full fine-tuning (FFT) on downstream tasks, demonstrating its effectiveness in balancing knowledge retention with task-specific learning.

    \item \textbf{Robustness:} The method is robust to different selection strategies and the number of parameters chosen for updating, ensuring consistent performance across various configurations.

    \item \textbf{Simplicity and Scalability:} HFT does not alter the model architecture, which simplifies implementation and allows for scalable applications, particularly beneficial in successive fine-tuning scenarios.
    
    \item \textbf{Versatility:} The technique has proven effective across diverse fine-tuning scenarios, including supervised fine-tuning, direct preference optimisation, and continual learning.
\end{enumerate}

\subsection{Comparison between HFT and LoRA}
\begin{table}[H]
\centering
\begin{tabularx}{\textwidth}{|>{\raggedright\arraybackslash}p{4cm}|X|X|}
\hline
\textbf{Criteria} & \textbf{HFT} & \textbf{LoRA} \\ \hline
\textbf{Objective} & The goal is to retain the foundational knowledge acquired during pre-training while learning new task-specific skills, thus balancing between maintaining existing capabilities and acquiring new ones. & LoRA aims to reduce computational and memory requirements during fine-tuning, making it more efficient and feasible to train large models on limited hardware resources. \\ \hline
\textbf{Approach} & HFT involves freezing half of the model's parameters during each fine-tuning round and updating only the other half. & LoRA reduces the number of trainable parameters by introducing low-rank decomposition into the weight matrices of the neural network. This involves injecting low-rank matrices into the model’s layers during fine-tuning. \\ \hline
\textbf{Model Architecture} & HFT does not alter the model's architecture or introduce new parameters, making it straightforward to apply without additional structural changes. & LoRA modifies the model by adding low-rank matrices, which changes the training dynamics and requires additional computations for the low-rank updates. \\ \hline
\textbf{Performance} & Research has shown that HFT can restore forgotten basic knowledge while maintaining high performance in general abilities. & LoRA is designed to achieve competitive performance with full fine-tuning but with significantly fewer trainable parameters and lower computational costs. \\ \hline
\end{tabularx}
\caption{Comparative Analysis of Half Fine-Tuning (HFT) and Low-Rank Adaptation (LoRA).}
\label{tab:hftVsLora}
\end{table}

\section{Lamini Memory Tuning}

Lamini \cite{laminiPaper} was introduced as a specialised approach to fine-tuning Large Language Models (LLMs), targeting the reduction of hallucinations. This development was motivated by the need to enhance the reliability and precision of LLMs in domains requiring accurate information retrieval. Traditional training methods typically consist of running stochastic gradient descent on vast datasets, which, despite fitting the training data well, often produce models that fail to generalise effectively and are prone to such errors. \\

\noindent Foundation models often follow a training regimen similar to the Chinchilla recipe, which prescribes training for a single epoch on a massive corpus, such as training Llama 2 7B on about one trillion tokens. This approach results in substantial loss and is geared more towards enhancing generalisation and creativity where a degree of randomness in token selection is permissible. However, it falls short for tasks demanding high factual precision. In contrast, Lamini Memory Tuning delves deeper by analysing the loss of individual facts, significantly improving the accuracy of factual recall. By augmenting a model with additional parameters specifically for memory (e.g., an 8B parameter model with an extra 2B parameters for weights), Lamini enables the model to memorise and accurately recall a significant number of facts, closely aligning performance with LLM scaling laws without compromising on generalisation.

\subsection{Lamini-1 - A model architecture based on Lamini}

\begin{figure}[h]
\centering
\includegraphics[width=0.85\textwidth]{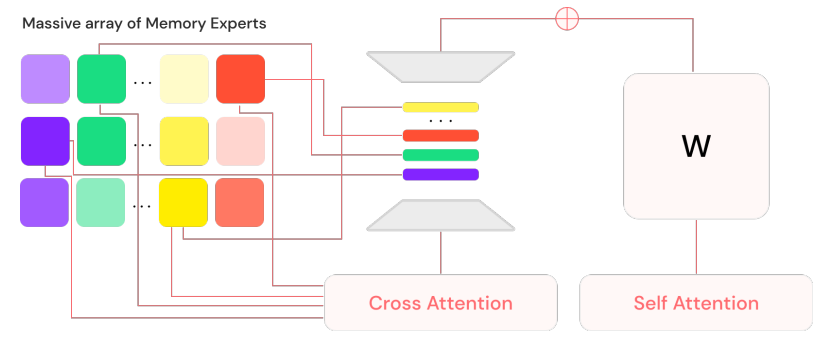}
\caption{Diagram of the Lamini-1 Model Architecture, featuring a Massive Array of Memory Experts (MoME). This architecture integrates a pre-trained transformer backbone with dynamically selected adapters via cross-attention mechanisms. Each adapter, functioning as a memory expert, is capable of storing specific factual data.  (adopted from \cite{laminiPaper})}
\label{laminiArchitecture}
\end{figure}

Departing from traditional transformer-based designs, the Lamini-1 model architecture (Figure \ref{laminiArchitecture}) employs a massive mixture of memory experts (MoME). This system features a pre-trained transformer backbone augmented by adapters that are dynamically selected from an index using cross-attention mechanisms. These adapters function similarly to experts in MoE architectures, and the network is trained end-to-end while freezing the backbone. This setup allows for specific facts to be stored exactly in the selected experts. \\

\noindent At inference time, only the relevant experts are retrieved from the index, enabling the LLM to store a large number of facts while maintaining low inference latency. Specialised GPU kernels written in Triton are used to accelerate the lookup of experts, optimising the system for quick access to stored knowledge.

\subsubsection{Systems Optimisations for Banishing Hallucinations}
The MoME architecture is designed to minimise the computational demand required to memorise facts. During training, a subset of experts, such as 32 out of a million, is selected for each fact. The weights of the backbone network and the cross attention used to select the expert are frozen, and gradient descent steps are taken until the loss is sufficiently reduced to memorise the fact. This approach prevents the same expert from being selected multiple times for different facts by first training the cross attention selection mechanism during a generalisation training phase, then freezing its weights. \\

\noindent This method ensures that computation scales with the number of training examples, not the total number of parameters, thereby significantly reducing the computation required for memory tuning. This optimised approach allows Lamini-1 to achieve near-zero loss in memory tuning on real and random answers efficiently, demonstrating its efficacy in eliminating hallucinations while improving factual recall. 

\section{Mixture of Experts}

A mixture of experts (MoE) is an architectural design for neural networks that divides the computation of a layer or operation (e.g., linear layers, MLPs, or attention projection) into several specialised subnetworks, referred to as "experts". Each expert independently carries out its computation, and the results are aggregated to produce the final output of the MoE layer. MoE architectures can be categorised as either dense, where every expert is engaged for each input, or sparse, where only a subset of experts is utilised for each input.

\subsection{Mixtral 8x7B Architecture and Performance}

\begin{figure}[h]
\centering
\includegraphics[width=0.75\textwidth]{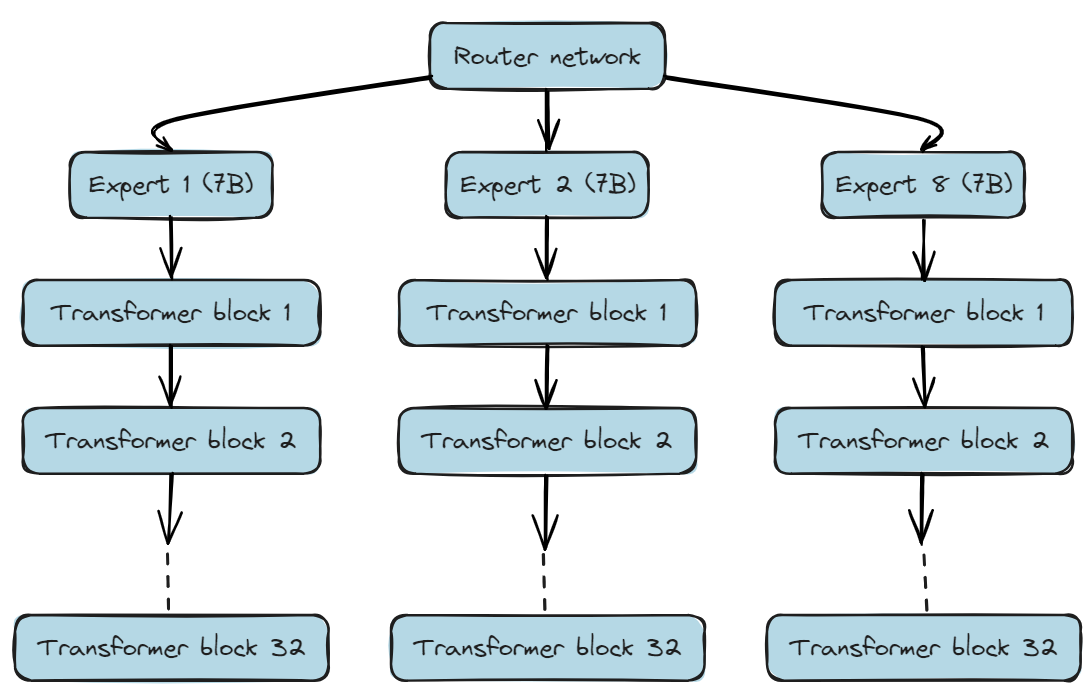}
\caption{Diagram of the Mixtral 8x7B Mixture of Experts (MoE) model architecture. The model is composed of a router network that dynamically selects the most relevant experts from a pool of eight transformer-based experts, each with 7 billion parameters. The experts are organised into transformer blocks, where the router directs data to the appropriate expert based on the input, optimising computational efficiency and model performance. This architecture allows for scalability and specialised processing within large language models. (adapted from \cite{mixtralMoeModel})}
\label{moeModel}
\end{figure}

\noindent Mixtral \cite{mixtralOfExperts} 8x7B employs a Sparse Mixture of Experts (SMoE) architecture (Figure \ref{moeModel}), mirroring the structure of Mistral 7B but incorporating eight feedforward blocks (experts) in each layer. For every token at each layer, a router network selects two experts to process the current state and combine their outputs. Although each token interacts with only two experts at a time, the selected experts can vary at each timestep. Consequently, each token has access to 47 billion parameters but utilises only 13 billion active parameters during inference. Mixtral 8x7B not only matches but often surpasses Llama 2 70B and GPT-3.5 across all evaluated benchmarks. Its performance is notably superior to Llama 2 70B in mathematics, code generation, and multilingual tasks.

\section{Mixture of Agents}

Despite the numerous LLMs and their notable accomplishments, they continue to encounter fundamental limitations regarding model size and training data. Scaling these models further is prohibitively expensive, often necessitating extensive retraining on multiple trillion tokens. Simultaneously, different LLMs exhibit distinct strengths and specialise in various aspects of tasks. A recent study has investigated leveraging the collective expertise of multiple LLMs to develop a more capable and robust model, a method known as Mixture of Agents (MoA) \cite{mixtureOfAgentsPaper}.

\begin{figure}[h]
\centering
\includegraphics[width=1\textwidth]{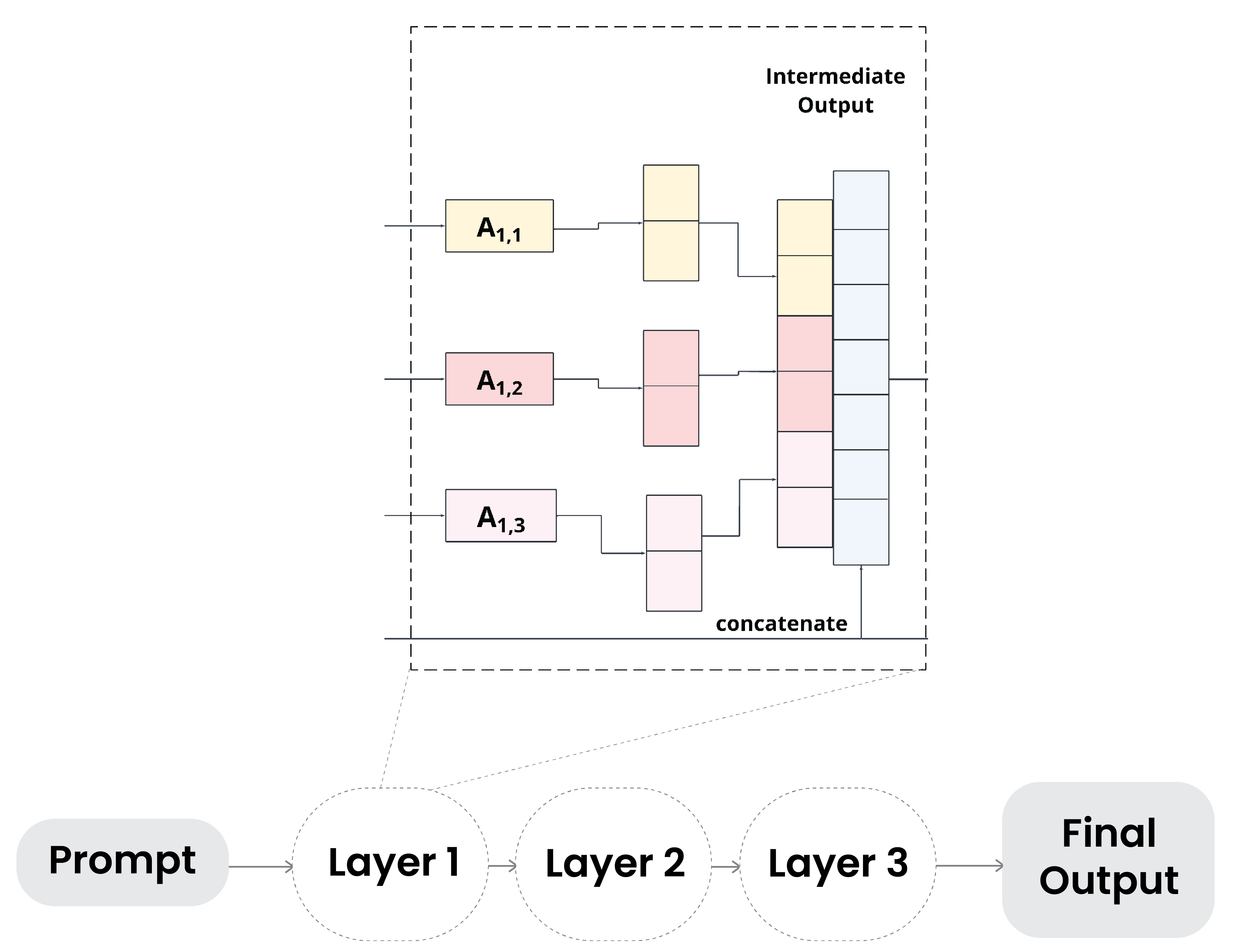}
\caption{Illustration for Mixture of Agents (MoA) LLM configuration. The model consists of multiple layers, each incorporating several agents that process the input independently before concatenating their outputs to form an intermediate result. The process continues across layers, refining the output at each stage to generate the final output based on the given prompt (adapted from \cite{mixtureOfAgentsPaper}).}
\label{moaModel}
\end{figure}

\noindent MoA functions using a layered architecture, where each layer comprises multiple LLM agents (Figure ~\ref{moaModel}). This structure reveals a phenomenon known as the “collaborativeness of LLMs.” The innovative MoA framework utilises the combined capabilities of several LLMs to enhance both reasoning and language generation proficiency. Research indicates that LLMs naturally collaborate, demonstrating improved response quality when incorporating outputs from other models, even if those outputs are not ideal.

\subsection{Methodology}
To enhance collaboration among multiple LLMs, it is essential to understand their individual strengths and classify them accordingly. The classification includes:

\begin{enumerate}
    \item \textbf{Proposers:} These models excel at generating valuable reference responses for other models. While they may not perform exceptionally on their own, they provide useful context and varied perspectives that improve the final output when utilised by an aggregator.

    \item \textbf{Aggregators:} These models are adept at merging responses from various models into a single high-quality result. An effective aggregator should maintain or even enhance the quality of the final response, regardless of the quality of the individual inputs.
\end{enumerate}

\noindent The careful selection of LLMs for each MoA layer is crucial
Performance metrics, such as average win rates in a given layer, help assess the suitability of models for subsequent layers, ensuring the production of higher-quality outputs.
Diversity in model outputs is vital, as varied responses from different models contribute significantly more than homogeneous outputs from a single model.
In MoA, given an input prompt, the output of the $i^\text{th}$ MoA layer $y_i$ is calculated as follows:

\begin{equation}
y_i = \bigoplus_{j=1}^n \left[ A_{i,j}(x_i) \right] + x_1, \, x_{i+1} = y_i 
\end{equation}

\subsection{Analogy with MoE}
Mixture-of-Experts (MoE) is a well-established machine learning technique where multiple expert networks, each with specialised skills, collaborate to address complex problems. This approach has demonstrated significant success across various applications and serves as the inspiration for the Mixture-of-Agents (MoA) method. In a typical MoE design, a stack of layers, known as MoE layers, consists of multiple expert networks, a gating network, and residual connections to improve gradient flow. The output for layer $y_i$ is calculated as follows:

\begin{equation}
y_i = \sum_{j=1}^n G_{i,j}(x_i) E_{i,j}(x_i) + x_i    
\end{equation}

\noindent The MoA framework advances the MoE concept by operating at the model level through prompt-based interactions rather than altering internal activations or weights. Instead of relying on specialised sub-networks within a single model, MoA utilises multiple full-fledged LLMs across different layers. In this approach, the gating and expert networks' functions are integrated within an LLM, leveraging its ability to interpret prompts and generate coherent outputs without additional coordination mechanisms.

\subsection{What makes MoA works well?}
\begin{enumerate}
    \item \textbf{MoA's Superior Performance:} MoA significantly outperforms LLM-based rankers, which select one answer from the proposals rather than generating new responses. This suggests that MoA’s approach of aggregating all generated responses provides more effective results than simply choosing from pre-existing options.

    \item \textbf{Effective Incorporation of Proposals:} The aggregator in MoA demonstrates a tendency to integrate the best proposed answers. This is supported by positive correlations between aggregator responses and various similarity metrics, such as BLEU scores, which measure n-gram overlaps. The use of alternative similarity measures also shows a consistent positive correlation with preference scores, indicating that the aggregator effectively utilises the proposed responses.

    \item \textbf{Influence of Model Diversity and Proposer Count:} Increasing the number of proposers improves output quality, highlighting the benefits of additional auxiliary information. Additionally, using a diverse set of LLMs as proposers consistently yields better results compared to using a single LLM. This suggests that both the number and diversity of LLM agents in each MoA layer contribute to enhanced performance, with potential for further improvement through scaling.

    \item \textbf{Model Specialisation:} Analysis of model roles within the MoA ecosystem reveals that GPT-4o, Qwen, and LLaMA-3 are effective in both assisting and aggregating tasks. In contrast, WizardLM excels as a proposer but struggles with aggregating responses from other models.
\end{enumerate}

\section{Proximal Policy Optimisation (PPO)}
PPO \cite{ppoPaper} is a widely recognised reinforcement learning algorithm used for training agents to perform tasks in diverse environments. This algorithm leverages policy gradient methods, where policies—represented by neural networks—determine the actions taken by the agent based on the current state. PPO effectively handles the dynamic nature of training data generated through continuous agent-environment interactions, a feature that differentiates it from static datasets used in supervised learning.

\noindent The innovation of PPO lies in its "surrogate" objective function, optimised via stochastic gradient ascent. This approach allows for multiple updates from the same batch of data, enhancing both training efficiency and stability over traditional policy gradient methods. Developed by OpenAI, PPO was designed to balance ease of implementation with the robust performance characteristics of more complex algorithms like Trust Region Policy Optimisation (TRPO), but without the associated computational complexity.

\noindent PPO operates by maximising expected cumulative rewards through iterative policy adjustments that increase the likelihood of actions leading to higher rewards. A key feature of PPO is its use of a clipping mechanism in the objective function, which limits the extent of policy updates, thus preventing drastic changes and maintaining stability during training. \\

\begin{figure}[h]
\centering
\includegraphics[width=1\textwidth]{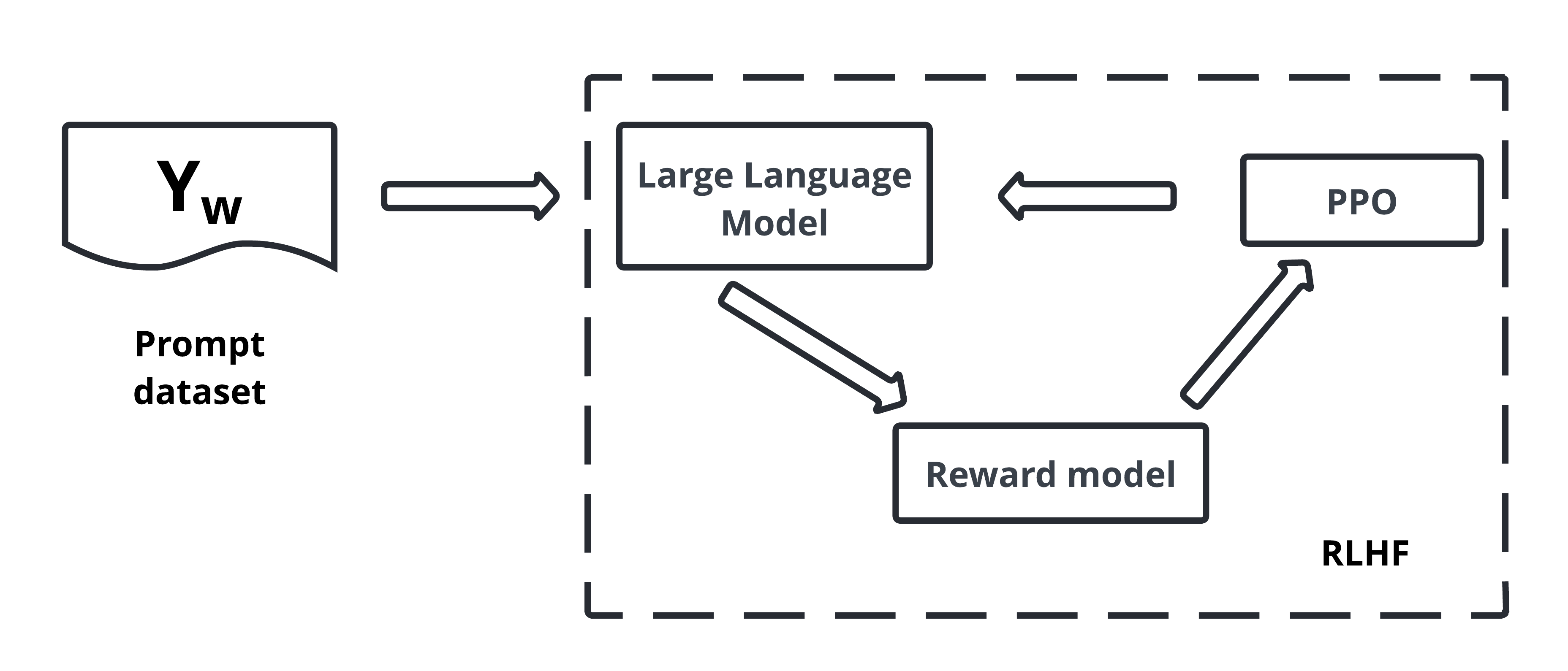}
\caption{Schematic of Proximal Policy Optimisation (PPO) applied in the context of Reinforcement Learning from Human Feedback (RLHF) for fine-tuning a Large Language Model (LLM). The process involves using a prompt dataset to train the LLM. The PPO algorithm adjusts the LLM’s policy based on rewards provided by the reward model, which is fine-tuned through human feedback. (adapted from \cite{ppoPaper})}
\label{ppo}
\end{figure}

\noindent \textbf{Python Library -} HuggingFace Transformer Reinforcement Learning (TRL\footnote{\url{https://huggingface.co/docs/trl/en/index}}) package supports the PPO Trainer\footnote{\url{https://huggingface.co/docs/trl/main/en/ppo_trainer}} for training language models from the preference data. \\

\noindent The PPOTrainer expects to align a generated response with a query given the rewards obtained from the Reward model. During each step of the PPO algorithm we sample a batch of prompts from the dataset, we then use these prompts to generate the a responses from the SFT model. Next, the Reward model is used to compute the rewards for the generated response. Finally, these rewards are used to optimise the SFT model using the PPO algorithm. Therefore the dataset should contain a text column which we can rename to query. Each of the other data-points required to optimise the SFT model are obtained during the training loop.

\subsection{Benefits of PPO}
\begin{enumerate}
    \item \textbf{Stability:} Proximal Policy Optimisation (PPO) is designed to ensure stable and reliable policy updates. The clipped surrogate objective function is central to this stability, as it limits policy updates to prevent large, potentially destabilising changes. This results in smoother and more consistent learning.

    \item \textbf{Ease of Implementation:} Compared to advanced algorithms TRPO, PPO is relatively straightforward to implement. It avoids the need for second-order optimisation techniques, making it more accessible to less experienced practitioners.

    \item \textbf{Sample Efficiency:} PPO achieves data efficiency through its use of the clipped surrogate objective. This mechanism regulates policy updates, ensuring stability while effectively reusing training data. Consequently, PPO tends to be more sample-efficient than other reinforcement learning algorithms, performing well with fewer samples, which is advantageous in scenarios where data collection is costly or time-consuming.
\end{enumerate}

\subsection{Limitations of PPO}

\begin{enumerate}
    \item \textbf{Complexity and Computational Cost:} Proximal Policy Optimisation (PPO) involves intricate policy and value networks, necessitating substantial computational resources for training. This complexity often results in extended training durations and increased operational expenses.

    \item \textbf{Hyperparameter Sensitivity:} PPO's performance is highly dependent on several hyperparameters, such as the clipping range, learning rate, and discount factor. Achieving optimal performance requires meticulous tuning of these parameters. Incorrect settings can lead to suboptimal policy outcomes or instability during the learning process.

    \item \textbf{Stability and Convergence Issues:} Although PPO is designed to enhance stability compared to earlier methods, it can still encounter convergence issues, particularly in highly dynamic or complex environments. Maintaining stable policy updates remains a significant challenge.

    \item \textbf{Reward Signal Dependence:} PPO's effectiveness is heavily reliant on a well-defined reward signal to guide the learning process. In scenarios where designing an appropriate reward function is challenging or impractical, PPO may struggle to attain the desired results.
\end{enumerate}

\subsection{Tutorial for training models using PPO technique}
The tutorial for tuning GPT2 to generate positive movie reviews based on the IMDB dataset using PPO technique can be found \href{https://github.com/huggingface/trl/blob/main/examples/notebooks/gpt2-sentiment.ipynb}{here.}

\section{Direct Preference Optimisation (DPO)}
Direct Preference Optimisation (DPO) \cite{dpoPaper} offers a streamlined approach to aligning language models (LMs) with human preferences, bypassing the complexity of reinforcement learning from human feedback (RLHF). Large-scale unsupervised LMs typically lack precise behavioural control, necessitating methods like RLHF that fine-tune models using human feedback. However, RLHF is intricate, involving the creation of reward models and the fine-tuning of LMs to maximise estimated rewards, which can be unstable and computationally demanding. DPO addresses these challenges by directly optimising LMs with a simple classification objective that aligns responses with human preferences. This approach eliminates the need for explicit reward modelling and extensive hyperparameter tuning, enhancing stability and efficiency. DPO optimises the desired behaviours by increasing the relative likelihood of preferred responses while incorporating dynamic importance weights to prevent model degeneration. Thus, DPO simplifies the preference learning pipeline, making it an effective method for training LMs to adhere to human preferences. \\

\begin{figure}[h]
\centering
\includegraphics[width=1\textwidth]{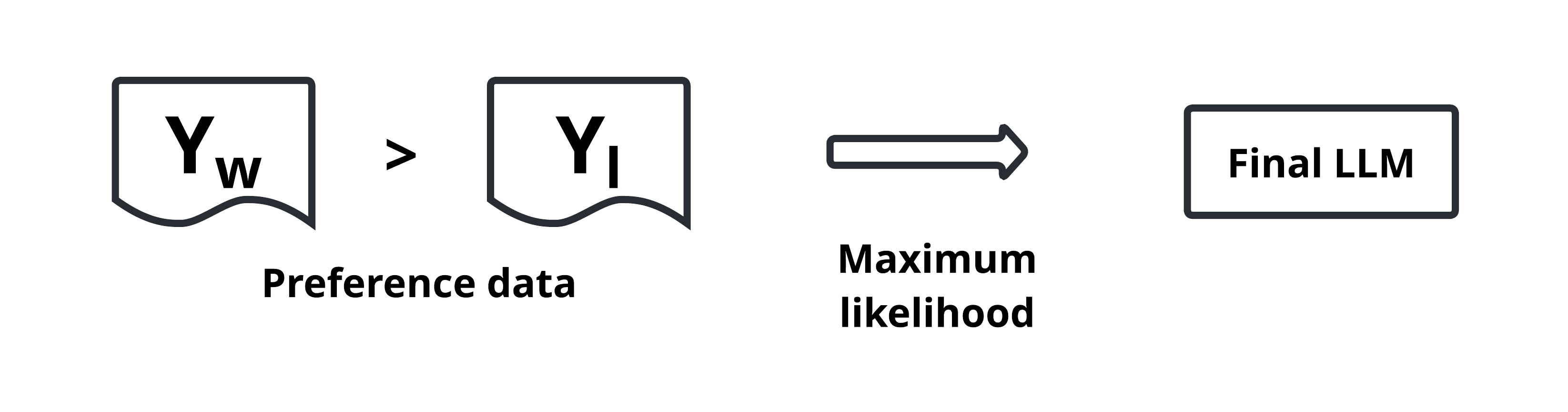}
\caption{Direct Preference Optimisation (DPO) Process Flow. This figure illustrates the Direct Preference Optimisation (DPO) technique used in fine-tuning large language models. The process begins with preference data (\(Y_w > Y_l\)), where \(Y_w\) represents preferred outputs, and \(Y_l\) represents less preferred outputs. Through a maximum likelihood estimation process, this preference data is used to optimise the model's parameters, resulting in the final large language model (LLM). The method is designed to improve the alignment of model outputs with desired user preferences, enhancing the model's effectiveness in specific tasks. (adapted from \cite{dpoPaper})}
\label{dpo}
\end{figure}

\noindent \textbf{Python Library -} HuggingFace TRL package supports the DPO Trainer\footnote{\url{https://huggingface.co/docs/trl/main/en/dpo_trainer}} for training language models from the preference data. 
The DPO training process requires a dataset formatted in a very specific manner. 
If you are utilising the default DPODataCollatorWithPadding data collator, your final dataset object must include three specific entries, which should be labelled as follows:

\begin{itemize}
    \item Prompt
    \item Chosen
    \item Rejected
\end{itemize}

\noindent HuggingFace offers datasets compatible with DPO and can be accessed \href{https://huggingface.co/datasets?other=dpo}{here.}

\subsection{Benefits of DPO}

\begin{enumerate}
    \item \textbf{Direct Alignment with Human Preferences:} DPO directly optimises models to generate responses that align with human preferences, thereby producing more favourable outputs.
    
    \item \textbf{Minimised Dependence on Proxy Objectives:} In contrast to methods that rely on next-word prediction, DPO leverages explicit human preferences, resulting in responses that are more reflective of human behaviour.
    
    \item \textbf{Enhanced Performance on Subjective Tasks:} For tasks requiring subjective judgement, such as dialogue generation or creative writing, DPO excels in aligning the model with human preferences.
\end{enumerate}

\subsection{Best Practices for DPO}

\begin{enumerate}
    \item \textbf{High-Quality Preference Data:} The performance of the model is heavily influenced by the quality of preference data. Ensure the dataset includes clear and consistent human preferences.
    
    \item \textbf{Optimal Beta Value:} Experiment with various beta values to manage the influence of the reference model. Higher beta values prioritise the reference model’s preferences more strongly.

    \item \textbf{Hyperparameter Tuning:} optimise hyperparameters such as learning rate, batch size, and LoRA configuration to determine the best settings for your dataset and task.

    \item \textbf{Evaluation on Target Tasks:} Continuously assess the model’s performance on the target task using appropriate metrics to monitor progress and ensure the achievement of desired results.

    \item \textbf{Ethical Considerations:} Pay attention to potential biases in the preference data and take steps to mitigate them, preventing the model from adopting and amplifying these biases.
\end{enumerate}

\subsection{Tutorial for training models using DPO technique}
The tutorial for DPO training, including the full source code of the training scripts for SFT and DPO, is available \href{https://github.com/huggingface/blog/blob/main/dpo-trl.md}{here.}

\subsection{Is DPO Superior to PPO for LLM Alignment?}
The recent study on DPO superior to PPO for LLM Alignment\cite{dpoPpoSuperiorPaper} investigates the efficacy of reward-based and reward-free methods within RLHF. Reward-based methods, such as those developed by OpenAI, utilise a reward model constructed from preference data and apply actor-critic algorithms like Proximal Policy Optimisation (PPO) to optimise the reward signal. Conversely, reward-free methods, including Direct Preference Optimisation (DPO), RRHF, and PRO, forego an explicit reward function, with DPO focusing exclusively on policy optimisation through a logarithmic representation of the reward function. \\

\noindent One of the objectives of this study is to determine whether DPO is genuinely superior to PPO in the RLHF domain. The study combines theoretical and empirical analyses to uncover the inherent limitations of DPO and identify critical factors that enhance PPO's practical performance in RLHF. \\

\noindent Theoretical findings suggest that DPO may yield biased solutions by exploiting out-of-distribution responses. Empirical results indicate that DPO's performance is notably affected by shifts in the distribution between model outputs and the preference dataset. Furthermore, the study highlights that while iterative DPO may offer improvements over static data training, it still fails to enhance performance in challenging tasks such as code generation.
Ablation studies on PPO reveal essential components for optimal performance, including advantage normalisation, large batch sizes, and exponential moving average updates for the reference model's parameters. These findings form the basis of practical tuning guidelines, demonstrating PPO's robust effectiveness across diverse tasks and its ability to achieve state-of-the-art results in challenging code competition tasks. Specifically, on the CodeContest dataset, the PPO model with 34 billion parameters surpasses AlphaCode-41B, showing a significant improvement in performance metrics.

\section{Odds-Ratio Preference Optimization (ORPO)}

Odds-Ratio Preference Optimization (ORPO) is a novel approach designed to align the output of language models with desired responses by introducing a penalisation mechanism for undesirable outputs. Unlike traditional supervised fine-tuning (SFT) approaches, which focus solely on maximising the likelihood of correct responses, ORPO adds a specific odds-ratio based loss to penalise unwanted generations. This technique provides a refined method for improving preference alignment without relying on a reference model, making it efficient for large-scale implementations.

Given an input sequence \( x \), the log-likelihood of generating an output sequence \( y \) of length \( m \) is computed as:

\[
\log P_\theta(y|x) = \frac{1}{m} \sum_{i=1}^{m} \log P_\theta(y_i|x)
\]

The odds of generating the output sequence \( y \) given input \( x \) is expressed as:

\[
odds_\theta(y|x) = \frac{P_\theta(y|x)}{1 - P_\theta(y|x)}
\]

ORPO introduces an odds-ratio that contrasts the likelihood of generating a preferred (chosen) response \( y_w \) with a less preferred (rejected) response \( y_l \), defined as:

\[
OR_\theta(y_w, y_l|x) = \frac{odds_\theta(y_w|x)}{odds_\theta(y_l|x)}
\]

The ORPO loss function incorporates two components:
\begin{itemize}
    \item \textbf{Supervised Fine-tuning Loss (SFT)}: 
    \[
    L_{SFT} = -\frac{1}{M} \sum_{k=1}^{M}\sum_{i=1}^{|V|} y_{i}^{k} \log p_{i}^{k}
    \]
    where \( y_{i}^{k} \) is a binary indicator for the \( i \)-th token in the vocabulary, and \( p_{i}^{k} \) is its predicted probability.
    
    \item \textbf{Odds-Ratio Loss}: 
    \[
    L_{OR} = - \log \sigma\left( \log \frac{odds_\theta(y_w|x)}{odds_\theta(y_l|x)} \right)
    \]
    where \( \sigma \) is the sigmoid function applied to stabilise the log odds ratio.
\end{itemize}

Thus, the total ORPO objective is:

\[
L_{ORPO} = L_{SFT} + \lambda L_{OR}
\]

where \( \lambda \) controls the strength of preference alignment. This loss function effectively guides the model towards generating the chosen response while discouraging the rejected one, facilitating efficient alignment without the need for additional reference models \cite{orpo_paper}.

\textbf{Advantages of ORPO}: ORPO's strength lies in its ability to perform preference alignment in a monolithic manner, bypassing the need for separate phases of fine-tuning and preference optimisation. This reduces computational overhead and provides state-of-the-art performance across various models, including LLaMA and Mistral, when evaluated on benchmark tasks such as AlpacaEval and MT-Bench \cite{orpo_evaluation}.

\section{Pruning LLMs}

Pruning LLMs involves eliminating unnecessary or redundant components from a neural network to reduce its size and complexity, thereby enhancing its efficiency and performance. This process assists AI developers and engineers in addressing the challenges associated with deploying AI models in resource-limited environments, such as mobile devices, edge computing, or embedded systems. Pruning AI models can be achieved through various techniques, each suited to the type and structure of the neural network, the pruning objective, and the pruning criterion. The following are common approaches:

\begin{enumerate}
\item \textbf{Weight Pruning:} Involves removing weights or connections with minimal magnitude or impact on the output. This method reduces the number of parameters and operations in the model, although it may not necessarily decrease memory footprint or latency.
\item \textbf{Unit Pruning:} Eliminates entire units or neurons with the lowest activation or contribution to the output. This technique can reduce the model's memory footprint and latency but may require retraining or fine-tuning to maintain performance.
\item \textbf{Filter Pruning:} Involves removing entire filters or channels in convolutional neural networks that have the least importance or relevance to the output. This strategy also reduces memory footprint and latency, though it may necessitate retraining or fine-tuning to preserve performance \cite{pruningTechniques}.
\end{enumerate}

\subsection{When to Prune AI Models?}
Pruning AI models can be conducted at various stages of the model development and deployment cycle, contingent on the chosen technique and objective.

\begin{enumerate}
\item \textbf{Pre-Training Pruning:} Leverages prior knowledge or heuristics to determine the optimal network structure before training begins. This approach can save time and resources during training but may necessitate careful design and experimentation to identify the best configuration.
\item \textbf{Post-Training Pruning:} Involves using metrics or criteria to assess the importance or impact of each network component after training. This method helps maintain model performance but may require additional validation and testing to ensure quality and robustness.
\item \textbf{Dynamic Pruning:} Adjusts the network structure during inference or runtime based on feedback or signals. This approach can optimise the model for different scenarios or tasks but may involve higher computational overhead and complexity to implement and execute.
\end{enumerate}

\subsection{Benefits of Pruning}
\begin{enumerate}
\item \textbf{Reduced Size and Complexity:} Pruning decreases the size and complexity of AI models, making them easier to store, transmit, and update.
\item \textbf{Improved Efficiency and Performance:} Pruned models are faster, more energy-efficient, and more reliable.
\item \textbf{Enhanced generalisation and Accuracy:} Pruning can make models more robust, less prone to overfitting, and more adaptable to new data or tasks.
\end{enumerate}

\subsection{Challenges of Pruning}
\begin{enumerate}
\item \textbf{Balance Between Size Reduction and Performance:} Achieving the optimal balance between reducing size and complexity and maintaining performance is challenging; excessive or insufficient pruning can degrade model quality and functionality.
\item \textbf{Choosing Appropriate Techniques:} Selecting the right pruning technique, criterion, and objective for the specific neural network type and structure is crucial, as different methods can produce varying effects and outcomes.
\item \textbf{Evaluation and Validation:} Pruned models need thorough evaluation and validation to ensure pruning has not introduced errors, biases, or vulnerabilities that could impact performance and robustness.
\end{enumerate}

\chapter{Stage 5: Evaluation and Validation}

\section{Steps Involved in Evaluating and Validating Fine-Tuned Models}

\begin{enumerate}
    \item \textbf{Set Up Evaluation Metrics:} Choose appropriate evaluation metrics, such as cross-entropy, to measure the difference between the predicted and actual distributions of the data.
    \item \textbf{Interpret Training Loss Curve:} Monitor and analyse the training loss curve to ensure the model is learning effectively, avoiding patterns of underfitting or overfitting.
    \item \textbf{Run Validation Loops:} After each training epoch, evaluate the model on the validation set to compute relevant performance metrics and track the model's generalisation ability.
    \item \textbf{Monitor and Interpret Results:} Consistently observe the relationship between training and validation metrics to ensure stable and effective model performance.
    \item \textbf{Hyperparameter Tuning and Adjustments:} Adjust key hyperparameters such as learning rate, batch size, and number of training epochs to optimise model performance and prevent overfitting.
\end{enumerate}

\section{Setting Up Evaluation Metrics}

Cross-entropy is a key metric for evaluating LLMs during training or fine-tuning. Originating from information theory, it quantifies the difference between two probability distributions.

\subsection{Importance of Cross-Entropy for LLM Training and Evaluation}

Cross-entropy is crucial for training and fine-tuning LLMs. It serves as a loss function, guiding the model to produce high-quality predictions by minimising discrepancies between the predicted and actual data. In LLMs, each potential word functions as a separate class, and the model's task is to predict the next word given the context. This task is inherently complex, requiring the model to understand syntax, semantics, and context deeply.

\subsection{Beyond Cross-Entropy: Advanced LLM Evaluation Metrics}

While cross-entropy remains fundamental, evaluating LLMs effectively necessitates additional metrics tailored to various aspects of model performance. Here are some advanced metrics employed in LLM evaluation:

\subsubsection{Perplexity}
Perplexity measures how well a probability distribution or model predicts a sample. In the context of LLMs, it evaluates the model’s uncertainty about the next word in a sequence. Lower perplexity indicates better performance, as the model is more confident in its predictions.

\subsubsection{Factuality}
Factuality assesses the accuracy of the information produced by the LLM. It is particularly important for applications where misinformation could have serious consequences. Higher factuality scores correlate with higher output quality.

\subsubsection{LLM Uncertainty}
LLM uncertainty is measured using log probability, helping to identify low-quality generations. Lower uncertainty indicates higher output quality. This metric leverages the log probability of each generated token, providing insights into the model’s confidence in its responses.

\subsubsection{Prompt Perplexity}
This metric evaluates how well the model understands the input prompt. Lower prompt perplexity indicates a clear and comprehensible prompt, which is likely to yield better model performance.

\subsubsection{Context Relevance}
In retrieval-augmented generation (RAG) systems, context relevance measures how pertinent the retrieved context is to the user query. Higher context relevance improves the quality of generated responses by ensuring that the model utilises the most relevant information.

\subsubsection{Completeness}
Completeness assesses whether the model's response fully addresses the query based on the provided context. High completeness ensures that all relevant information is included in the response, enhancing its utility and accuracy.

\subsubsection{Chunk Attribution and Utilisation}
These metrics evaluate how effectively the retrieved chunks of information contribute to the final response. Higher chunk attribution and utilisation scores indicate that the model is efficiently using the available context to generate accurate and relevant answers.

\subsubsection{Data Error Potential}
This metric quantifies the difficulty the model faces in learning from the training data. Higher data quality results in lower error potential, leading to better model performance.

\subsubsection{Safety Metrics}
Safety metrics ensure that the LLM’s outputs are appropriate and non-harmful. These are included in the final sections of the chapter.

\noindent Integrating these advanced metrics provides a holistic view of LLM performance, enabling developers to fine-tune and optimise models more effectively. By employing a metrics-first approach, it is possible to ensure that LLMs not only produce accurate and high-quality outputs but also do so consistently and reliably across diverse applications\footnote{\url{https://www.rungalileo.io/blog/metrics-first-approach-to-llm-evaluation}}.

\section{Understanding the Training Loss Curve}

The training loss curve plots the loss value against training epochs and is essential for monitoring model performance.

\subsection{Interpreting Loss Curves}

An ideal training loss curve shows a rapid decrease in loss during initial stages, followed by a gradual decline and eventual plateau. Specific patterns to look for include:

\begin{enumerate}
    \item \textbf{Underfitting:} High loss value that does not decrease significantly over time, suggesting the model cannot learn the data.
    \item \textbf{Overfitting:} Decreasing training loss with increasing validation loss, indicating the model memorises the training data.
    \item \textbf{Fluctuations:} Significant variations may indicate a high learning rate or noisy gradients.
\end{enumerate}

\begin{figure}[h]
    \centering
    \includegraphics[width=0.8\textwidth]{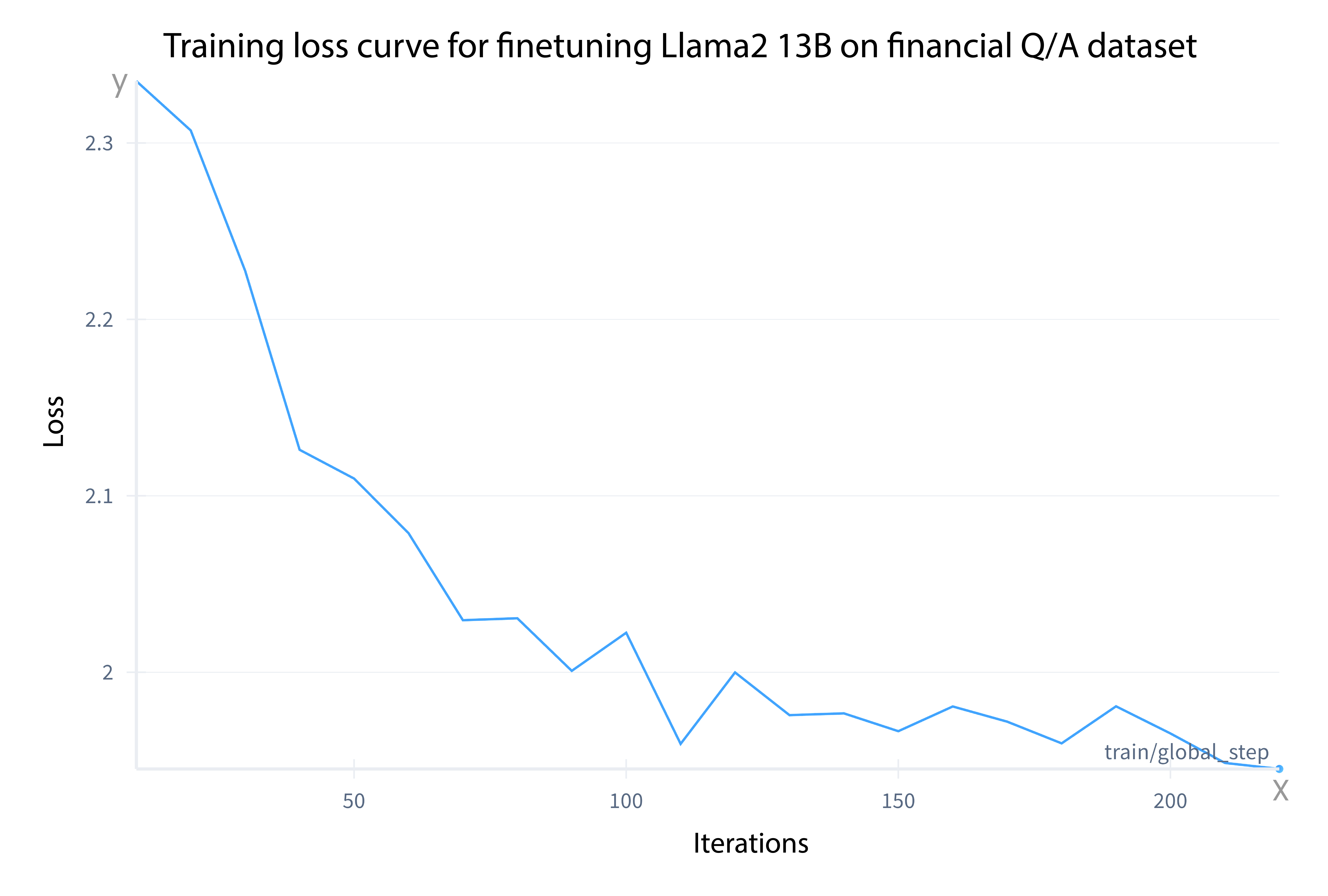}
    \caption{Example training loss curve showing the decline in loss over iterations during the fine-tuning of Llama2 13B on a financial Q/A dataset. The curve illustrates the effectiveness of the fine-tuning process in reducing the loss and improving model performance.}
    \label{fig:training_loss_curve}
\end{figure}

\subsection{Avoiding Overfitting}

Techniques to prevent overfitting include:

\begin{enumerate}
    \item \textbf{Regularisation:} Adds a penalty term to the loss function to encourage smaller weights.
    \item \textbf{Early Stopping:} Stops training when validation performance no longer improves.
    \item \textbf{Dropout:} Randomly deactivates neurons during training to reduce sensitivity to noise.
    \item \textbf{Cross-Validation:} Splits data into multiple subsets for training and validation to assess model generalisation.
    \item \textbf{Batch Normalisation:} Normalises inputs to each layer during training to stabilise the learning process.
    \item \textbf{Larger Datasets and Batch Sizes:} Reduces overfitting by increasing the amount of diverse data and batch sizes.
\end{enumerate}

\subsection{Sources of Noisy Gradients}

Noisy gradients are common during the training of machine learning models, including LLMs. They arise from variability in gradient estimates due to stochastic gradient descent and its variants. Strategies to manage noisy gradients include:

\begin{enumerate}
    \item \textbf{Learning Rate Scheduling:} Gradually decreasing the learning rate during training can reduce the impact of noisy gradients.
    \item \textbf{Gradient Clipping:} Setting a threshold for gradient values prevents large updates that can destabilise training.
\end{enumerate}

\section{Running Validation Loops}

Validation loops provide an unbiased evaluation of model performance. Typical steps include:

\begin{enumerate}
    \item \textbf{Split Data:} Divide the dataset into training and validation sets.
    \item \textbf{Initialise Validation:} Evaluate the model on the validation set at the end of each epoch.
    \item \textbf{Calculate Metrics:} Compute relevant performance metrics, such as cross-entropy loss.
    \item \textbf{Record Results:} Log validation metrics for each epoch.
    \item \textbf{Early Stopping:} Optionally stop training if validation loss does not improve for a predefined number of epochs.
\end{enumerate}

\section{Monitoring and Interpreting Results}

Monitoring validation results involves analysing trends in validation metrics over epochs. Key aspects include:

\begin{enumerate}
    \item \textbf{Consistent Improvement:} Indicates good model generalisation if both training and validation metrics improve and plateau.
    \item \textbf{Divergence:} Suggests overfitting if training metrics improve while validation metrics deteriorate.
    \item \textbf{Stability:} Ensure validation metrics do not fluctuate significantly, indicating stable training.
\end{enumerate}

\section{Hyperparameter Tuning and Other Adjustments}

Fine-tuning involves adjusting key hyperparameters to achieve optimal performance. Important hyperparameters include:

\begin{enumerate}
    \item \textbf{Learning Rate:} Determines the step size for updating model weights. A good starting point is 2e-4, but this can vary.
    \item \textbf{Batch Size:} Larger batch sizes lead to more stable updates but require more memory.
    \item \textbf{Number of Training Epochs:} Balancing the number of epochs ensures the model learns sufficiently without overfitting or underfitting.
    \item \textbf{Optimiser:} Optimisers like Paged ADAM optimise memory usage, advantageous for large models.
\end{enumerate}

Other tunable parameters include dropout rate, weight decay, and warmup steps.

\subsection{Data Size and Quality}

The efficacy of LLMs is directly impacted by the quality of their training data. Ensuring that datasets are clean, relevant, and adequate is crucial. Data cleanliness refers to the absence of noise, errors, and inconsistencies within the labelled data. For example, having a phrase like “This article suggests…” multiple times in the training data can corrupt the response of LLMs and add a bias towards using this specific phrase more often and in inappropriate situations.

\section{Benchmarking Fine-Tuned LLMs}

Modern LLMs are assessed using standardised benchmarks such as GLUE, SuperGLUE, HellaSwag, TruthfulQA, and MMLU (See Table \ref{llmBenchmarks}). These benchmarks evaluate various capabilities and provide an overall view of LLM performance.

\begin{table}[H]
\centering
\resizebox{\textwidth}{!}{%
\begin{tabular}{|l|p{0.5\linewidth}|l|}
\hline
\textbf{Benchmark} & \textbf{Description} & \textbf{Reference URL} \\ \hline
GLUE & Provides a standardised set of diverse NLP tasks to evaluate the effectiveness of different language models & \href{https://gluebenchmark.com/}{Source} \\ \hline
SuperGLUE & Compares more challenging and diverse tasks with GLUE, with comprehensive human baselines & \href{https://super.gluebenchmark.com/}{Source} \\ \hline
HellaSwag & Evaluates how well an LLM can complete a sentence & \href{https://rowanzellers.com/hellaswag/}{Source} \\ \hline
TruthfulQA & Measures truthfulness of model responses & \href{https://github.com/sylinrl/TruthfulQA}{Source} \\ \hline
MMLU & Evaluates how well the LLM can multitask & \href{https://github.com/hendrycks/test}{Source} \\ \hline
IFEval & Tests a model’s ability to follow explicit instructions, focusing on formatting adherence & \href{https://github.com/google-research/google-research/tree/master/instruction_following_eval}{Source} \\ \hline
BBH (Big Bench Hard) & 23 challenging tasks from the BigBench dataset to evaluate LLMs using objective metrics & \href{https://github.com/suzgunmirac/BIG-Bench-Hard}{Source} \\ \hline
MATH & Compilation of high-school level competition problems formatted using LaTeX and Asymptote & \href{https://github.com/hendrycks/apps}{Source} \\ \hline
GPQA & Challenging knowledge dataset with questions crafted by PhD-level domain experts & \href{https://github.com/idavidrein/gpqa}{Source} \\ \hline
MuSR & Dataset with complex problems requiring models to integrate reasoning with long-range context parsing & \href{https://github.com/Zayne-Sprague/MuSR}{Source} \\ \hline
MMLU-PRO & Refined version of MMLU with higher quality and more challenging multiple-choice questions & \href{https://github.com/TIGER-AI-Lab/MMLU-Pro}{Source} \\ \hline
ARC & Measures machine reasoning with a dataset of grade-school science questions & \href{https://allenai.org/data/arc}{Source} \\ \hline
COQA & A dataset for building conversational question-answering systems & \href{https://stanfordnlp.github.io/coqa/}{Source} \\ \hline
DROP & Evaluates the ability to perform discrete reasoning over paragraphs of text & \href{https://allennlp.org/drop}{Source} \\ \hline
SQuAD & A reading comprehension dataset for evaluating models' ability to answer questions based on passages of text & \href{https://rajpurkar.github.io/SQuAD-explorer/}{Source} \\ \hline
TREC & A benchmark for evaluating text retrieval methodologies & \href{https://trec.nist.gov/}{Source} \\ \hline
WMT & A dataset and benchmark for evaluating machine translation models & \href{http://www.statmt.org/wmt20/}{Source} \\ \hline
XNLI & A dataset for evaluating cross-lingual language understanding & \href{https://cims.nyu.edu/~sbowman/xnli/}{Source} \\ \hline
PiQA & A dataset for evaluating models' understanding of physical interactions & \href{https://yonatanbisk.com/piqa/}{Source} \\ \hline
Winogrande & A large-scale dataset for evaluating commonsense reasoning & \href{https://mosaic.allenai.org/projects/winogrande}{Source} \\ \hline
\end{tabular}%
}
\caption{Detailed Overview of Benchmark Datasets Used for Evaluating Language Model Performance.}
\label{llmBenchmarks}
\end{table}

\noindent As LLMs evolve, so do benchmarks, with new standards such as BigCodeBench challenging current benchmarks and setting new standards in the domain. Given the diverse nature of LLMs and the tasks they can perform, the choice of benchmarks depends on the specific tasks the LLM is expected to handle. For generic applicability, various benchmarks for different downstream applications and reasoning should be utilised. For domain/task-specific LLMs, benchmarking can be limited to relevant benchmarks like BigCodeBench for coding.

\section{Evaluating Fine-Tuned LLMs on Safety Benchmark}

The safety aspects of Large Language Models (LLMs) are increasingly scrutinised due to their ability to generate harmful content when influenced by jailbreaking prompts. These prompts can bypass the embedded safety and ethical guidelines within the models, similar to code injection techniques used in traditional computer security to circumvent safety protocols. Notably, models like ChatGPT, GPT-3, and InstructGPT are vulnerable to such manipulations that remove content generation restrictions, potentially violating OpenAI’s guidelines. This underscores the necessity for robust safeguards to ensure LLM outputs adhere to ethical and safety standards.

\noindent DecodingTrust \cite{decodingTrustPaper} provides a comprehensive evaluation of the trustworthiness of LLMs, notably comparing GPT-4 with GPT-3.5 (ChatGPT). This evaluation spans several critical areas:

\begin{enumerate}
    \item \textbf{Toxicity:} Optimisation algorithms and generative models are employed to create challenging prompts that test the model's ability to avoid generating harmful content.
    
    \item \textbf{Stereotype Bias:} An array of demographic groups and stereotype topics are utilised to assess model bias, helping to understand and mitigate prejudiced responses.

    \item \textbf{Adversarial Robustness:} The resilience of models against adversarial attacks is tested by challenging them with sophisticated algorithms intended to deceive or mislead.

    \item \textbf{Out-of-Distribution (OOD) Robustness:} Models are evaluated on their ability to handle inputs that differ significantly from their training data, such as poetic or Shakespearean styles.

    \item \textbf{Robustness to Adversarial Demonstrations:} Demonstrations that contain misleading information are used to test the model's robustness across various tasks.

    \item \textbf{Privacy:} Various levels of privacy evaluation assess how well models safeguard sensitive information during interactions and understand privacy-related contexts.

    \item \textbf{Hallucination Detection:} Identifies instances where the model generates information not grounded in the provided context or factual data. Lower hallucination rates improve the reliability and trustworthiness of the LLM’s outputs.
        
    \item \textbf{Tone Appropriateness:} Assesses whether the model’s output maintains an appropriate tone for the given context. This is particularly important for applications in customer service, healthcare, and other sensitive areas.

    \item \textbf{Machine Ethics:} Ethical assessments involve testing models with scenarios that require moral judgments, using datasets like ETHICS and Jiminy Cricket.

    \item \textbf{Fairness:} The fairness of models is evaluated by generating tasks that vary protected attributes, ensuring equitable responses across different demographic groups.

\end{enumerate}

\noindent The dataset employed for evaluating the aforementioned eight safety dimensions can be found \href{https://github.com/AI-secure/DecodingTrust/tree/main/data}{here}.

\noindent In partnership with HuggingFace, the \href{https://huggingface.co/blog/leaderboard-decodingtrust}{LLM Safety Leaderboard} utilises DecodingTrust's framework to provide a unified evaluation platform for LLM safety. This allows researchers and practitioners to better understand the capabilities, limitations, and risks associated with LLMs. Users are encouraged to submit their models to HuggingFace for evaluation, ensuring they meet the evolving standards of safety and reliability in the field.

\section{Evaluating Safety of Fine-Tuned LLM using AI Models}

\subsection{Llama Guard}

Llama Guard 2\cite{llamaGuardPaper} is a safeguard model built on LLMs for managing risks in conversational AI applications. It effectively categorises both input prompts and responses from AI agents using a detailed safety risk taxonomy tailored to identify potential legal and policy risks in AI interactions. It utilises a detailed safety risk taxonomy designed to identify and manage potential legal and policy risks in interactions involving conversational AI. This taxonomy enables effective classification in areas such as:

\begin{itemize}
    \item \textbf{Violence \& Hate,} addressing content that could incite violent acts or discrimination.
    
    \item \textbf{Sexual Content,} targeting sexually explicit material or behaviour, especially involving minors.
    
    \item \textbf{Guns \& Illegal Weapons,} concerning the promotion or instruction of illegal armaments.
    
    \item \textbf{Regulated or Controlled Substances,} covering illegal drugs and other controlled substances.
    
    \item \textbf{Suicide \& Self-Harm,} aimed at content that could encourage self-destructive behaviour.
    
    \item \textbf{Criminal Planning,} for content that could assist in planning or executing criminal activities.
    
\end{itemize}

\noindent The core of Llama Guard 2 is its robust framework that allows for both prompt and response classification, supported by a high-quality dataset that enhances its ability to monitor conversational exchanges. Operating on a Llama2-7b model, Llama Guard 2 has been instruction-tuned to deliver strong performance on benchmarks like the OpenAI Moderation Evaluation dataset and ToxicChat, where it matches or surpasses the capabilities of existing content moderation tools. 

\noindent The model supports multi-class classification and generates binary decision scores. Its instruction fine-tuning allows for extensive customisation of tasks and adaptation of output formats. This feature enables users to modify taxonomy categories to align with specific use cases and supports flexible prompting capabilities, including zero-shot and few-shot applications. The adaptability and effectiveness of Llama Guard make it a vital resource for developers and researchers. By making its model weights publicly available, Llama Guard 2 encourages ongoing development and customisation to meet the evolving needs of AI safety within the community. 

\noindent Llama Guard 3 represents the latest advancement over Llama Guard 2, having been fine-tuned on the Llama 3 8b model. The key difference between the two versions is that Llama Guard 3 expands upon the capabilities of Llama Guard 2 by introducing three new categories: \textbf{Defamation, Elections, and Code Interpreter Abuse.} 

\noindent \textbf{Python Library:} Llama Guard 3 is accessible via HuggingFace's AutoModelForCausalLM.\footnote{\url{https://huggingface.co/docs/transformers/en/model_doc/auto\#transformers.AutoModelForCausalLM}} A detailed tutorial is available at this \href{https://huggingface.co/meta-llama/Llama-Guard-3-8B}{link}. Please note that access to the model requires submitting a request to Hugging Face with the user details. Additionally, the model weights can be downloaded from the Meta platform by providing user details, and the link can be found \href{https://llama.meta.com/llama-downloads}{here}.

\noindent The prompt formats for these two models also differ, with the specific formats for Llama Guard 2 available \href{https://llama.meta.com/docs/model-cards-and-prompt-formats/meta-llama-guard-2}{here} and Llama Guard 3 is accessible \href{https://llama.meta.com/docs/model-cards-and-prompt-formats/llama-guard-3}{here}. 

\subsection{Shield Gemma}
ShieldGemma \cite{shieldGemmaPaper} is an advanced content moderation model built on the Gemma2 platform, designed to enhance the safety and reliability of interactions between LLMs and users. It effectively filters both user inputs and model outputs to mitigate key harm types, including offensive language, hate speech, misinformation, and explicit content. The model's scalability, with options ranging from 2B to 27B parameters, allows for tailored applications that meet specific needs, such as reducing latency in online safety applications or enhancing performance in complex decision-making tasks.

\noindent A distinguishing feature of ShieldGemma is its novel approach to data curation. It leverages synthetic data generation techniques to create high-quality datasets that are robust against adversarial prompts and fair across diverse identity groups. This reduces the need for extensive human annotation, streamlining the data preparation process while ensuring the model's effectiveness. Compared to existing content moderation tools like LlamaGuard and WildGuard, which typically offer fixed-size models and limited customisation, ShieldGemma's flexible architecture and advanced data handling capabilities provide a more adaptable and efficient solution. These innovations position ShieldGemma as a significant advancement in LLM-based content moderation, offering developers and researchers a versatile tool that promotes safer and more reliable AI interactions across various platforms. 

\noindent \textbf{Python Library:} The ShieldGemma series is available on HuggingFace via AutoModelForCausalLM. The models can be accessed \href{https://huggingface.co/google}{here}. A tutorial for running ShieldGemma 2B on Google Colab can be found \href{https://colab.research.google.com/github/google/generative-ai-docs/blob/main/site/en/responsible/docs/safeguards/shieldgemma_on_keras.ipynb}{here}. Similar to Llama Guard series, ShieldGemma series also has guidelines for prompting and it can be found \href{https://ai.google.dev/gemma/docs/shieldgemma/model_card}{here}.

\subsection{WILDGUARD}
WILDGUARD \cite{wildGuardPaper} is an innovative open-source tool developed to enhance the safety of interactions with large language models (LLMs). This tool addresses three critical moderation tasks: detecting harmful intent in user prompts, identifying safety risks in model responses, and determining when a model appropriately refuses unsafe requests. Central to its development is WILDGUARD MIX\footnote{\url{https://huggingface.co/datasets/allenai/wildguardmix}}, a meticulously curated dataset comprising 92,000 labelled examples that include both benign prompts and adversarial attempts to bypass safety measures. The dataset is divided into WILDGUARD TRAIN, used for training the model, and WILDGUARD TEST, consisting of high-quality human-annotated examples for evaluation.

\noindent The WILDGUARD model itself is fine-tuned on the Mistral-7B language model using the WILDGUARD TRAIN dataset, enabling it to perform all three moderation tasks in a unified, multi-task manner. Results show that WILDGUARD surpasses existing open-source moderation tools in effectiveness, particularly excelling in handling adversarial prompts and accurately detecting model refusals. On many benchmarks, WILDGUARD's performance is on par with or exceeds that of GPT-4, a much larger, closed-source model. 

\noindent The quick start guide and additional information on WILDGUARD are available in GitHub and it can be accessed \href{https://github.com/allenai/wildguard}{here}.

\chapter{Stage 6: Deployment}

\section{Steps Involved in Deploying the Fine-Tuned Model}

\begin{enumerate}
    \item \textbf{Model Export:} Save the fine-tuned model in a suitable format (e.g., ONNX, TensorFlow SavedModel, PyTorch) for deployment.
    
    \item \textbf{Infrastructure Setup:} Prepare the deployment environment, including necessary hardware, cloud services, and containerisation tools.
    
    \item \textbf{API Development:} Create APIs to allow applications to interact with the model, facilitating prediction requests and responses.

    \item \textbf{Deployment:} Deploy the model to the production environment, making it accessible to end-users or applications.
\end{enumerate}

\section{Cloud-Based Providers for LLM Deployment}

Cloud-based large language model (LLM) inferencing frequently employs a pricing model based on the number of tokens processed. Users are charged according to the volume of text analysed or generated by the model. While this pricing structure can be cost-effective for sporadic or small-scale usage, it may not always be economical for larger or continuous workloads.

\noindent In some scenarios, hosting an LLM solution in-house may offer better long-term cost savings, especially if there is consistent or high-volume usage. Managing your own infrastructure provides greater control over resource allocation and allows for cost optimisation based on specific needs. Additionally, self-hosting offers advantages in terms of data privacy and security, as sensitive information remains within your own environment. \\

\noindent However, it is crucial to carefully evaluate the total cost of ownership when comparing cloud-based solutions with self-hosted alternatives. This evaluation should consider factors such as hardware expenses, maintenance, and operational overheads. Ultimately, the decision should be informed by a comprehensive cost-benefit analysis, considering both short-term affordability and long-term sustainability.

\noindent Several companies offer deployment services for large language models (LLMs), providing a range of tools and platforms to efficiently implement and manage these models. Here’s a detailed list of some prominent providers and their services:

\begin{itemize}
    \item \textbf{Amazon Web Services (AWS)}
    \begin{itemize}
        \item \textbf{Amazon Bedrock:} This service offers a suite of foundation models including Amazon Titan, which supports various NLP tasks such as summarisation and text generation. Bedrock integrates seamlessly with other AWS services for scalable and secure deployment.
        
        \item \textbf{Amazon SageMaker:} Provides an end-to-end machine learning service that includes tools for building, training, and deploying LLMs. SageMaker JumpStart offers pre-trained models and step-by-step guides to simplify the deployment process.
        
        \item \textbf{Tutorial:} \href{https://docs.aws.amazon.com/bedrock/latest/userguide/agents-deploy.html}{This tutorial} explains the deployment of LLM Agents on Amazon Bedrock. Another \href{https://aws.amazon.com/blogs/machine-learning/fine-tune-and-deploy-language-models-with-amazon-sagemaker-canvas-and-amazon-bedrock/}{tutorial} explains end-to-end fine-tuning and deployment of LLMs with Sagemaker Canvas and Amazon Bedrock. \href{https://docs.aws.amazon.com/bedrock/latest/userguide/general-guidelines-for-bedrock-users.html}{General guidelines of Amazon Bedrock} for LLM users can be found here.
    \end{itemize}
    
    \item \textbf{Microsoft Azure}
    \begin{itemize}
        \item \textbf{Azure OpenAI Service:} This service offers access to OpenAI’s powerful models like GPT-3.5 and Codex. It provides capabilities for embedding, image generation with DALL-E, and speech-to-text with Whisper. Azure’s integration with OpenAI models ensures robust deployment options for various applications.
        
        \item \textbf{Azure Machine Learning:} Supports the deployment of custom and pre-trained models, offering tools for model management, deployment, and monitoring. It integrates with Azure’s broader ecosystem for scalable and secure ML operations.
        
        \item \textbf{Tutorial:} \href{https://learn.microsoft.com/en-us/azure/ai-services/openai/how-to/create-resource?pivots=web-portal}{Here} is the tutorial for creating and deploying an Azure OpenAI Service in Microsoft Azure platform.
    \end{itemize}
    
    \item \textbf{Google Cloud Platform (GCP)}
    \begin{itemize}
        \item \textbf{Vertex AI:} This platform allows the deployment of large language models with tools for training, tuning, and serving models. Vertex AI supports models like BERT and GPT-3, providing extensive MLOps capabilities for end-to-end management.
        
        \item \textbf{Cloud AI API:} Offers APIs for NLP tasks such as translation, sentiment analysis, and entity recognition. These APIs are backed by Google’s powerful infrastructure, ensuring high performance and reliability.
        
        \item \textbf{Tutorial:} \href{https://cloud.google.com/vertex-ai/docs/tutorials/tabular-bq-prediction/train-and-deploy-model}{This document} contains a tutorial for training and deploying an LLM in GCP.
    \end{itemize}
    
    \item \textbf{Hugging Face}
    \begin{itemize}
        \item \textbf{Inference API:} This service allows users to deploy and manage LLMs hosted on Hugging Face’s infrastructure. It supports various models from the Transformers library and provides an easy-to-use API for integrating these models into applications.
        
        \item \textbf{Spaces:} A collaborative environment where users can deploy and share models using Hugging Face’s hosting platform. It supports deploying custom models and interactive demos.
        
        \item \textbf{Tutorial:} \href{https://huggingface.co/blog/inference-endpoints-llm}{This document} contains a tutorial for training and deploying an LLM using HuggingFace Inference API.
    \end{itemize}
    
    \item \textbf{Other Platforms}
    \begin{itemize}
        \item \textbf{OpenLLM:} Provides deployment solutions \href{https://github.com/bentoml/OpenLLM?ref=content.whylabs.ai}{here}.
        \item \textbf{Deepseed:} Offers deployment solutions \href{https://github.com/microsoft/DeepSpeed?ref=content.whylabs.ai}{here}.
    \end{itemize}
\end{itemize}

\section{Techniques for Optimising Model Performance During Inference}

Optimising model performance during inference is crucial for the efficient deployment of large language models (LLMs). The following advanced techniques offer various strategies to enhance performance, reduce latency, and manage computational resources effectively.

\subsection{Traditional On-Premises GPU-Based Deployments}

This conventional approach to deploying large language models (LLMs) involves using Graphics Processing Units (GPUs) due to their parallel processing capabilities, which enable fast and efficient inference. However, this method requires upfront hardware investment and may not be suitable for applications with fluctuating demand or limited budgets. GPU-based deployments face several challenges:

\begin{enumerate}
    \item Resource utilisation may suffer during periods of low demand due to idle servers.
    \item Scaling up or down often requires physical hardware modifications, which can be time-consuming.
    \item Centralised servers can introduce single points of failure and scalability limitations.
\end{enumerate}

\noindent To mitigate these issues, strategies such as load balancing between multiple GPUs, fallback routing, model parallelism, and data parallelism can be employed to achieve better results. Optimisation techniques like distributed inference using PartialState from accelerate can further enhance efficiency.

\subsubsection{Example use case: Large-Scale NLP Application}

For instance, a large e-commerce platform implemented traditional on-premises GPU-based deployment to handle millions of customer queries daily. By utilising load balancing and model parallelism, they were able to achieve a significant reduction in latency and improved customer satisfaction.

\subsection{Distributed LLM: Torrent-Style Deployment and Parallel Forward Passes}

An innovative deployment strategy for large language models (LLMs) involves distributing them across multiple GPUs in a decentralised, torrent-style manner. Libraries like Petals\footnote{\url{https://github.com/bigscience-workshop/petals}} can perform this task. Petals functions as a decentralised pipeline designed for rapid neural network inference by partitioning the model into distinct blocks or layers, which are distributed across multiple geographically dispersed servers. Users can connect their own GPUs to this network, acting as both contributors and clients who can access and apply the model to their data.\\

\noindent When a client request is received, the network routes it through a series of servers optimised to minimise the total forward pass time. Each server dynamically selects the most optimal set of blocks, adapting to the current bottlenecks in the pipeline. This framework leverages decentralisation principles to distribute computational load across diverse regions, sharing computational resources and GPUs in a way that reduces the financial burden on individual organisations. This collaborative approach not only optimises resource utilisation but also fosters a global community dedicated to shared AI goals.

\begin{figure}[H]
    \centering
    \includegraphics[width=0.7\textwidth]{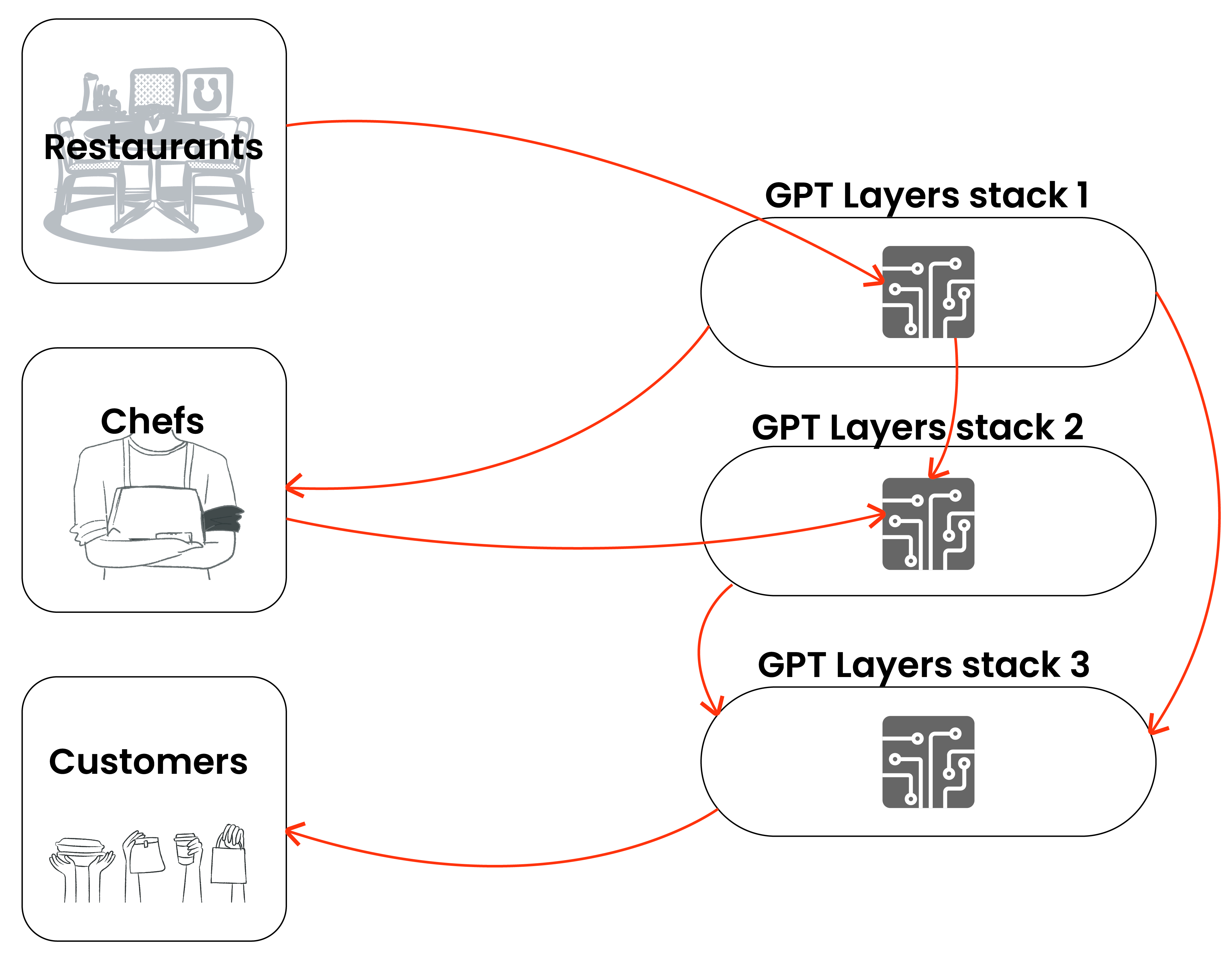}
    \caption{Conceptual Representation of Distributed LLM Deployment Using a Torrent-Style Approach. This figure illustrates the distributed deployment of a Large Language Model (LLM) using a torrent-style approach, where multiple GPT model layers (stacks) are distributed across different nodes (represented by chefs) and perform parallel forward passes. The process mimics the flow of orders from customers (input data) through restaurants (intermediate processing layers) to chefs (model layers), highlighting the efficiency of parallel processing and distributed computing in handling large-scale language models. This approach is essential for reducing inference latency and improving the scalability of LLMs across diverse computational environments. (adapted from \cite{deploymentStrategies})}
    \label{DistributedLLM:Torrent-Style}
\end{figure}

\subsubsection{Example use case: Global Research Collaboration}

A consortium of research institutions implemented a distributed LLM using the Petals framework to analyse large datasets across different continents. By leveraging the decentralised nature of Petals, they achieved high efficiency in processing and collaborative model development.

\subsection{WebGPU-Based Deployment of LLM}

This deployment option for large language models (LLMs) involves utilising WebGPU, a web standard that provides a low-level interface for graphics and compute applications on the web platform. With WebGPU, organisations can harness the power of GPUs directly within web browsers, enabling efficient inference for LLMs in web-based applications. WebGPU enables high-performance computing and graphics rendering directly within the client’s web browser. It allows developers to utilise the client’s GPU for tasks such as rendering graphics, accelerating computational workloads, and performing parallel processing, all without the need for plugins or additional software installations. This capability permits complex computations to be executed efficiently on the client’s device, leading to faster and more responsive web applications.

\subsection{LLM on WebGPU using WebLLM}

Clients can access powerful large language models and chatbots directly in their browser, leveraging WebGPU acceleration. This approach eliminates server dependencies, providing users with exceptional performance and enhanced privacy. WebLLM facilitates the use of large language models directly in the client’s browser to perform tasks such as filtering out personally identifiable information (PII) or named entity recognition (NER) on data without transmitting it over the network. This ensures enhanced privacy and security by retaining sensitive information on the client side.

\begin{figure}[H]
    \centering
    \includegraphics[width=0.8\textwidth]{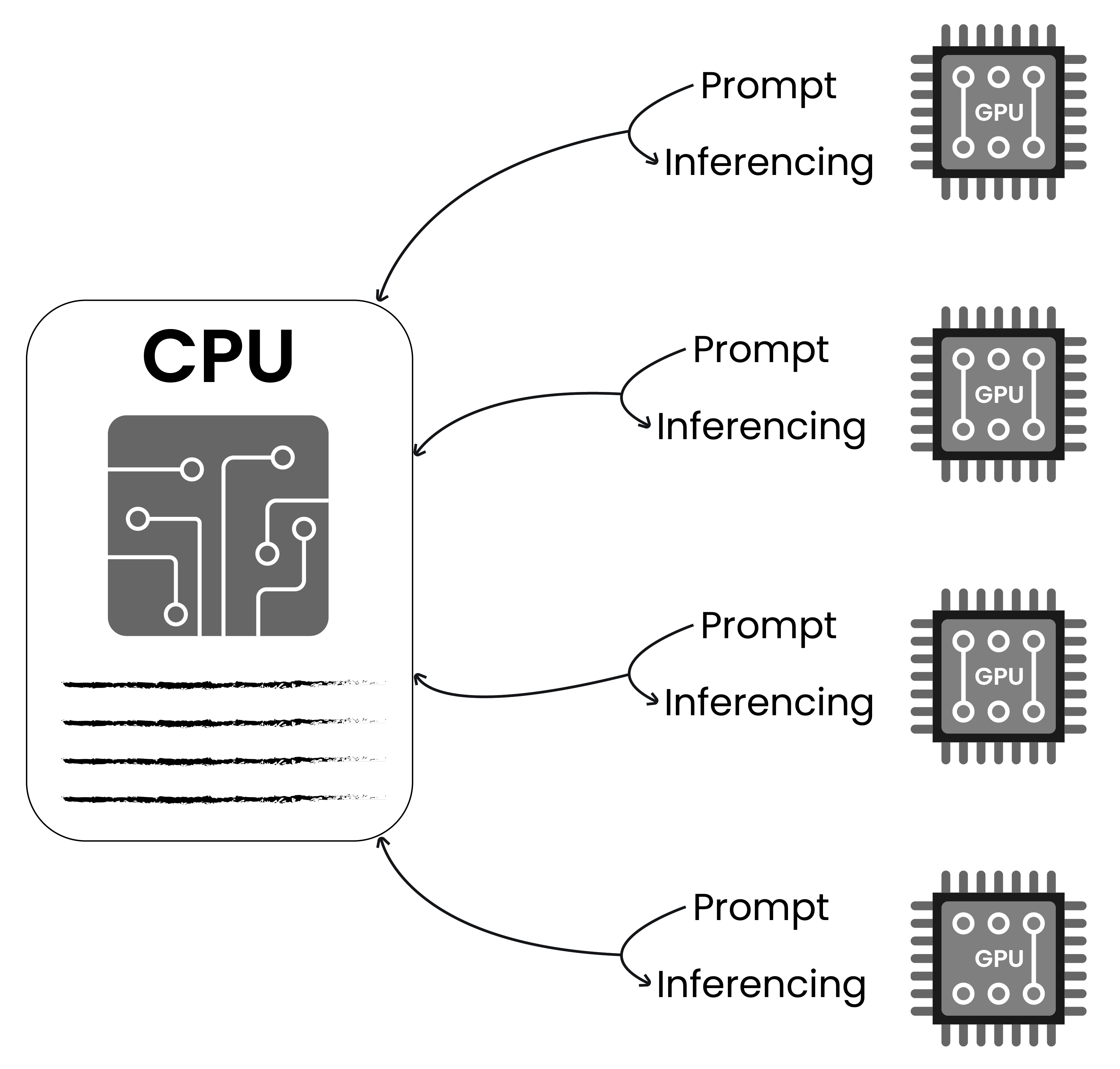}
    \caption{WebGPU-Based Deployment of LLM: This diagram illustrates the architecture of deploying a large language model (LLM) using WebGPU technology. The CPU manages the distribution of prompt inferencing tasks to multiple GPUs, which then process these prompts in parallel, enhancing efficiency and scalability in LLM deployment across web-based platforms. (adapted from \cite{deploymentStrategies})}
    \label{webgpullm}
\end{figure}

\subsubsection{Additional Use Cases for WebLLM}

\begin{enumerate}
    \item \textbf{Language Translation:} Enable real-time translation of text directly in the browser, allowing users to communicate across language barriers without transmitting their messages over the network.
    \item \textbf{Code Autocompletion:} Develop code editors that provide intelligent autocompletion suggestions based on context, leveraging WebLLM to understand and predict code snippets.
    \item \textbf{Customer Support Chatbots:} Implement chatbots on websites to provide instant customer support and answer frequently asked questions without relying on external servers.
    \item \textbf{Data Analysis and Visualisation:} Create browser-based tools for analysing and visualising data, with WebLLM assisting in data processing, interpretation, and generating insights.
    \item \textbf{Personalised Recommendations:} Develop recommendation engines that offer personalised product recommendations, content suggestions, or movie/music recommendations based on user preferences and behaviour.
    \item \textbf{Privacy-Preserving Analytics:} Develop analytics platforms that perform data analysis directly in the browser, ensuring that sensitive information remains on the client side and reducing the risk of data breaches.
\end{enumerate}

\subsubsection{Example use case: Privacy-Focused Web Application}

A healthcare startup deployed an LLM using WebLLM to process patient information directly within the browser, ensuring data privacy and compliance with healthcare regulations. This approach significantly reduced the risk of data breaches and improved user trust.

\subsection{Quantised LLMs}

Model quantisation is a technique utilised to reduce the size of an AI model by representing its parameters with fewer bits. In traditional machine learning models, each parameter (e.g., weights and biases in neural networks) is typically stored as a 32-bit floating-point number, necessitating significant memory and computational resources, particularly for large models. Quantisation aims to alleviate this by reducing the precision of these parameters. For instance, instead of storing each parameter as a 32-bit floating-point number, they may be represented using fewer bits, such as 8-bit integers. This compression reduces the memory footprint of the model, making it more efficient to deploy and execute, especially in resource-constrained environments like mobile devices or edge devices. QLoRA is a popular example of this quantisation for LLMs and can be used to deploy LLMs locally or host them on external servers.

\subsubsection{Example use case: Edge Device Deployment}

A tech company used quantised LLMs to deploy advanced NLP models on mobile devices, enabling offline functionality for applications such as voice recognition and translation. This deployment significantly improved app performance and user experience by reducing latency and reliance on internet connectivity.

\subsection{vLLMs}

The vLLM\footnote{\url{https://docs.vllm.ai/en/stable/}} system efficiently handles requests by employing a block-level memory management method and preemptive request scheduling. It utilises the PagedAttention\cite{pagedAttentionPaper} algorithm to manage the key-value (KV) cache, thereby reducing memory waste and fragmentation. By batching requests and sharing physical blocks across multiple samples, vLLM optimises memory usage and enhances throughput. Performance tests indicate that vLLM surpasses other systems in various decoding scenarios. Consider a transformer-based model tasked with summarising a lengthy book. Traditional transformers process the entire book simultaneously, which can be both computationally and memory-intensive, especially for extended texts. With PagedAttention, the book is divided into smaller segments or pages. The model then focuses on summarising one page at a time, rather than the entire book simultaneously. This approach reduces computational complexity and memory requirements, making it more feasible to process and summarise lengthy texts efficiently.

\subsubsection{Example use case: High-Volume Content Generation}

A content marketing agency implemented vLLMs for generating large volumes of SEO-optimised content. By leveraging the efficient memory management of vLLMs, they were able to handle multiple concurrent requests, significantly increasing their content production rate while maintaining high quality.

\section{Key Considerations for Deployment of LLMs}

Deploying large language models (LLMs) effectively requires careful planning and consideration of various factors to ensure optimal performance, cost-efficiency, and security. Key considerations include:

\begin{itemize}
    \item \textbf{Infrastructure Requirements:}
    \begin{itemize}
        \item \textbf{Compute Resources:} Ensure adequate CPU/GPU resources to handle the model's computational demands. High-performance GPUs are typically required for efficient inference and training.
        \item \textbf{Memory:} LLMs, especially those with billions of parameters, require substantial memory. Memory management techniques such as quantisation and model parallelism can be employed to optimise usage.
    \end{itemize}

    \item \textbf{Scalability:}
    \begin{itemize}
        \item \textbf{Horizontal Scaling:} Plan for horizontal scaling to distribute the load across multiple servers, which can improve performance and handle increased demand.
        \item \textbf{Load Balancing:} Implement load balancing strategies to ensure even distribution of requests and prevent any single point of failure.
    \end{itemize}

    \item \textbf{Cost Management:}
    \begin{itemize}
        \item \textbf{Token-based Pricing:} Understand the cost implications of token-based pricing models offered by cloud providers. This model charges based on the number of tokens processed, which can become expensive with high usage.
        \item \textbf{Self-Hosting:} Evaluate the costs and benefits of self-hosting versus cloud hosting. Self-hosting might offer long-term savings for consistent, high-volume usage but requires significant upfront investment in hardware and ongoing maintenance.
    \end{itemize}

    \item \textbf{Performance Optimisation:}
    \begin{itemize}
        \item \textbf{Latency:} Minimise latency to ensure real-time performance, particularly for applications requiring instant responses like chatbots and virtual assistants.
        \item \textbf{Throughput:} Maximise throughput to handle a high volume of requests efficiently. Techniques like batching and efficient memory management (e.g., PagedAttention) can help.
    \end{itemize}

    \item \textbf{Security and Privacy:}
    \begin{itemize}
        \item \textbf{Data Security:} Implement robust security measures to protect sensitive data, including encryption and secure access controls.
        \item \textbf{Privacy:} Ensure compliance with data privacy regulations by keeping sensitive data within your environment if self-hosting, or ensuring cloud providers comply with relevant privacy standards.
    \end{itemize}

    \item \textbf{Maintenance and Updates:}
    \begin{itemize}
        \item \textbf{Model Updates:} Regularly update the model to incorporate new data and improve performance. Automate this process if possible to reduce manual effort.
        \item \textbf{System Maintenance:} Plan for regular maintenance of the infrastructure to prevent downtime and ensure smooth operation.
    \end{itemize}

    \item \textbf{Flexibility and Customisation:}
    \begin{itemize}
        \item \textbf{Fine-Tuning:} Allow for model fine-tuning to adapt the LLM to specific use cases and datasets. Fine-tuning can improve accuracy and relevance in responses.
        \item \textbf{API Integration:} Ensure the deployment platform supports easy integration with existing systems and workflows through APIs and SDKs.
    \end{itemize}

    \item \textbf{User Management:}
    \begin{itemize}
        \item \textbf{Access Control:} Implement role-based access control to manage who can deploy, use, and maintain the LLM.
        \item \textbf{Monitoring and Logging:} Set up comprehensive monitoring and logging to track usage, performance, and potential issues. This helps in proactive troubleshooting and optimisation.
    \end{itemize}

    \item \textbf{Compliance:}
    \begin{itemize}
        \item \textbf{Regulatory Compliance:} Ensure that the deployment adheres to all relevant regulatory and legal requirements, including data protection laws like GDPR, HIPAA, etc.
        \item \textbf{Ethical Considerations:} Implement ethical guidelines to avoid biases and ensure the responsible use of LLMs.
    \end{itemize}

    \item \textbf{Support and Documentation:}
    \begin{itemize}
        \item \textbf{Technical Support:} Choose a deployment platform that offers robust technical support and resources.
        \item \textbf{Documentation:} Provide comprehensive documentation for developers and users to facilitate smooth deployment and usage.
    \end{itemize}
\end{itemize}

\chapter{Stage 7: Monitoring and Maintenance}

\section{Steps Involved in Monitoring and Maintenance of Deployed Fine-Tuned LLMs}

Continuous monitoring and maintenance of fine-tuned LLMs are essential to ensure their optimal performance, accuracy, and security over time. Below are the key steps involved in this process:
 
\begin{enumerate}
    \item \textbf{Setup Initial Baselines:} Establish initial performance baselines by evaluating the model on a comprehensive test dataset. Record metrics such as accuracy, latency, throughput, and error rates to serve as reference points for future monitoring.
    
    \item \textbf{Performance Monitoring:} Implement systems to continuously track key performance metrics such as response time, server load, and token usage. Regularly compare these metrics against the established baselines to detect any deviations.

    \item \textbf{Accuracy Monitoring:} Continuously evaluate the model's predictions against a ground truth dataset. Use metrics like precision, recall, F1 score, and cross-entropy loss to ensure the model maintains high accuracy levels.

    \item \textbf{Error Monitoring:} Track and analyse errors, including runtime errors and prediction errors. Implement logging mechanisms to capture detailed information about each error for troubleshooting and improvement.

    \item \textbf{Log Analysis:} Maintain comprehensive logs for each prediction request and response, including input data, output predictions, response times, and encountered errors. Regularly review logs to identify patterns and areas for improvement.

    \item \textbf{Alerting Mechanisms:} Set up automated alerting systems to notify stakeholders of any anomalies or deviations from expected performance metrics. Integrate alerts with communication tools like Slack, PagerDuty, or email for timely responses.

    \item \textbf{Feedback Loop:} Establish a feedback loop with end-users to gather insights on model performance and user satisfaction. Use this feedback to continuously refine and improve the model.

    \item \textbf{Security Monitoring:} Implement robust security measures to monitor for threats, including unauthorised access, data breaches, and adversarial attacks. Use encryption, access control, and regular security audits to protect the model and data.

    \item \textbf{Drift Detection:} Continuously monitor for data and concept drift using statistical tests and drift detectors. Regularly evaluate the model on holdout datasets to detect changes in input data distribution or model performance.

    \item \textbf{Model Versioning:} Maintain version control for different iterations of the model. Track performance metrics for each version to ensure that the best-performing model is in production.

    \item \textbf{Documentation and Reporting:} Keep detailed documentation of monitoring procedures, metrics, and findings. Generate regular reports to provide stakeholders with insights into the model's performance and maintenance activities.
    
    \item \textbf{Periodic Review and Update:} Regularly assess and update the monitoring processes to incorporate new techniques, tools, and best practices, ensuring the monitoring system remains effective and up-to-date.
\end{enumerate}

\section{Continuous Monitoring of Model Performance}

While large language model (LLM) applications undergo some form of evaluation, continuous monitoring remains inadequately implemented in most cases. This section outlines the components necessary to establish an effective monitoring programme aimed at safeguarding users and preserving brand integrity.

\subsection{Functional Monitoring}

Initially, it is crucial to monitor fundamental metrics consistently. This includes tracking metrics such as request volume, response times, token utilisation, costs incurred, and error rates.

\subsection{Prompt Monitoring}

Following functional metrics, attention should be directed towards monitoring user-generated prompts or inputs. Metrics like readability can provide valuable insights. LLM evaluators should be employed to detect potential toxicity in responses. Additionally, metrics such as embedding distances from reference prompts prove insightful, ensuring adaptability to varying user interactions over time.

\noindent Introducing a new evaluation category involves identifying adversarial attempts or malicious prompt injections, often overlooked in initial evaluations. Comparison against reference sets of known adversarial prompts helps identify and flag malicious activities. Evaluative LLMs play a crucial role in classifying prompts as benign or malicious.

\subsection{Response Monitoring}

Monitoring responses involves several critical checks to ensure alignment with expected outcomes. Parameters such as relevance, coherence (hallucination), topical alignment, sentiment, and their evolution over time are essential. Metrics related to toxicity and harmful output require frequent monitoring due to their critical impact. Prompt leakage represents an adversarial tactic wherein sensitive prompt information is illicitly extracted from the application's stored data. Monitoring responses and comparing them against the database of prompt instructions can help detect such breaches. Embedding distance metrics are particularly effective in this regard. Regular testing against evaluation datasets provides benchmarks for accuracy and highlights any performance drift over time. Tools capable of managing embeddings allow exportation of underperforming output datasets for targeted improvements.

\subsection{Alerting Mechanisms and Thresholds}

Effective monitoring necessitates well-calibrated alerting thresholds to avoid excessive false alarms. Implementing multivariate drift detection and alerting mechanisms can enhance accuracy. Consideration of false alarm rates and best practices for setting thresholds is paramount for effective monitoring system design. Alerting features should include integration with communication tools such as Slack and PagerDuty. Some systems offer automated response blocking in case of alerts triggered by problematic prompts. Similar mechanisms can be employed to screen responses for personal identifiable information (PII), toxicity, and other quality metrics before delivery to users. Custom metrics tailored to specific application nuances or innovative insights from data scientists can significantly enhance monitoring efficacy. Flexibility to incorporate such metrics is essential to adapt to evolving monitoring needs and advancements in the field.

\subsection{Monitoring User Interface (UI)}

The monitoring system's UI is pivotal, typically featuring time-series graphs of monitored metrics. Differentiated UIs facilitate in-depth analysis of alert trends, aiding root cause analysis. Advanced UI capabilities may include visualisations of embedding spaces through clustering and projections, providing insights into data patterns and relationships. Mature monitoring systems categorise data by users, projects, and teams, ensuring role-based access control (RBAC) to protect sensitive information. Optimising alert analysis within the UI interface remains an area where improvements can significantly reduce false alarm rates and enhance operational efficiency.

\section{Updating LLM Knowledge}

To improve the knowledge base of an LLM, continued pretraining is used to help LLM evolve with the latest knowledge and information. The world and language are constantly evolving. New information emerges, trends shift, and cultural references change. LLMs trained on static data can become outdated, leading to:

\begin{itemize}
    \item \textbf{Factual Errors:} Outdated information can cause LLMs to provide inaccurate responses.
    \item \textbf{Irrelevance:} Models might miss the context of current events or use outdated references.
    \item \textbf{Bias Perpetuation:} Biases present in training data can become entrenched if not addressed through updates.
\end{itemize}

\subsection{Retraining Methods}

\begin{itemize}
    \item \textbf{Periodic Retraining:} This involves refreshing the model's knowledge base at regular intervals (weekly, monthly, yearly) with new data. This is a straightforward method but requires a steady stream of high-quality, unbiased data.
    
    \item \textbf{Trigger-Based Retraining:} This approach monitors the LLM's performance. When metrics like accuracy or relevance fall below a certain threshold, a retraining process is triggered. This method is more dynamic but requires robust monitoring systems and clear performance benchmarks.
\end{itemize}

\subsection{Additional Methods}

\begin{itemize}
    \item \textbf{Fine-Tuning:} LLMs can be fine-tuned for specific tasks by training them on smaller, domain-specific datasets. This allows for specialisation without complete retraining.

    \item \textbf{Active Learning:} This approach involves selectively querying the LLM to identify areas where it lacks knowledge. The retrieved information is then used to update the model.
\end{itemize}

\subsection{Key Considerations}

\begin{itemize}
    \item \textbf{Data Quality and Bias:} New training data must be carefully curated to ensure quality and mitigate bias. Techniques like human annotation and fairness checks are crucial.

    \item \textbf{Computational Cost:} Retraining LLMs can be computationally expensive, requiring significant resources. Optimisations like transfer learning (using pre-trained models as a starting point) can help reduce costs.

    \item \textbf{Downtime:} Retraining often takes time, leading to LLM downtime. Strategies like rolling updates or deploying multiple models can minimise service disruptions.

    \item \textbf{Version Control:} Tracking different versions of the LLM and their training data is essential for rollbacks in case of performance issues.
\end{itemize}

\section{The Future of LLM Updates}

Research is ongoing to develop more efficient and effective LLM update strategies. One promising area is \textit{continuous learning}, where LLMs can continuously learn and adapt from new data streams without retraining from scratch. Continuous learning aims to reduce the need for frequent full-scale retraining by enabling models to update incrementally with new information. This approach can significantly enhance the model's ability to remain current with evolving knowledge and language use, improving its long-term performance and relevance. \\

\noindent Innovations in transfer learning and meta-learning are also contributing to advancements in LLM updates. These techniques allow models to leverage pre-existing knowledge and adapt quickly to new tasks or domains with minimal additional training. By integrating these advanced learning methods, future LLMs can become more adaptable and efficient in processing and understanding new information.

\noindent Furthermore, ongoing improvements in hardware and computational resources will support more frequent and efficient updates. As processing power increases and becomes more accessible, the computational burden of updating large models will decrease, enabling more regular and comprehensive updates. \\

\noindent Collaboration between academia and industry is vital in driving these advancements. By sharing research findings and best practices, the field can collectively move towards more robust and efficient LLM update methodologies, ensuring that models remain accurate, relevant, and valuable over time.

\chapter{Industrial Fine-Tuning Platforms and Frameworks for LLMs}

The evolution of fine-tuning techniques has been propelled by leading tech companies and platforms that have introduced innovative frameworks and services. Companies like HuggingFace, Amazon Web Services (AWS), Microsoft Azure, and OpenAI have developed tools and platforms that simplify and democratise the fine-tuning process. These advancements have not only lowered the barrier to entry for leveraging state-of-the-art AI models but have also enabled a wide range of applications across various industries, from healthcare and finance to customer service and content creation. Each of these platforms offers unique capabilities that cater to different needs, whether it be through automated fine-tuning workflows, scalable cloud-based training environments, or accessible API interfaces for deploying custom models.\\

\noindent HuggingFace, for example, has made significant strides with its Transformers library\footnote{\url{https://huggingface.co/docs/transformers/en/index/}}
 and tools like Autotrain\footnote{\url{https://huggingface.co/autotrain}} and SetFit, which allow users to fine-tune models with minimal coding and data. Their platform provides a robust infrastructure that supports both the research community and industry practitioners, facilitating the rapid development and deployment of custom AI solutions. Similarly, AWS's SageMaker\footnote{\url{https://huggingface.co/autotrain}} and SetFit\footnote{\url{https://aws.amazon.com/sagemaker/}} provides an extensive suite of services that cover the entire machine learning lifecycle, from data preparation and training to model deployment and optimisation, making it a comprehensive solution for enterprise-level applications.\\

\noindent On the other hand, Microsoft Azure integrates its fine-tuning capabilities with enterprise-grade tools and services, offering solutions like Azure Machine Learning and the Azure OpenAI Service that cater to large organisations looking to incorporate advanced AI into their operations. Azure’s focus on MLOps and seamless integration with other Azure services ensures that fine-tuned models can be efficiently deployed and maintained in production environments. Meanwhile, OpenAI has pioneered the concept of "fine-tuning as a service" allowing businesses to leverage their powerful models like GPT-4 through a user-friendly API \footnote{\url{https://platform.openai.com/docs/guides/fine-tuning/fine-tuning-integrations}}, enabling custom model adaptations without the need for in-house AI expertise or infrastructure.\\

\noindent The collective efforts of these tech companies have not only enhanced the efficiency and scalability of fine-tuning but also democratised access to sophisticated AI tools. By reducing the technical barriers and providing comprehensive, user-friendly platforms, these innovations have enabled a wider range of industries to deploy advanced AI models tailored to their specific needs. Tables \ref{tab:compLLMfinetunetable1} and \ref{tab:compLLMfinetunetable2} offer a quick comparison of LLM fine-tuning tools and frameworks from different providers.

\begin{table}[H]
\centering
\footnotesize
\begin{tabularx}{\textwidth}{|>{\raggedright\arraybackslash}p{2.5cm}|X|X|X|X|X|}
\hline
\textbf{Parameter} & \textbf{NVIDIA NeMo} & \textbf{Hugging Face AutoTrain API} & \textbf{Amazon Bedrock} & \textbf{AWS SageMaker JumpStart} & \textbf{Hugging Face Trainer API} \\
\hline
\textbf{Primary Use Case} & Custom fine-tuning of LLMs with advanced NVIDIA GPUs. & Fine-tuning and deployment of LLMs with minimal code. & Fine-tuning and deploying LLMs on AWS infrastructure. & Simplified fine-tuning and deployment within the AWS ecosystem. & Manual fine-tuning of LLMs with detailed control over training processes. \\
\hline
\textbf{Model Support} & Supports a variety of large, pre-trained models, including Megatron series. & Supports a wide range of pre-trained models from the Hugging Face model hub. & Supports various LLMs like Amazon Titan and third-party models. & Pre-trained models from AWS and partners; integration with custom models. & Supports a vast array of models from the Hugging Face model hub. \\
\hline
\textbf{Data Handling} & Users provide task-specific data for fine-tuning, processed using NVIDIA's infrastructure. & Uploads datasets via a simple interface; AutoTrain handles preprocessing and model training. & Data is uploaded and managed within the AWS environment; integrates with AWS data services. & Uploads and processes data within AWS; supports various data formats. & Users manually preprocess data and manage training steps. \\
\hline
\textbf{Customisation Level} & High; extensive control over fine-tuning process and model parameters. & Moderate; automated process with some customisation options. & High; detailed configuration and integration with AWS services. & Moderate; pre-configured settings with some customisation available. & Very High; detailed control over every aspect of fine-tuning. \\
\hline
\textbf{Scalability} & High; leverages NVIDIA’s GPU capabilities for efficient scaling. & High; scalable via Hugging Face’s cloud infrastructure. & Very High; scalable across AWS's extensive cloud infrastructure. & High; scalable within the AWS cloud ecosystem. & High; scalability depends on the infrastructure used (e.g., local vs. cloud). \\
\hline
\textbf{Deployment Options} & On-premises or cloud deployment via NVIDIA infrastructure. & Deployed via Hugging Face's cloud or can be exported for local deployment. & Integrated into AWS services, easily deployable across AWS's global infrastructure. & AWS cloud deployment; integrates with other AWS services. & Deployable locally, in cloud, or exported to other platforms. \\
\hline
\textbf{Integration with Ecosystem} & Deep integration with NVIDIA tools (e.g., TensorRT) and GPU-based workflows. & Integrates well with the Hugging Face ecosystem and other ML tools. & Seamless integration with AWS services (e.g., S3, Lambda, SageMaker). & Strong integration with AWS services; easy to connect with data pipelines and analytics. & Integrates with Hugging Face ecosystem and other Python-based ML tools. \\
\hline
\textbf{Data Privacy} & Users must ensure data privacy compliance; NVIDIA handles data during processing. & Data handled within Hugging Face's environment; privacy depends on data handling practices. & Strong focus on data privacy within AWS environment; compliant with various standards. & Strong AWS privacy and security measures; compliant with industry standards. & User-managed; depends on where the models and data are hosted. \\
\hline
\textbf{Target Users} & Enterprises and developers needing advanced customisation and performance in LLM fine-tuning. & Developers and businesses looking for easy, automated LLM fine-tuning solutions. & Businesses and developers integrated into or seeking to leverage AWS cloud services. & Enterprises and developers seeking streamlined AI/ML solutions within AWS. & Researchers, developers, and ML engineers needing detailed control over training. \\
\hline
\textbf{Limitations} & High resource demand and potential costs; dependency on NVIDIA ecosystem. & Less control over fine-tuning specifics; cloud-based, may not suit all on-premises needs. & Dependency on AWS; potential vendor lock-in, cost management complexity. & Limited to AWS services; pre-configured options may limit deep customisation. & Requires technical expertise; more complex setup and management. \\
\hline
\end{tabularx}
\caption{Detailed Comparison of LLM Fine-Tuning Platforms (Part I). This table provides a comprehensive comparison of various fine-tuning tools for Large Language Models (LLMs), including NVIDIA NeMo, Hugging Face AutoTrain API, Amazon Bedrock, AWS SageMaker JumpStart, and Hugging Face Trainer API. It covers multiple aspects such as the primary use case, model support, data handling, customisation level, scalability, deployment options, integration with the ecosystem, data privacy, target users, and limitations for each tool.}
\label{tab:compLLMfinetunetable1}
\end{table}

\begin{table}[H]
\centering
\footnotesize
\begin{tabularx}{\textwidth}{|>{\raggedright\arraybackslash}p{2.5cm}|X|X|X|X|}
\hline
\textbf{Parameter} & \textbf{OpenAI Fine-Tuning API} & \textbf{Google Vertex AI Studio} & \textbf{Microsoft Azure AI Studio} & \textbf{LangChain} \\
\hline
\textbf{Primary Use Case} & API-based fine-tuning for OpenAI models with custom datasets. & End-to-end ML model development and deployment within Google Cloud. & End-to-end AI development, fine-tuning, and deployment on Azure. & Building applications using LLMs with modular and customisable workflows. \\
\hline
\textbf{Model Support} & Limited to OpenAI models like GPT-3 and GPT-4. & Supports Google’s pre-trained models and user-customised models. & Supports Microsoft’s models and custom models fine-tuned within Azure. & Supports integration with various LLMs and AI tools (e.g., OpenAI, GPT-4, Cohere). \\
\hline
\textbf{Data Handling} & Users upload datasets via API; OpenAI handles preprocessing and fine-tuning. & Data managed within Google Cloud; supports multiple data formats. & Data integrated within Azure ecosystem; supports various formats and sources. & Data handling is flexible, dependent on the specific LLM and integration used. \\
\hline
\textbf{Customisation Level} & Moderate; focuses on ease of use with limited deep customisation. & High; offers custom model training and deployment with detailed configuration. & High; extensive customisation options through Azure’s AI tools. & Very High; allows detailed customisation of workflows, models, and data processing. \\
\hline
\textbf{Scalability} & High; scalable through OpenAI's cloud infrastructure. & Very High; leverages Google Cloud’s infrastructure for scaling. & Very High; scalable across Azure’s global infrastructure. & High; scalability depends on the specific infrastructure and models used. \\
\hline
\textbf{Deployment Options} & Deployed via API, integrated into applications using OpenAI's cloud. & Deployed within Google Cloud; integrates with other GCP services. & Deployed within Azure; integrates with Azure’s suite of services. & Deployed within custom infrastructure; integrates with various cloud and on-premises services. \\
\hline
\textbf{Integration with Ecosystem} & Limited to OpenAI ecosystem; integrates well with apps via API. & Seamless integration with Google Cloud services (e.g., BigQuery, AutoML). & Deep integration with Azure’s services (e.g., Data Factory, Power BI). & Flexible integration with multiple tools, APIs, and data sources. \\
\hline
\textbf{Data Privacy} & Managed by OpenAI; users must consider data transfer and privacy implications. & Strong privacy and security measures within Google Cloud environment. & Strong privacy and security measures within Azure environment. & Dependent on the integrations and infrastructure used; users manage privacy. \\
\hline
\textbf{Target Users} & Developers and enterprises looking for straightforward, API-based LLM fine-tuning. & Developers and businesses integrated into Google Cloud or seeking to leverage GCP. & Enterprises and developers integrated into Azure or seeking to leverage Azure’s AI tools. & Developers needing to build complex, modular LLM-based applications with custom workflows. \\
\hline
\textbf{Limitations} & Limited customisation; dependency on OpenAI’s infrastructure; potential cost. & Limited to Google Cloud ecosystem; potential cost and vendor lock-in. & Limited to Azure ecosystem; potential cost and vendor lock-in. & Complexity in chaining multiple models and data sources; requires more setup. \\
\hline
\end{tabularx}
\caption{Detailed Comparison of LLM Fine-Tuning Platforms (Part II). This table continues the comparison of LLM fine-tuning tools, focusing on OpenAI Fine-Tuning API, Google Vertex AI Studio, Microsoft Azure AI Studio, and LangChain. It evaluates the tools based on the primary use case, model support, data handling, customisation level, scalability, deployment options, integration with the ecosystem, data privacy, target users, and limitations, offering a complete view of their capabilities and constraints.}
\label{tab:compLLMfinetunetable2}
\end{table}

\section{Autotrain}

Autotrain is HuggingFace’s innovative platform that automates the fine-tuning of large language models, making it accessible even to those with limited machine learning expertise. The complexity and resource demands of fine-tuning LLMs can be daunting, but Autotrain simplifies the process by handling the most challenging aspects, such as data preparation, model configuration, and hyperparameter optimisation. This automation is particularly valuable for small teams or individual developers who need to deploy custom LLMs quickly and efficiently.

\subsection{Steps Involved in Fine-Tuning Using Autotrain}

\begin{figure}[h!]
    \centering
    \includegraphics[width=\textwidth]{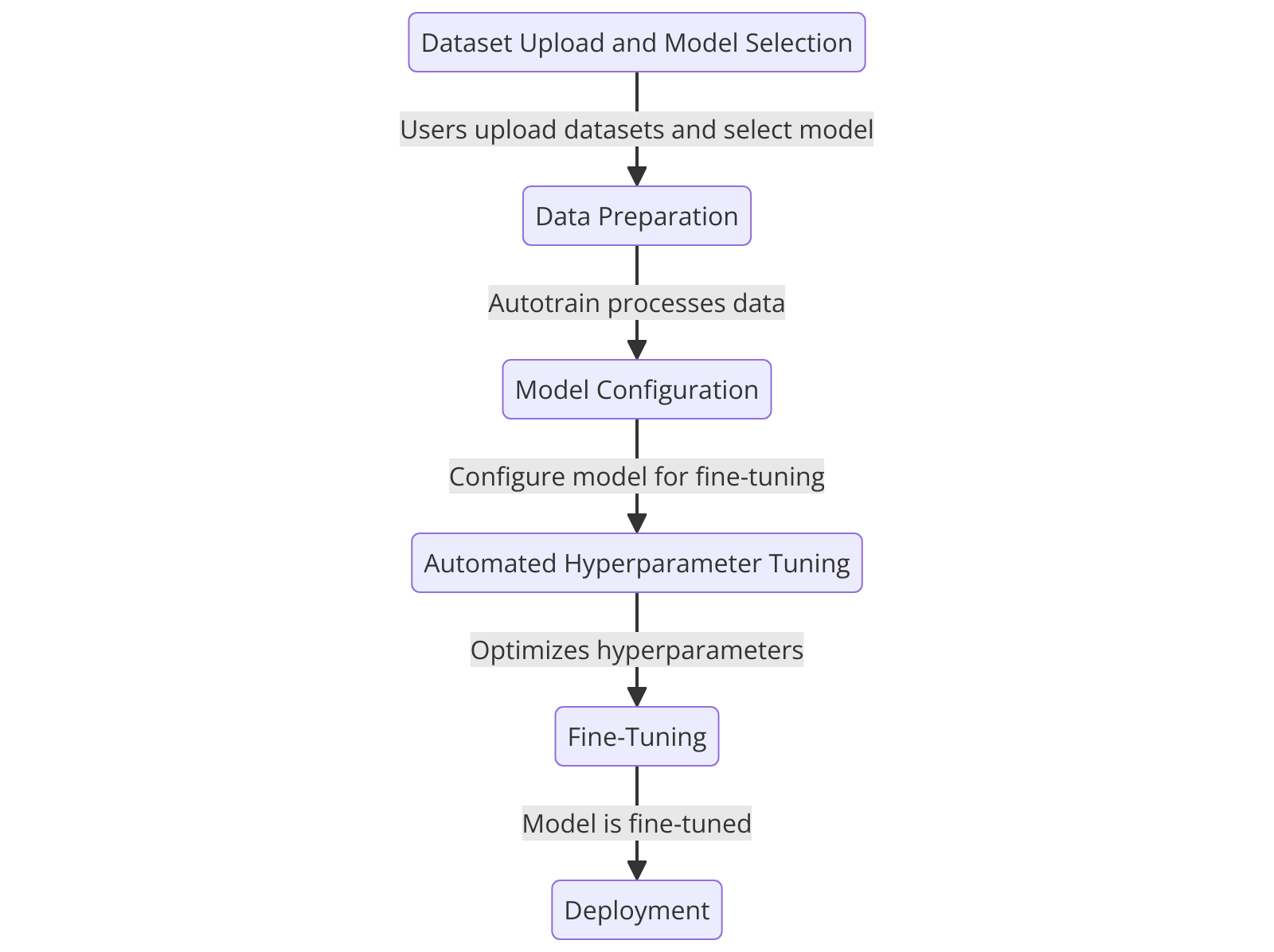} 
    \caption{Overview of the Autotrain Workflow. This diagram illustrates the step-by-step process within the Autotrain system, beginning with the upload of datasets and model selection by users. The workflow then moves to data preparation and model configuration, followed by automated hyperparameter tuning to optimise model performance. The fine-tuning phase adjusts the model based on the provided datasets, culminating in the deployment of the fully fine-tuned model for practical use.}
    \label{fig:autotrain_workflow}
\end{figure}

Following are the steps involved in fine-tuning LLMs using Autotrain. Figure \ref{fig:autotrain_workflow} represents the visual workflow.

\begin{itemize}
    \item \textbf{Dataset Upload and Model Selection:}
    \begin{itemize}
        \item Users begin by uploading their datasets to the Autotrain platform.
        \item They then select a pre-trained model from the extensive HuggingFace Model Hub.
    \end{itemize}
    \item \textbf{Data Preparation:}
    \begin{itemize}
        \item Autotrain automatically processes the uploaded data, including tasks like tokenization to convert text into a format the LLM can understand.
    \end{itemize}
    \item \textbf{Model Configuration:}
    \begin{itemize}
        \item The platform configures the model for fine-tuning, setting up the training environment and necessary parameters.
    \end{itemize}
    \item \textbf{Automated Hyperparameter Tuning:}
    \begin{itemize}
        \item Autotrain explores various hyperparameter configurations (such as learning rate, batch size, and sequence length) and selects the best-performing ones.
    \end{itemize}
    \item \textbf{Fine-Tuning:}
    \begin{itemize}
        \item The model is fine-tuned on the prepared data with the optimised hyperparameters.
    \end{itemize}
    \item \textbf{Deployment:}
    \begin{itemize}
        \item Once fine-tuning is complete, the model is ready for deployment in various NLP applications, such as text generation, completion, and language translation.
    \end{itemize}
\end{itemize}

\subsection{Best Practices of Using Autotrain}

\begin{itemize}
    \item \textbf{Data Quality:} Ensure high-quality, well-labelled data for better model performance.
    \item \textbf{Model Selection:} Choose pre-trained models that are well-suited to your specific task to minimize fine-tuning effort.
    \item \textbf{Hyperparameter Optimisation:} Leverage Autotrain’s automated hyperparameter tuning to achieve optimal performance without manual intervention.
\end{itemize}

\subsection{Challenges of Using Autotrain}

\begin{itemize}
    \item \textbf{Data Privacy:} Ensuring the privacy and security of sensitive data during the fine-tuning process.
    \item \textbf{Resource Constraints:} Managing computational resources effectively, especially in environments with limited access to powerful hardware.
    \item \textbf{Model Overfitting:} Avoiding overfitting by ensuring diverse and representative training data and using appropriate regularization techniques.
\end{itemize}

\subsection{When to Use Autotrain}

\begin{enumerate}
    \item \textbf{Lack of Deep Technical Expertise:} Ideal for individuals or small teams without extensive machine learning or LLM background who need to fine-tune models quickly and effectively.
    \item \textbf{Quick Prototyping and Deployment:} Suitable for rapid development cycles where time is critical, such as proof-of-concept projects or MVPs.
    \item \textbf{Resource-Constrained Environments:} Useful for scenarios with limited computational resources or where a quick turnaround is necessary.
\end{enumerate}

\noindent In summary, Autotrain is an excellent tool for quick, user-friendly fine-tuning of LLMs for standard NLP tasks, especially in environments with limited resources or expertise. However, it may not be suitable for highly specialised applications or those requiring significant customisation and scalability.

\subsection{Tutorials}

\begin{enumerate}
    \item \href{https://cobusgreyling.medium.com/how-to-create-huggingface-custom-ai-models-using-autotrain-72d75484b82b}{\color{blue} How To Create HuggingFace Custom AI Models Using AutoTrain}
    \item \href{https://www.kdnuggets.com/how-to-finetune-mistral-ai-7b-llm-with-hugging-face-autotrain}{\color{blue} Finetune models with HuggingFace AutoTrain}
\end{enumerate}

\section{Transformers Library and Trainer API}

The Transformers Library by HuggingFace stands out as a pivotal tool for fine-tuning large language models (LLMs) such as BERT, GPT-3, and GPT-4. This comprehensive library offers a wide array of pre-trained models tailored for various LLM tasks, making it easier for users to adapt these models to specific needs with minimal effort. Whether you're fine-tuning for tasks like sentiment analysis, text classification, or generating customer support responses, the library simplifies the process by allowing seamless model selection from the HuggingFace Model Hub and straightforward customisation through its high-level APIs.\\

\noindent Central to the fine-tuning process within the Transformers Library is the Trainer API. This API includes the Trainer class, which automates and manages the complexities of fine-tuning LLMs. After completing data preprocessing, the Trainer class streamlines the setup for model training, including data handling, optimisation, and evaluation. Users only need to configure a few parameters, such as learning rate and batch size, and the API takes care of the rest. However, it's crucial to note that running Trainer.train() can be resource-intensive and slow on a CPU. For efficient training, a GPU or TPU is recommended. Platforms like Google Colab provide free access to these resources, making it feasible for users without high-end hardware to fine-tune models effectively.\\

\noindent The Trainer API also supports advanced features like distributed training and mixed precision, which are essential for handling the large-scale computations required by modern LLMs. Distributed training allows the fine-tuning process to be scaled across multiple GPUs or nodes, significantly reducing training time. Mixed precision training, on the other hand, optimises memory usage and computation speed by using lower precision arithmetic without compromising model performance. HuggingFace’s dedication to accessibility is evident in the extensive documentation and community support they offer, enabling users of all expertise levels to fine-tune LLMs. This democratisation of advanced NLP technology empowers developers and researchers to deploy sophisticated, fine-tuned models for a wide range of applications, from specialised language understanding to large-scale data processing.

\subsection{Limitations of the Transformers Library and Trainer API}

\begin{itemize}
    \item \textbf{Limited Customisation for Advanced Users:} While the Trainer API simplifies many aspects of training, it might not offer the deep customisation that advanced users or researchers might need for novel or highly specialised applications.
    \item \textbf{Learning Curve:} Despite the simplified API, there is still a learning curve associated with understanding and effectively using the Transformers Library and Trainer API, particularly for those new to NLP and LLM.
    \item \textbf{Integration Limitations:} The seamless integration and ease of use are often tied to the HuggingFace ecosystem, which might not be compatible with all workflows or platforms outside their environment.
\end{itemize}

\noindent In summary, the Transformers Library and Trainer API provide robust, scalable solutions for fine-tuning LLMs across a range of applications, offering ease of use and efficient training capabilities. However, users must be mindful of the resource requirements and potential limitations in customisation and complexity management.

\section{Optimum: Enhancing LLM Deployment Efficiency}

Optimum\footnote{\url{https://huggingface.co/docs/optimum/en/index}} is HuggingFace’s tool designed to optimise the deployment of large language models (LLMs) by enhancing their efficiency across various hardware platforms. As LLMs grow in size and complexity, deploying them in a cost-effective and performant manner becomes increasingly challenging. Optimum addresses these challenges by applying a range of hardware-specific optimisations, such as quantisation, pruning, and model distillation, which reduce the model’s size and improve inference speed without significantly affecting accuracy. The following are the key techniques supported by Optimum:

\begin{itemize}
    \item \textbf{Quantisation:} Quantisation is one of the key techniques supported by Optimum. This process involves converting the model’s weights from high-precision floating-point numbers to lower-precision formats, such as int8 or float16. This reduction in precision decreases the model’s memory footprint and computational requirements, enabling faster execution and lower power consumption, especially on edge devices and mobile platforms. Optimum automates the quantisation process, making it accessible to users who may not have expertise in low-level hardware optimisation.

    \item \textbf{Pruning:} Pruning is another critical optimisation strategy offered by Optimum. It involves identifying and removing less significant weights from the LLM, reducing its overall complexity and size. This leads to faster inference times and lower storage needs, which are particularly beneficial for deploying models in environments with limited computational resources. Optimum’s pruning algorithms carefully eliminate these redundant weights while maintaining the model’s performance, ensuring that it continues to deliver high-quality results even after optimisation.

    \item \textbf{Model Distillation:} In addition to these techniques, Optimum supports model distillation, a process where a smaller, more efficient model is trained to replicate the behaviour of a larger, more complex model. This distilled model retains much of the knowledge and capabilities of the original while being significantly lighter and faster. Optimum provides tools to facilitate the distillation process, allowing users to create compact LLMs that are well-suited for real-time applications. By offering a comprehensive suite of optimisation tools, Optimum ensures that HuggingFace’s LLMs can be deployed effectively across a wide range of environments, from powerful cloud servers to resource-constrained edge devices.
\end{itemize}

\subsection{Best Practices of Using Optimum}

\begin{itemize}
    \item \textbf{Understand Hardware Requirements:} Assess the target deployment environment (e.g., edge devices, cloud servers) to optimise model configuration accordingly.
    \item \textbf{Iterative Optimisation:} Experiment with different optimisation techniques (quantisation levels, pruning thresholds) to find the optimal balance between model size, speed, and accuracy.
    \item \textbf{Validation and Testing:} Validate optimised models thoroughly to ensure they meet performance and accuracy requirements across different use cases.
    \item \textbf{Documentation and Support:} Refer to HuggingFace's resources for detailed guidance on using Optimum's tools effectively, and leverage community support for troubleshooting and best practices sharing.
    \item \textbf{Continuous Monitoring:} Monitor deployed models post-optimisation to detect any performance degradation and adjust optimisation strategies as needed to maintain optimal performance over time.
\end{itemize}

\subsection{Tutorials}

\begin{enumerate}
    \item \href{https://www.datacamp.com/tutorial/an-introduction-to-using-transformers-and-hugging-face}{\color{blue} An Introduction to Using Transformers and Hugging Face}
\end{enumerate}

\section{Amazon SageMaker JumpStart}

\begin{figure}[!ht]
    \centering
    \includegraphics[width=0.5\textwidth]{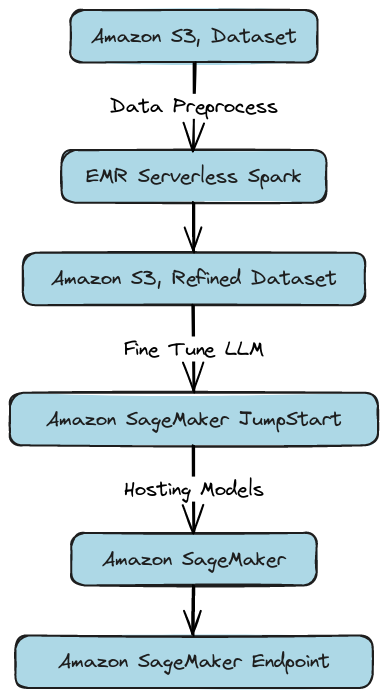}
    \caption{A step-by-step workflow illustrating the Amazon SageMaker JumpStart process, starting from data preprocessing using EMR Serverless Spark to the fine-tuning of LLMs, and ending with model deployment on Amazon SageMaker Endpoints. (adapted from \cite{amazonPreprocessFinetune})}
    \label{fig:amazon-jumpstart}
\end{figure}

Amazon SageMaker JumpStart is a feature within the SageMaker ecosystem designed to simplify and expedite the fine-tuning of large language models (LLMs). It provides users with a rich library of pre-built models and solutions that can be quickly customised for various use cases. This tool is particularly valuable for organisations looking to deploy NLP solutions efficiently without deep expertise in machine learning or the extensive computational resources typically required for training LLMs from scratch. The architecture depicted in Figure~\ref{fig:amazon-jumpstart} outlines a comprehensive pipeline for the fine-tuning and deployment of large language models (LLMs) Utilising AWS services.

\subsection{Steps Involved in Using JumpStart}

\begin{itemize}
    \item \textbf{Data Preparation and Preprocessing:}
    \begin{itemize}
        \item \textbf{Data Storage:} Begin by securely storing raw datasets in Amazon S3, AWS's scalable object storage service.
        \item \textbf{Preprocessing:} Utilise the EMR Serverless framework with Apache Spark for efficient data preprocessing. This step refines and prepares the raw data for subsequent model training and evaluation.
        \item \textbf{Data Refinement:} Store the processed dataset back into Amazon S3 after preprocessing, ensuring accessibility and readiness for the next stages.
    \end{itemize}
    \item \textbf{Model Fine-Tuning with SageMaker JumpStart:}
    \begin{itemize}
        \item \textbf{Model Selection:} Choose from a variety of pre-built models and solutions available through SageMaker JumpStart's extensive library, tailored for tasks such as sentiment analysis, text generation, or customer support automation.
        \item \textbf{Fine-Tuning Execution:} Utilise Amazon SageMaker's capabilities, integrated with SageMaker JumpStart, to fine-tune the selected model. This involves adjusting parameters and configurations to optimise the model's performance for specific use cases.
        \item \textbf{Workflow Simplification:} Leverage pre-built algorithms and model templates provided by SageMaker JumpStart to streamline the fine-tuning workflow, reducing the time and effort required for deployment.
    \end{itemize}
    \item \textbf{Model Deployment and Hosting:}
    \begin{itemize}
        \item \textbf{Deployment Setup:} Deploy the fine-tuned model using Amazon SageMaker's endpoint deployment capabilities. This setup ensures that the model is hosted in a scalable environment capable of handling real-time predictions efficiently.
        \item \textbf{Scalability:} Benefit from AWS's infrastructure scalability, allowing seamless scaling of resources to accommodate varying workloads and operational demands.
        \item \textbf{Efficiency and Accessibility:} Ensure that the deployed model is accessible via SageMaker endpoints, enabling efficient integration into production applications for real-time inference tasks.
    \end{itemize}
\end{itemize}

\subsection{Best Practices for Using JumpStart}

\begin{itemize}
    \item \textbf{Robust Data Management:} Maintain secure and organised data storage practices in Amazon S3, facilitating efficient data access and management throughout the pipeline.
    \item \textbf{Cost-Effective Processing:} Utilise serverless computing frameworks like EMR Serverless with Apache Spark for cost-effective and scalable data preprocessing.
    \item \textbf{Optimised Fine-Tuning:} Capitalise on SageMaker JumpStart's pre-built models and algorithms to expedite and optimise the fine-tuning process, ensuring optimal model performance without extensive manual configuration.
    \item \textbf{Continuous Monitoring and Optimisation:} Implement robust monitoring mechanisms post-deployment to track model performance metrics. This allows for timely optimisations and adjustments to maintain accuracy and efficiency over time.
    \item \textbf{Integration with AWS Services:} Leverage AWS's comprehensive suite of services and integration capabilities to create end-to-end pipelines that ensure reliable and scalable deployment of large-scale language models across diverse operational environments.
\end{itemize}

\subsection{Limitations of Using JumpStart}

\begin{itemize}
    \item \textbf{Limited Customisation:} While JumpStart simplifies the process for common use cases, it may offer limited flexibility for highly specialised or complex applications that require significant customisation beyond the provided templates and workflows.
    \item \textbf{Dependency on AWS Ecosystem:} JumpStart is tightly integrated with AWS services, which may pose challenges for users who prefer or need to operate in multi-cloud environments or those with existing infrastructure outside of AWS.
    \item \textbf{Resource Costs:} Utilising SageMaker’s scalable resources for fine-tuning LLMs, especially large models, can incur substantial costs, which might be a barrier for smaller organisations or those with limited budgets.
\end{itemize}

\subsection{Tutorials}

\begin{enumerate}
    \item \href{https://www.linkedin.com/pulse/fine-tuning-llama-2-amazon-sagemaker-jumpstart-elhousieny-phd%E1%B4%AC%E1%B4%AE%E1%B4%B0-8zp9c/}{\color{blue} Fine-Tuning LLaMA 2 with Amazon SageMaker JumpStart}
    \item \href{https://aws.amazon.com/blogs/machine-learning/learn-how-to-build-and-deploy-tool-using-llm-agents-using-aws-sagemaker-jumpstart-foundation-models/}{\color{blue} LLM Agents Using AWS SageMaker JumpStart Foundation Models}
\end{enumerate}

\section{Amazon Bedrock}

Amazon Bedrock\footnote{\url{https://aws.amazon.com/bedrock/}} is a fully managed service designed to simplify access to high-performing foundation models (FMs) from top AI innovators like AI21 Labs, Anthropic, Cohere, Meta, Mistral AI, Stability AI, and Amazon. It provides a unified API that integrates these models and offers extensive capabilities for developing secure, private, and responsible generative AI applications. With Amazon Bedrock, users can effortlessly experiment with and assess leading FMs tailored to their specific needs. The service supports private customisation of models through fine-tuning and Retrieval Augmented Generation (RAG), enabling the creation of intelligent agents that leverage enterprise data and systems. Amazon Bedrock's serverless architecture allows for quick deployment, seamless integration, and secure customisation of FMs without the burden of infrastructure management, Utilising AWS tools to deploy these models into applications efficiently and securely.

\subsection{Steps Involved in Using Amazon Bedrock}

Amazon Bedrock offers a streamlined workflow for deploying and fine-tuning LLMs, making it an ideal choice for businesses looking to quickly integrate advanced AI capabilities into their operations. Here’s a high-level overview of how Bedrock operates:

\begin{itemize}
    \item \textbf{Model Selection:} Users start by choosing from a curated selection of foundation models available through Bedrock. These include models from AWS (like Amazon Titan) and third-party providers (such as Anthropic Claude and Stability AI).
    \item \textbf{Fine-Tuning:}
    \begin{itemize}
        \item Once a model is selected, users can fine-tune it to better fit their specific needs. This involves feeding the model with domain-specific data or task-specific instructions to tailor its outputs.
        \item The fine-tuning process is handled via simple API calls, eliminating the need for extensive setup or detailed configuration. Users provide their custom data, and Bedrock manages the training process in the background.
    \end{itemize}
    \item \textbf{Deployment:}
    \begin{itemize}
        \item After fine-tuning, Bedrock takes care of deploying the model in a scalable and efficient manner. This means that users can quickly integrate the fine-tuned model into their applications or services.
        \item Bedrock ensures that the model scales according to demand and handles performance optimisation, providing a seamless user experience.
    \end{itemize}
    \item \textbf{Integration and Monitoring:}
    \begin{itemize}
        \item Bedrock integrates smoothly with other AWS services, allowing users to embed AI capabilities directly into their existing AWS ecosystem.
        \item Users can monitor and manage the performance of their deployed models through AWS’s comprehensive monitoring tools, ensuring that the models continue to perform optimally.
    \end{itemize}
\end{itemize}

\subsection{Limitations of Using Amazon Bedrock}

While Amazon Bedrock offers a robust suite of tools and services for addressing certain AI challenges, it is not a comprehensive solution for all AI needs. One key limitation is that it does not eliminate the requirement for human expertise. Organisations still need skilled professionals who understand the intricacies of AI technology to effectively develop, fine-tune, and optimise the models provided by Bedrock.\\

\noindent Additionally, Amazon Bedrock is not designed to function as a standalone service. It relies on integration with other AWS services, such as Amazon S3 for data storage, AWS Lambda for serverless computing, and AWS SageMaker for machine learning model development. Therefore, businesses leveraging Amazon Bedrock will also need to use these complementary AWS services to fully realise its potential. This interconnectedness means that while Amazon Bedrock enhances the AI capabilities within an AWS ecosystem, it may present a steep learning curve and require significant infrastructure management for those new to AWS.

\subsection{Tutorials}

\begin{enumerate}
    \item \href{https://medium.com/@abdullahiolaoye4/finetuning-llms-on-amazon-bedrock-887ebc547adc}{\color{blue} Finetuning LLMs on Amazon Bedrock}
    \item \href{https://cloudnature.net/blog/the-complete-guide-to-amazon-bedrock-for-generative-ai}{\color{blue} Amazon Bedrock for Generative AI}
\end{enumerate}

\section{OpenAI’s Fine-Tuning API}

OpenAI’s Fine-Tuning API is a comprehensive platform that facilitates the customisation of OpenAI’s pre-trained LLMs to cater to specific tasks and domains. This service is designed to be user-friendly, enabling a broad range of users, from businesses to individual developers, to harness the power of advanced AI without the complexities typically associated with model training and deployment.

\subsection{Steps Involved in Using OpenAI's Fine-Tuning API}

\begin{itemize}
    \item \textbf{Model Selection:}
    \begin{itemize}
        \item \textbf{Choosing a Pre-Trained Model:} Users begin by selecting a base model from OpenAI’s extensive lineup. This includes powerful models like GPT-4, which offer a robust starting point for a wide range of language processing tasks.
        \item \textbf{Customisable Base:} These models come pre-trained with vast amounts of data, providing a solid foundation that can be further refined to suit specific requirements.
    \end{itemize}
    \item \textbf{Data Preparation and Upload:}
    \begin{itemize}
        \item \textbf{Curating Relevant Data:} Users need to gather and prepare a dataset that reflects the specific task or domain they wish to fine-tune the model for. This data is crucial for teaching the model to perform the desired function more effectively.
        \item \textbf{Uploading Data to the API:} The Fine-Tuning API facilitates easy data upload. Users can feed their curated datasets into the API through straightforward commands, making the process accessible even to those with limited technical backgrounds.
    \end{itemize}
    \item \textbf{Initiating Fine-Tuning:}
    \begin{itemize}
        \item \textbf{Automated Process:} Once the data is uploaded, OpenAI’s infrastructure handles the fine-tuning process. The API adjusts the model’s parameters based on the new data to improve performance on the specified tasks.
    \end{itemize}
    \item \textbf{Deploying the Fine-Tuned Model:}
    \begin{itemize}
        \item \textbf{API Integration:} The fine-tuned model can be accessed and deployed via OpenAI’s API. This allows for seamless integration into various applications, such as chatbots, automated content creation tools, or specialised customer service systems.
    \end{itemize}
\end{itemize}

\subsection{Limitations of OpenAI’s Fine-Tuning API}

\begin{itemize}
    \item \textbf{Pricing Models:} Fine-tuning and using OpenAI’s models through the API can be costly, especially for large-scale deployments or continuous usage. This can be a significant consideration for smaller organisations or budget-constrained projects.
    \item \textbf{Data Privacy and Security:} Users must upload their data to OpenAI’s servers for the fine-tuning process. This raises potential concerns about data privacy and the security of sensitive or proprietary information.
    \item \textbf{Dependency on OpenAI Infrastructure:} The reliance on OpenAI’s infrastructure for model hosting and API access can lead to vendor lock-in, limiting flexibility and control over the deployment environment.
    \item \textbf{Limited Control Over Training Process:} The fine-tuning process is largely automated and managed by OpenAI, offering limited visibility and control over the specific adjustments made to the model.
\end{itemize}

\subsection{Tutorials}

\begin{enumerate}
    \item \href{https://www.datacamp.com/tutorial/fine-tuning-gpt-3-using-the-open-ai-api-and-python}{\color{blue} Fine-Tuning GPT-3 Using the OpenAI API}
\end{enumerate}

\section{NVIDIA NeMo Customizer}
NVIDIA NeMo Customiser\footnote{\url{https://developer.nvidia.com/blog/fine-tune-and-align-llms-easily-with-nvidia-nemo-customizer/}} is part of the NeMo framework, a suite of tools and models designed by NVIDIA to facilitate the development and fine-tuning of LLM models. The Customiser focuses specifically on making it easier to fine-tune large language models (LLMs) for specialised tasks and domains. Like other fine-tuning tools, NeMo Customiser is geared toward users who want to adapt pre-trained models for specific applications, such as conversational AI, translation, or domain-specific text generation. It delivers enterprise-ready models by offering accurate data curation, extensive customisation options, retrieval-augmented generation (RAG), and improved performance features. The platform supports training and deploying generative AI models across diverse environments, including cloud, data center, and edge locations. It provides a comprehensive package with support, security, and reliable APIs as part of the NVIDIA AI Enterprise.

\subsection{Key Features of NVIDIA NeMo}
NVIDIA NeMo is designed to enhance AI projects with several standout features.\cite{runNVIDIANeMo}
\begin{itemize}
    \item \textbf{State-of-the-Art Training Techniques}
    NeMo employs GPU-accelerated tools like NeMo Curator for preparing large-scale, high-quality datasets. These tools facilitate efficient pretraining of generative AI models by leveraging thousands of compute cores, which significantly reduces training time and enhances the accuracy of large language models (LLMs).
    
    \item \textbf{Advanced Customisation for LLMs}
    The NeMo Customiser microservice allows for precise fine-tuning and alignment of LLMs for specific domains. It uses model parallelism to speed up training and supports scaling across multiple GPUs and nodes, enabling the fine-tuning of larger models.
    
    \item \textbf{Optimised AI Inference with NVIDIA Triton}
    NeMo includes NVIDIA Triton Inference Server to streamline AI inference at scale. This integration accelerates generative AI inference, ensuring confident deployment of AI applications both on-premises and in the cloud.
    
    \item \textbf{User-Friendly Tools for Generative AI}
    NeMo features a modular, reusable architecture that simplifies the development of conversational AI models. It supports comprehensive workflows from data processing to deployment and includes pre-trained models for automatic speech recognition (ASR), natural language processing (NLP), and text-to-speech (TTS), which can be fine-tuned or used as-is.
    
    \item \textbf{Best-in-Class Pretrained Models}
    NeMo Collections offer a variety of pre-trained models and training scripts, facilitating rapid application development or fine-tuning for specific tasks. Currently, NeMo supports models like Llama 2, Stable Diffusion, and NVIDIA’s Nemotron-3 8B family.
    
    \item \textbf{Optimised Retrieval-Augmented Generation}
    NeMo Retriever delivers high-performance, low-latency information retrieval, enhancing generative AI applications with enterprise-grade retrieval-augmented generation (RAG) capabilities. This feature supports real-time business insights and data Utilisation.

\end{itemize}

\subsection{Components of NVIDIA NeMo}
\begin{itemize}
    \item \textbf{NeMo Core} Provides essential elements like the Neural Module Factory for training and inference, streamlining the development of conversational AI models.
    
    \item \textbf{NeMo Collections} Offers specialised modules and models for ASR, NLP, and TTS, including pre-trained models and training scripts, making the platform versatile.
    
    \item \textbf{Neural Modules} Serve as the building blocks of NeMo, defining trainable components such as encoders and decoders, which can be connected to create comprehensive models.
    
    \item \textbf{Application Scripts} Simplify the deployment of conversational AI models with ready-to-use scripts, enabling quick training or fine-tuning on specific datasets for various AI applications.
\end{itemize}

\subsection{Customising Large Language Models (LLMs)}
\begin{figure}[!ht]
    \centering
    \includegraphics[width=1\textwidth]{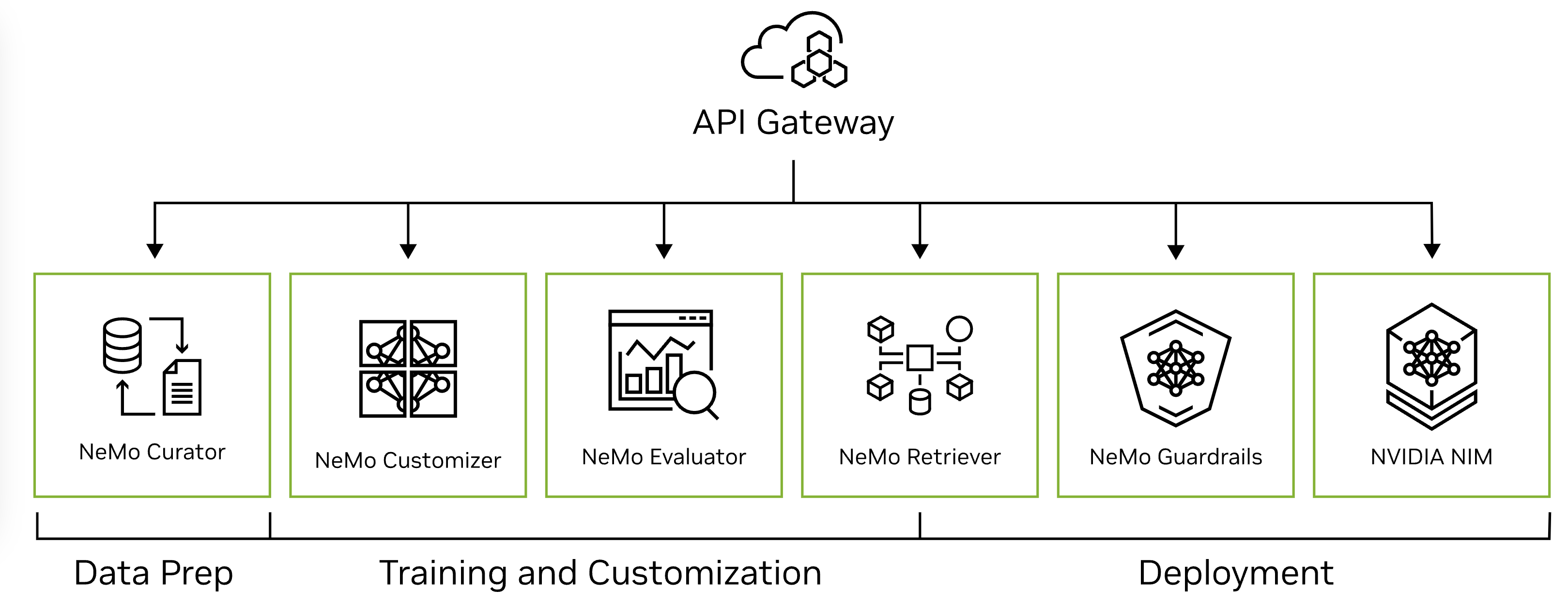}
    \caption{Nvidia NeMo Framework for Customising and Deploying LLMs. The Nvidia NeMo framework is designed for end-to-end customisation and deployment of large language models (LLMs). This diagram illustrates the process from data curation and distributed training of foundation models, through model customisation, to accelerated inference with guardrails. The platform enables AI developers to integrate in-domain, secure, and cited responses into enterprise applications, ensuring that LLMs are effectively tailored for specific tasks and industries. The NeMo framework, supported by Nvidia AI Enterprise, also offers robust support for various pre-trained foundation models like OpenAI's GPT family, ensuring scalability and reliability in AI deployments. (adapted from \cite{substackWhatTakes})}
    \label{fig:Nvidia nemo}
\end{figure}

While general-purpose LLMs, enhanced with prompt engineering or light fine-tuning, have enabled organisations to achieve successful proof-of-concept projects, transitioning to production presents additional challenges. Figure \ref{fig:Nvidia nemo} illustrates NVIDIA’s detailed LLM customisation lifecycle, offering valuable guidance for organisations that are preparing to deploy customised models in a production environment \cite{substackWhatTakes}.

\begin{enumerate}
    \item \textbf{Model Selection or Development} \\
    NVIDIA provides a range of pre-trained models, from 8B to 43B parameters, and supports the integration of other open-source models of any size. Alternatively, users can develop their own models, starting with data curation, which includes selecting, labeling, cleansing, validating, and integrating data. This process, better termed data engineering, involves additional analysis, designing storage, evaluating model training results, and incorporating reinforcement learning with human feedback (RLHF). While building a custom foundation model is often costly, complex, and time-consuming, most enterprises opt to start with a pre-trained model and focus on customisation.

    \item \textbf{Model Customisation} \\
    Model customisation involves optimising performance with task-specific datasets and adjusting model weights. NeMo offers recipes for customisation, and enterprises can choose models already tailored to specific tasks and then fine-tune them with proprietary data.

    \item \textbf{Inference} \\
    Inference refers to running models based on user queries. This phase involves considering hardware, architecture, and performance factors that significantly impact usability and cost in production.

    \item \textbf{Guardrails} \\
    NVIDIA employs guardrails as intermediary services between models and applications. These services review incoming prompts for policy compliance, execute arbitration or orchestration steps, and ensure model responses adhere to policies. Guardrails help maintain relevance, accuracy, safety, privacy, and security.

    \item \textbf{Applications} \\
    NVIDIA's framework presents enterprise applications as LLM-ready, though this is not always the case. Existing applications may be connected to LLMs to enable new features. However, creating assistants for knowledge access or task execution often involves designing new applications specifically for natural language interfaces.
\end{enumerate}

\subsection{Tutorials}

\begin{enumerate}
    \item \href{https://medium.com/@khang.pham.exxact/introduction-to-nvidia-nemo-tutorial-example-478f6ba6b160}{\color{blue} Introduction to NVIDIA NeMo — Tutorial and Example}
    \item \href{https://docs.nvidia.com/deeplearning/riva/user-guide/docs/tutorials/nmt-python-advanced-finetune-nmt-model-with-nemo.html}{\color{blue} How to fine-tune a Riva NMT Bilingual model with Nvidia NeMo}
\end{enumerate}

\chapter{Multimodal LLMs and their Fine-tuning}

A multimodal model is a machine learning model that can process information from various modalities, such as images, videos, and text. For instance, Google's multimodal model, Gemini\cite{geminiPaper}, can analyse a photo of a plate of cookies and produce a written recipe in response, and it can perform the reverse as well.\\

\noindent The difference between Generative AI and Multimodal AI is that generative AI refers to the use of machine learning models to create new content, such as text, images, music, audio, and videos, typically from a single type of input. Multimodal AI extends these generative capabilities by processing information from multiple modalities, including images, videos, and text. This enables the AI to understand and interpret different sensory modes, allowing users to input various types of data and receive a diverse range of content types in return.

\begin{figure}[!ht]
    \centering
    \includegraphics[width=1\textwidth]{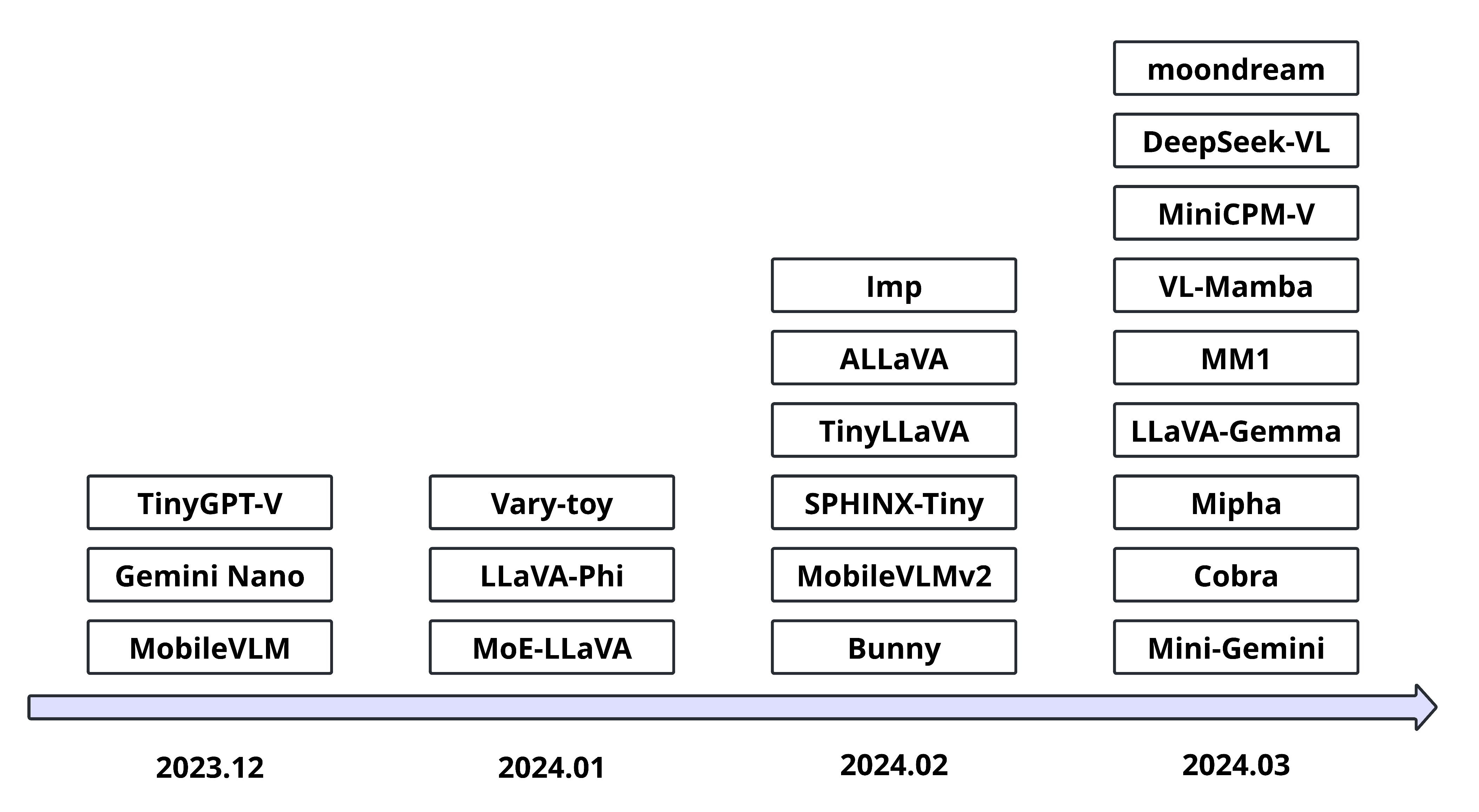} 
    \caption{Timeline of Multimodal Model Developments — This figure illustrates the progression of significant multimodal models, highlighting key releases from major tech companies and research institutions from December 2023 to March 2024. The timeline showcases models like Google's TinyGPT-V and Gemini Nano, along with other innovations such as MoE-LLAVA, DeepSeek-VL, and LLAVA-Gemma, indicating the rapid advancement in multimodal AI technologies (adapted from \cite{efficientMultimodals}).}
    \label{multimodalModelsTimeline}
\end{figure}

\section{Vision Language Model (VLMs)}

Vision language models encompass multimodal models capable of learning from both images and text inputs. They belong to the category of generative models that utilise image and text data to produce textual outputs. These models, especially at larger scales, demonstrate strong zero-shot capabilities, exhibit robust generalisation across various tasks, and effectively handle diverse types of visual data such as documents and web pages.
Typical applications include conversational interactions involving images, image interpretation based on textual instructions, answering questions related to visual content, understanding documents, generating captions for images, and more.
Certain advanced vision language models can also understand spatial attributes within images. They can generate bounding boxes or segmentation masks upon request to identify or isolate specific subjects, localise entities within images, or respond to queries regarding their relative or absolute positions.
The landscape of large vision language models is characterised by considerable diversity in training data, image encoding techniques, and consequently, their functional capabilities.
 
\subsection{Architecture}

Vision-language models adeptly integrate both visual and textual information, leveraging three fundamental components:
\begin{itemize}
    \item \textbf{Image Encoder:} This component translates visual data (images) into a format that the model can process.
    \item \textbf{Text Encoder:} Similar to the image encoder, this component converts textual data (words and sentences) into a format the model can understand.
    \item \textbf{Fusion Strategy:} This component combines the information from both the image and text encoders, merging the two data types into a unified representation.
\end{itemize}

\noindent These elements work collaboratively, with the model's learning process (loss functions) specifically tailored to the architecture and learning strategy employed.
Although the concept of vision-language models is not new, their construction has evolved significantly. Early models used manually crafted image descriptions and pre-trained word vectors. Modern models, however, utilise transformers—an advanced neural network architecture—for both image and text encoding. These encoders can learn features either independently or jointly.

\noindent A crucial aspect of these models is pre-training. Before being applied to specific tasks, the models are trained on extensive datasets using carefully selected objectives. This pre-training equips them with the foundational knowledge required to excel in various downstream applications. Following is one of the example architectures of VLMs.

\subsection{Contrastive Learning}

Contrastive learning is a technique that focuses on understanding the differences between data points. It computes a similarity score between instances and aims to minimise contrastive loss, making it particularly useful in semi-supervised learning where a limited number of labelled samples guide the optimisation process to classify unseen data points.

\subsubsection{How it works}

For instance, to recognise a cat, contrastive learning compares a cat image with a similar cat image and a dog image. The model learns to distinguish between a cat and a dog by identifying features such as facial structure, body size, and fur. By determining which image is closer to the "anchor" image, the model predicts its class.

\begin{figure}[!ht]
    \centering
    \includegraphics[width=1\textwidth]{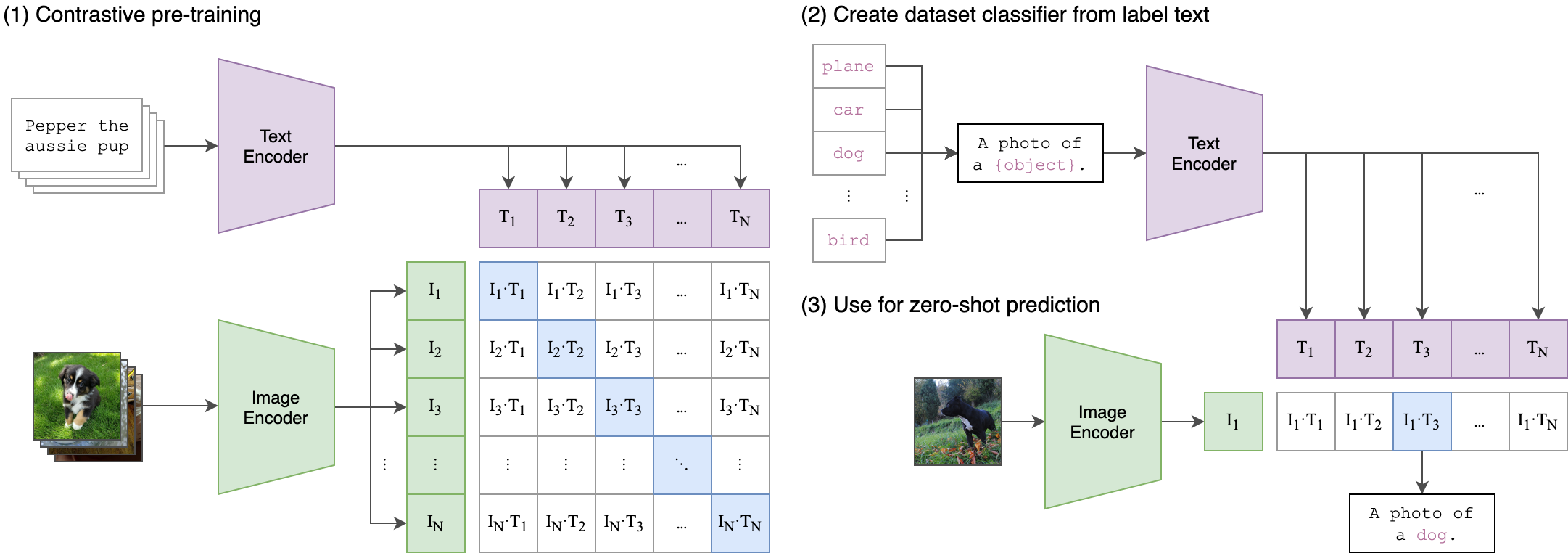} 
    \caption{Workflow of Contrastive Pre-Training for Multimodal Models. This figure illustrates the process of contrastive pre-training where text and image encoders are trained to align representations from both modalities. Step 1 involves contrastive pre-training by pairing text and image data, while Step 2 showcases the creation of a dataset classifier using label text encoded by the text encoder. Step 3 demonstrates the model's application for zero-shot prediction by leveraging the pre-trained text and image encoders. This method enables the model to generalise across various tasks without requiring task-specific fine-tuning (adopted from \cite{contrastivePretrainingPaper}).}
    \label{contranstivePretraining}
\end{figure}

\noindent CLIP is a model that utilises contrastive learning to compute similarity between text and image embeddings through textual and visual encoders. It follows a three-step process for zero-shot predictions:
\begin{itemize}
    \item \textbf{Pre-training:} Trains a text and image encoder to learn image-text pairs.
    
    \item \textbf{Caption Conversion:} Converts training dataset classes into captions.
    
    \item \textbf{Zero-Shot Prediction:} Estimates the best caption for a given input image based on learned similarities.
\end{itemize}

\section{Fine-tuning of multimodal models}

For fine-tuning a Multimodal Large Language Model (MLLM), PEFT techniques such as LoRA and QLoRA can be utilised. The process of fine-tuning for multimodal applications is analogous to that for large language models, with the primary difference being the nature of the input data. In addition to LoRA, which employs matrix factorisation techniques to reduce the number of parameters, other tools such as LLM-Adapters and (IA)³\cite{ia3Paper} can be effectively used. LLM-Adapters integrate various adapter modules into the pre-trained model’s architecture, enabling parameter-efficient fine-tuning for diverse tasks by updating only the adapter parameters while keeping the base model parameters fixed. (IA)³, or Infused Adapters by Inhibiting and Amplifying Inner Activations, enhances performance by learning vectors to weight model parameters through activation multiplications, supporting robust few-shot performance and task mixing without manual adjustments. Moreover, dynamic adaptation techniques like DyLoRA\cite{dyLoraPaper} allow for the training of low-rank adaptation blocks across different ranks, optimising the learning process by sorting the representations during training. LoRA-FA\cite{loraFaPaper}, a variant of LoRA, optimises the fine-tuning process by freezing the first low-rank matrix after initialisation and using it as a random projection while training the other, thereby reducing the number of parameters by half without compromising performance.

\noindent The Efficient Attention Skipping (EAS)\cite{efficientAttentionSkippingPaper} module introduces a novel parameter and computation-efficient tuning method for MLLMs, aiming to maintain high performance while reducing parameter and computation costs for downstream tasks. However, MemVP\cite{memVPPaper} critiques this approach, noting that it still increases the input length of language models. To address this, MemVP integrates visual prompts with the weights of Feed Forward Networks, thereby injecting visual knowledge to decrease training time and inference latency, ultimately outperforming previous PEFT methods.

\subsection{Full-parameter Fine-Tuning} 
Methods such as those introduced by LOMO\cite{lomoPaper} and MeZO\cite{mezoPaper} provide alternative solutions by focusing on memory efficiency. LOMO utilises a low-memory optimisation technique derived from Stochastic Gradient Descent (SGD), reducing memory consumption typically associated with the ADAM optimiser. MeZO, on the other hand, offers a memory-efficient optimiser that requires only two forward passes to compute gradients, enabling comprehensive fine-tuning of large models with a memory footprint equivalent to inference \cite{efficientMultimodals}.

\subsection{Case study of fine-tuning MLLMs for Medical domain}
The following section provides a case study on fine-tuning MLLMs for the Visual Question Answering (VQA) task. In this example, we present a PEFT for fine-tuning MLLM specifically designed for Med-VQA applications. To ensure accurate performance measurement, human evaluations were conducted, demonstrating that the model achieves an overall accuracy of 81.9\% and surpasses the GPT-4v model by a substantial margin of 26\% in absolute accuracy on closed-ended questions.

\begin{figure}[!ht]
    \centering
    \includegraphics[width=1\textwidth]{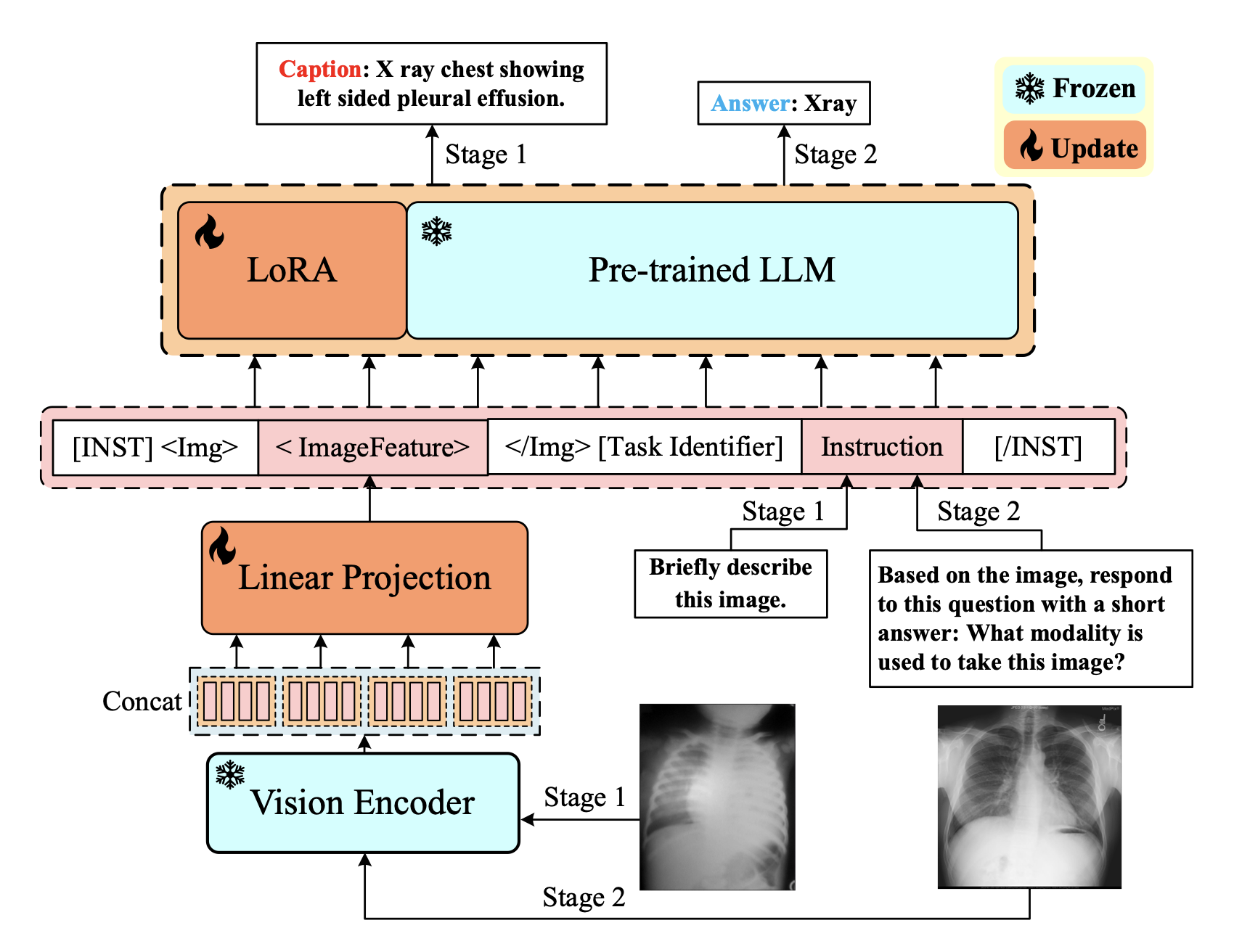} 
    \caption{Overview of Med VQA architecture integrating LoRA and a pre-trained LLM with a Vision Encoder for medical visual question answering tasks. The architecture includes stages for processing images and generating contextually relevant responses, demonstrating the integration of vision and language models in a medical setting (adopted from \cite{medVqaPaper}).}
    \label{medVqaArchitecture}
\end{figure}

\noindent The model consists of three components: the vision encoder, a pre-trained Large Language Model (LLM) for handling multimodal inputs and generating responses, and a single linear layer for projecting embeddings from the visual encoding space to the LLM space, as shown in figure \ref{medVqaArchitecture}.

\noindent The Vision Transformer (ViT) type backbone, EVA, encodes image tokens into visual embeddings, with model weights remaining frozen during the fine-tuning process. The technique from MiniGPT-v2 is utilised, grouping four consecutive tokens into one visual embedding to efficiently reduce resource consumption by concatenating on the embedding dimension.

\noindent These grouped visual tokens are then processed through the projection layer, resulting in embeddings (length 4096) in the LLM space. A multimodal prompt template integrates both visual and question information, which is input into the pre-trained LLM, LLaMA2-chat(7B), for answer generation. The low-rank adaptation (LoRA) technique is applied for efficient fine-tuning, keeping the rest of the LLM frozen during downstream fine-tuning. A beam search with a width of 1 is utilised.

\noindent The multimodal prompt includes input images, questions, and a specific token for VQA tasks, following the MiniGPT-v2 template. In Figure \ref{medVqaArchitecture}, the image features derived from linear projection are labelled as \texttt{ImageFeature}, with the corresponding questions serving as text instructions. The special token \texttt{[VQA]} is used as the task identifier, forming the complete multimodal instructional template: 

\begin{verbatim}
[INST]<img><ImageFeature></img>[VQA] Instruction [/INST].
\end{verbatim}

\subsubsection{Model Training}
Weights from MiniGPT-v2, pre-trained on general domain datasets, are further fine-tuned using multimodal medical datasets in two stages. The LoRA technique is employed for efficient fine-tuning, updating only a small portion of the entire model, as detailed below:

\begin{itemize}
    \item \textbf{Fine-tuning with image captioning:} During this stage, the model is fine-tuned using the ROCO medical image-caption dataset, which contains medical image-caption pairs of varying lengths.
    The prompt template used is \texttt{<Img><ImageHere></Img>[caption] <instruction>}, with the instruction prompt randomly selected from a pool of four candidates, such as “Briefly describe this image.”
    During training, only the linear projection layer and the LoRA layer in the LLM are fine-tuned, while other parts of the model remain frozen.

    \item \textbf{Fine-tuning on VQA:} In the second stage, the model is fine-tuned on the Med-VQA dataset, VQA-RAD, which contains triplets of images, questions, and answers. Following the instruction template proposed in MiniGPT-v2, the template used is: “\texttt{[INST] <img><ImageFeature></img>[VQA] Instruction [/INST]}”, where the instruction prompt is: “Based on the image, respond to this question with a short answer: {question},” with {question} signifying the question corresponding to the given medical image.
    The motivation for generating short answers is to validate against the existing labelled data in VQA-RAD, where the answers are typically short in both open-ended and closed-ended QA pairs. Similar to the first stage, the vision encoder and the LLM remain frozen while only the linear projection and LoRA layers in the LLM are updated.
\end{itemize}

\section{Applications of Multimodal models}

\begin{enumerate}
    \item \textbf{Gesture Recognition} - These models interpret and recognise human gestures, which is crucial for sign language translation. Multimodal models facilitate inclusive communication by processing gestures and converting them into text or speech.
    
    \item \textbf{Video Summarisation} - Multimodal models can summarise lengthy videos by extracting key visual and audio elements. This capability streamlines content consumption, enables efficient content browsing, and enhances video content management platforms.

    \item \textbf{DALL-E} is a notable example of multimodal AI that generates images from textual descriptions. This technology expands creative possibilities in content creation and visual storytelling, with applications in art, design, advertising, and more.

    \item \textbf{Educational Tools} - Multimodal models enhance learning experiences by providing interactive educational content that responds to both visual and verbal cues from students. They are integral to adaptive learning platforms that adjust content and difficulty based on student performance and feedback.

    \item \textbf{Virtual Assistants} - Multimodal models power virtual assistants by understanding and responding to voice commands while processing visual data for comprehensive user interaction. They are essential for smart home automation, voice-controlled devices, and digital personal assistants.

\end{enumerate}

\section{Audio or Speech LLMs Or Large Audio Models}
Audio or speech LLMs are models designed to understand and generate human language based on audio inputs. They have applications in speech recognition, text-to-speech conversion, and natural language understanding tasks. These models are typically pre-trained on large datasets to learn generic language patterns, which are then fine-tuned on specific tasks or domains to enhance performance.

\noindent Audio and Speech Large Language Models (LLMs) represent a significant advancement in the integration of language processing with audio signals. These models leverage a robust Large Language Model as a foundational backbone, which is enhanced to handle multimodal data through the inclusion of custom audio tokens. This transformation allows the models to learn and operate within a shared multimodal space, where both text and audio signals can be effectively processed.

\begin{figure}[!ht]
    \centering
    \includegraphics[width=1\textwidth]{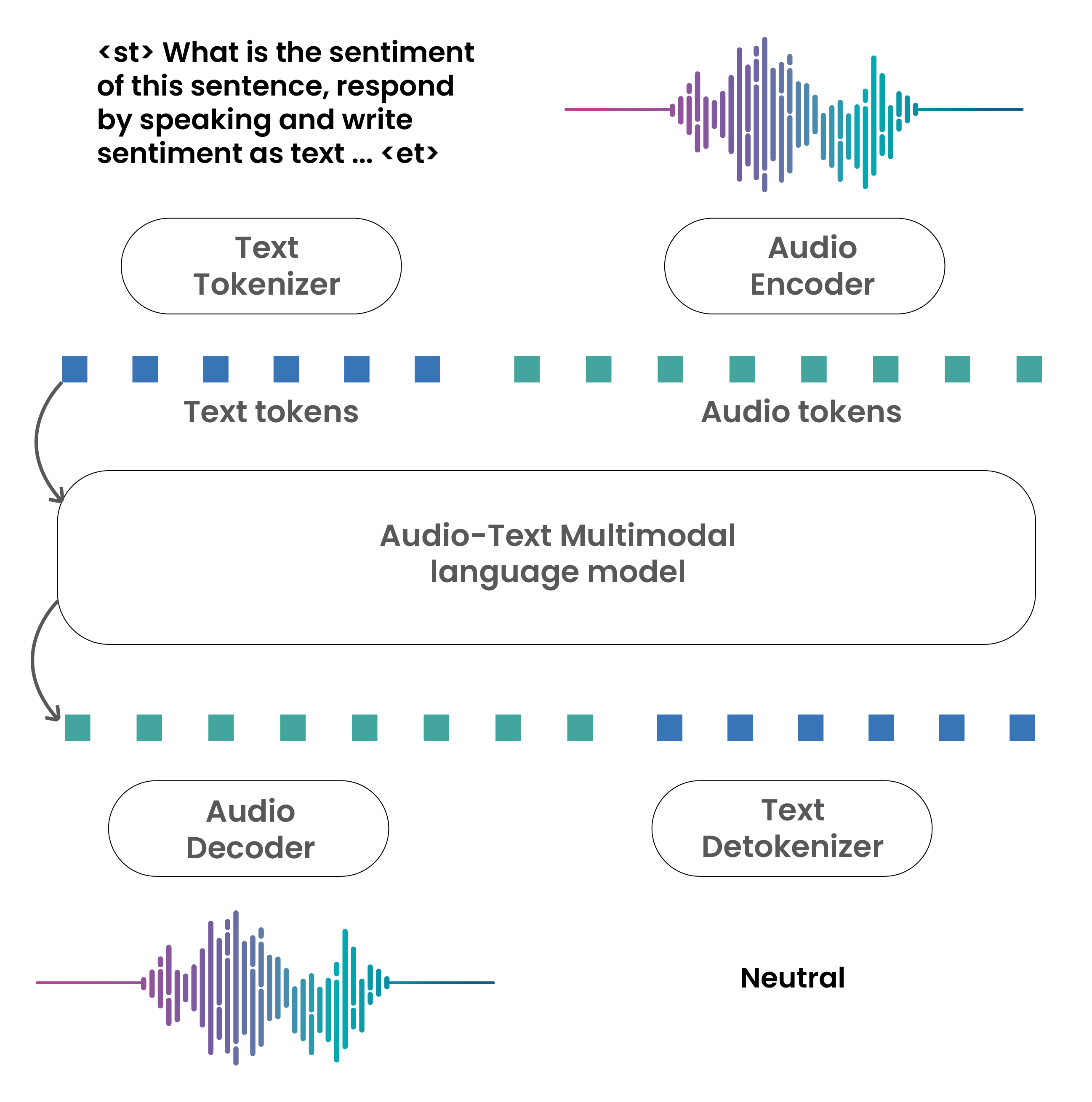} 
    \caption{Multimodal Audio-Text Language Model architecture that integrates text and audio inputs for advanced multimodal processing. The architecture utilises text tokenizers and audio encoders/tokenizers to convert inputs into tokens, which are then processed by the audio-text LM. This model supports both discrete and continuous speech processing and enables tasks such as sentiment analysis and response generation in natural language. The audio tokens are further refined using a vocoder, while text tokens are detokenized to produce coherent text outputs (adapted from \cite{AudioLanguage}).}
    \label{AudioLLMExample}
\end{figure}

\noindent Unlike text, which is inherently discrete, audio signals are continuous and need to be discretized into manageable audio tokens. Techniques like HuBERT\cite{hsu2021hubertselfsupervisedspeechrepresentation} and wav2vec\cite{baevski2020wav2vec20frameworkselfsupervised} are employed for this purpose, converting audio into a tokenized format that the LLM can process alongside text. The model, typically autoregressive and decoder-based, is pre-trained using a combination of self-supervised tasks, such as predicting masked tokens in interleaved text and audio, and supervised fine-tuning for specific tasks like transcription or sentiment analysis. This capability to handle and generate audio and text simultaneously allows for a wide range of applications, from audio question answering to speech-based sentiment detection, making Audio and Speech LLMs a versatile tool in multimodal AI. The figure \ref{AudioLLMExample} illustrates an example of a multimodal Audio LM architecture. In this setup, a prompt provides instructions in both text and audio formats. The audio is tokenized using an audio tokenizer. The multimodal model then combines these text and audio tokens and generates spoken speech through a vocoder (also known as a voice decoder).

\noindent Audio and speech LLMs like AudioPaLM\cite{rubenstein2023audiopalmlargelanguagemodel}, AudioLM\cite{borsos2023audiolmlanguagemodelingapproach}, and various adaptations of models like Whisper and LLaMA, integrate capabilities for understanding and generating audio data, including speech-to-text (STT), text-to-speech (TTS), and speech-to-speech (STS) translation. These models have shown that LLMs, initially designed for text, can be effectively adapted for audio tasks through sophisticated tokenization and fine-tuning techniques.

\subsection{Tokenization and Preprocessing}
A key aspect of adapting LLMs for audio is the tokenization of audio data into discrete representations that the model can process. For instance, AudioLM and AudioPaLM utilise a combination of acoustic and semantic tokens. Acoustic tokens capture the high-quality audio synthesis aspect, while semantic tokens help maintain long-term structural coherence in the generated audio. This dual-token approach allows the models to handle both the intricacies of audio waveforms and the semantic content of speech.

\subsection{Fine-Tuning Techniques}
Fine-tuning audio and speech LLMs typically involve several key strategies:
\begin{itemize}
    \item \textbf{Full Parameter Fine-Tuning:} This involves updating all the model's parameters during fine-tuning. For instance, LauraGPT and SpeechGPT fine-tune all parameters to adapt pre-trained text LLMs to various audio tasks, although this can be computationally expensive.
    \item \textbf{Layer-Specific Fine-Tuning:} Techniques like LoRA (Low-Rank Adaptation) update only specific layers or modules of the model. This method significantly reduces computational requirements while still allowing effective adaptation. Models like Qwen-Audio leverage LoRA to fine-tune pre-trained components for enhanced performance on speech recognition tasks.
    \item \textbf{Component-Based Fine-Tuning:} Recent models, such as those integrating the Whisper encoder, freeze certain parts of the model (like the speech encoder) and only fine-tune a linear projector or specific adapters to align the speech and text modalities. This approach simplifies the training process and enhances efficiency\cite{naveed2024comprehensiveoverviewlargelanguage}.
    \item \textbf{Multi-Stage Fine-Tuning:} Models like AudioPaLM perform multi-stage fine-tuning, starting with a text-based pre-training phase, followed by fine-tuning on a mixture of tasks that include both text and audio data. This staged approach leverages the strengths of pre-trained text models while adapting them for multimodal tasks.
\end{itemize}

\subsection{Fine-Tuning Whisper for Automatic Speech Recognition (ASR)}
Whisper\footnote{\url{https://openai.com/index/whisper/}} is an advanced Automatic Speech Recognition (ASR) model developed by OpenAI, designed to convert spoken language into text. Built upon the powerful Transformer architecture, Whisper excels at capturing and transcribing diverse speech patterns across various languages and accents. Unlike traditional ASR models that require extensive labelled data, Whisper leverages a vast dataset and self-supervised learning, enabling it to perform robustly in noisy environments and handle a wide range of speech variations. Its versatility and high accuracy make it an ideal choice for applications such as voice assistants, transcription services, and multilingual speech recognition systems.

\subsubsection{Why Fine-Tune Whisper?}
Fine-tuning Whisper for specific ASR tasks can significantly enhance its performance in specialised domains. Although Whisper is pre-trained on a large and diverse dataset, it might not fully capture the nuances of specific vocabularies or accents present in niche applications. Fine-tuning allows Whisper to adapt to particular audio characteristics and terminologies, leading to more accurate and reliable transcriptions. This process is especially beneficial in industries with domain-specific jargon, like medical, legal, or technical fields, where the generic model might struggle with specialised vocabulary.

\subsubsection{Steps to Fine-Tune Whisper}
\begin{itemize}
    \item \textbf{Data Collection and Preparation:}
        Gather a sizable dataset that matches the target domain or task. Ensure the dataset includes diverse examples with clear transcriptions. Clean and preprocess the audio files and transcripts, ensuring they are in a consistent format and aligned correctly. Tools like FFmpeg\footnote{\url{https://ffmpeg.org/ffmpeg.html}} can help standardise audio formats and sample rates.
    \item \textbf{Data Augmentation:}
        To improve robustness, augment the dataset with variations such as different noise levels, accents, or speeds. Techniques like adding background noise, altering pitch, or changing the tempo can help the model generalise better to real-world conditions.
    \item \textbf{Preprocessing:}
        Convert the audio files into a format suitable for Whisper, typically into mel spectrograms or another time-frequency representation. This transformation is crucial as Whisper relies on such representations to learn and transcribe speech effectively.
    \item \textbf{Model Configuration:}
        Initialise the Whisper model with pre-trained weights. Configure the model to accommodate the target language or domain-specific adjustments. This includes setting appropriate hyperparameters, like learning rate and batch size, tailored to the dataset's size and complexity.
    \item \textbf{Training:}
        Fine-tune the Whisper model on the prepared dataset using a framework like PyTorch or TensorFlow. Ensure to monitor the model’s performance on a validation set to avoid overfitting. Techniques like gradient clipping, learning rate scheduling, and early stopping can help maintain training stability and efficiency.
    \item \textbf{Evaluation and Testing:}
        After training, evaluate the model's performance on a separate test set to assess its accuracy and generalisability. Metrics like Word Error Rate (WER) or Character Error Rate (CER) provide insights into how well the model transcribes audio compared to ground truth transcriptions.
\end{itemize}

\subsection{Case Studies and Applications}
\begin{enumerate}
    \item \textbf{Medical Transcription:} Fine-tuning speech LLMs on medical data has led to significant improvements in transcribing doctor-patient interactions. Models like Whisper have been fine-tuned on medical terminologies, resulting in more accurate and reliable transcriptions.

    \item \textbf{Legal Document Processing:} Legal firms have employed fine-tuned audio LLMs to transcribe court proceedings and legal discussions. Domain-specific fine-tuning has enhanced the models' ability to recognise and accurately transcribe legal jargon.

    \item \textbf{Customer Service Automation:} Companies are using fine-tuned speech models to automate customer service interactions. These models are trained on customer support data to understand and respond to queries more effectively, providing a more seamless user experience.
\end{enumerate}

\chapter{Open Challenges and Research Directions}

\section{Scalability Issues}

The fine-tuning of Large Language Models (LLMs) such as GPT-4, PaLM\footnote{\url{https://ai.google/discover/palm2/}}
, and T5\footnote{\url{https://huggingface.co/docs/transformers/en/model_doc/t5}}
 has become a critical area of research, presenting several significant challenges and opening up new avenues for exploration, particularly in scaling these processes efficiently. This discussion focuses on the two main aspects: the challenges in scaling fine-tuning processes and potential research directions for scalable solutions.

\subsection{Challenges in Scaling Fine-Tuning Processes}

\begin{enumerate}
    \item \textbf{Computational Resources:} Large-scale models such as GPT-3 and PaLM require enormous computational resources for fine-tuning. For instance, fine-tuning a 175-billion parameter model like GPT-3 necessitates high-performance GPUs or TPUs capable of handling vast amounts of data and complex operations. The sheer volume of parameters translates to extensive computational demands. Even a relatively smaller model, such as BERT-large with 340 million parameters, can be computationally intensive to fine-tune.
    
    \item \textbf{Memory Requirements:} The memory footprint for fine-tuning LLMs is staggering. Each parameter in the model requires storage, and during training, additional memory is needed to store intermediate computations, gradients, and optimiser states. For example, loading a 7 billion parameter model (e.g., LLaMA 2) in FP32 (4 bytes per parameter) requires approximately 28 GB of GPU memory, while fine-tuning demands around 112 GB of GPU memory\cite{amdFinetuneLlama}. This memory demand is beyond the capability of most consumer-grade hardware, making fine-tuning accessible primarily to well-funded organisations or research institutions.
    
    \item \textbf{Data Volume:} LLMs typically require vast amounts of training data to achieve state-of-the-art performance during fine-tuning. This data needs to be loaded, preprocessed, and fed into the model at high speeds to maintain efficient training. Managing large datasets can become a bottleneck, especially if the data is stored in a distributed fashion across multiple systems or if it needs to be fetched from remote storage.
    
    \item \textbf{Throughput and Bottlenecks:} High throughput is essential to keep GPUs or TPUs fully utilised. However, data pipelines can become bottlenecks if not properly optimised. For example, shuffling large datasets or loading them into memory quickly enough to keep up with the training process can be challenging. Techniques like data packing, where multiple small examples are combined into larger batches, help improve throughput but add complexity to data handling routines.\cite{uniteUnderstandingFineTuning}
    
    \item \textbf{Efficient Use of Resources:} The financial and environmental costs of fine-tuning large models are significant. Large-scale fine-tuning involves not just the direct cost of computational resources but also the indirect costs associated with energy consumption and infrastructure maintenance. Techniques such as mixed-precision training and gradient checkpointing can reduce these costs by optimising memory and computational efficiency.
\end{enumerate}

\noindent The challenges in scaling the fine-tuning processes of LLMs are multifaceted and complex, involving significant computational, memory, and data handling constraints. Innovations in PEFT, data throughput optimisation, and resource-efficient training methods are critical for overcoming these challenges. As LLMs continue to grow in size and capability, addressing these challenges will be essential for making advanced AI accessible and practical for a wider range of applications.

\subsection{Research Directions for Scalable Solutions}

\subsubsection{Advanced PEFT Techniques and Sparse Fine-Tuning}
Recent advancements in PEFT techniques, like LoRA and its variant, Quantised LoRA, are revolutionising the scalability of LLMs. LoRA reduces the computational burden by updating only a low-rank approximation of the parameters, significantly lowering memory and processing requirements. Quantised LoRA further optimises resource usage by applying quantisation to these low-rank matrices, maintaining high model performance while minimising the need for extensive hardware. This has enabled efficient fine-tuning of massive models, such as in Meta's LLaMA project, where adapting a smaller set of influential parameters allowed the models to perform robustly across various tasks with less computational strain.\\

\noindent Sparse fine-tuning techniques, such as SpIEL \cite{ansell2024scalingsparsefinetuninglarge} complement these efforts by selectively updating only the most impactful parameters. SpIEL fine-tunes models by only changing a small portion of the parameters, which it tracks with an index. The process includes updating the parameters, removing the least important ones, and adding new ones based on their gradients or estimated momentum using an efficient optimiser.

\subsubsection{Data Efficient Fine-Tuning (DEFT)}
To address the scalability challenges, recently the concept of DEFT has emerged. This novel approach introduces data pruning as a mechanism to optimise the fine-tuning process by focusing on the most critical data samples.\\

\noindent DEFT aims to enhance the efficiency and effectiveness of fine-tuning LLMs by selectively pruning the training data to identify the most influential and representative samples. This method leverages few-shot learning principles, enabling LLMs to adapt to new data with minimal samples while maintaining or even exceeding performance levels achieved with full datasets \cite{lin2024dataefficientfinetuningllmbasedrecommendation}.

\subsubsection{Key Components of DEFT}
\textbf{High Accuracy Through Influence Score:} DEFT introduces the concept of an influence score to evaluate and rank the importance of each data sample in the context of LLM fine-tuning. The influence score estimates how removing a specific sample would impact the overall performance of the model. This approach allows for the selection of a small subset of data that is highly representative and influential, thereby enabling the model to maintain high accuracy with significantly fewer samples.\\

\noindent \textbf{High Efficiency Through Effort Score and Surrogate Models:} To address the cost and complexity of evaluating large datasets, DEFT employs a surrogate model—a smaller, computationally less intensive model—to approximate the influence scores. This surrogate model helps estimate the impact of each sample without the heavy computational burden associated with directly using the LLM. Additionally, DEFT introduces an effort score to identify and prioritise more challenging samples that may require special attention from the LLM. This dual-score system ensures that the fine-tuning process remains both efficient and effective.

\subsubsection{Practical Implications and Use Cases}
\begin{itemize}
    \item \textbf{Few-Shot Fine-Tuning for Rapid Adaptation:} DEFT is particularly beneficial for applications where models need to quickly adapt to new data with minimal samples. In scenarios such as personalised recommendations or adapting to sudden changes in user behaviour, DEFT allows for rapid fine-tuning, maintaining high performance with a fraction of the data typically required.
    \item \textbf{Reducing Computational Costs in Large-Scale Deployments:} By focusing on the most influential data samples and using surrogate models, DEFT significantly reduces the computational resources needed for fine-tuning. This makes it feasible to maintain high-performing LLMs even in large-scale deployments where data volumes are substantial.
\end{itemize}

\subsubsection{Future Directions}
The DEFT introduces a data pruning task for fine-tuning large language models (LLMs), setting the stage for new research into efficient LLM-based recommendation systems and presenting numerous opportunities for future exploration. Key areas for further investigation include:
\begin{itemize}
    \item Applying the proposed DEALRec\cite{liu2024endtoendlearnableclusteringintent} approach to a broader range of LLM-based recommender models across diverse cross-domain datasets, thereby enhancing fine-tuning performance within resource constraints.
    \item Addressing the limited context window of LLMs by selectively focusing on the most informative items in user interaction sequences for fine-tuning purposes.
\end{itemize}

\subsection{Hardware and Algorithm Co-Design}
Co-designing hardware and algorithms tailored for LLMs can lead to significant improvements in the efficiency of fine-tuning processes. Custom hardware accelerators optimised for specific tasks or types of computation can drastically reduce the energy and time required for model training and fine-tuning.

\begin{itemize}
    \item \textbf{Custom Accelerators:} Developing hardware accelerators specifically for the sparse and low-precision computations often used in LLM fine-tuning can enhance performance. These accelerators are designed to efficiently handle the unique requirements of LLMs, such as the high memory bandwidth and extensive matrix multiplications involved in transformer architectures.
    \item \textbf{Algorithmic Optimisation:} Combining hardware innovations with algorithmic optimisation techniques, such as those that minimise data movement or leverage hardware-specific features (e.g., tensor cores for mixed-precision calculations), can further enhance the efficiency of fine-tuning processes.
    \item \textbf{Example:} NVIDIA's TensorRT\footnote{\url{https://docs.nvidia.com/tensorrt/index.html}} is an example of hardware and algorithm co-design in action. It optimises deep learning models for inference by leveraging NVIDIA GPUs' capabilities, significantly speeding up the process while reducing the resource requirements. TensorRT's optimisations include support for mixed-precision and sparse tensor operations, making it highly suitable for fine-tuning large models.
\end{itemize}

\noindent As the scale of language models continues to grow, addressing the challenges of fine-tuning them efficiently becomes increasingly critical. Innovations in PEFT, sparse fine-tuning, data handling, and the integration of advanced hardware and algorithmic solutions present promising directions for future research. These scalable solutions are essential not only to make the deployment of LLMs feasible for a broader range of applications but also to push the boundaries of what these models can achieve.

\section{Ethical Considerations in Fine-Tuning LLMs}

\subsection{Bias and Fairness}
When fine-tuning LLMs, the goal is often to optimise their performance for specific tasks or datasets. However, these datasets may inherently carry biases that get transferred to the model during the fine-tuning process. Biases can arise from various sources, including historical data, imbalanced training samples, and cultural prejudices embedded in language. For instance, an LLM fine-tuned on a dataset primarily sourced from English-speaking countries might underperform or make biased predictions when applied to text from other linguistic or cultural backgrounds. Google AI's Fairness Indicators tool\footnote{\url{https://research.google/blog/fairness-indicators-scalable-infrastructure-for-fair-ml-systems/}} is a practical solution that allows developers to evaluate the fairness of their models by analysing performance metrics across different demographic groups. This tool can be integrated into the fine-tuning pipeline to monitor and address bias in real-time.

\subsubsection{Addressing Bias and Fairness}
\begin{itemize}
    \item \textbf{Diverse and Representative Data:} Ensuring that fine-tuning datasets are diverse and representative of all user demographics can help mitigate bias.
    \item \textbf{Fairness Constraints:} Incorporating fairness constraints, as suggested by the FairBERTa framework\footnote{\url{https://huggingface.co/facebook/FairBERTa}}, ensures that fine-tuned models maintain equitable performance across different groups.
    \item \textbf{Example Application:} In healthcare, an LLM fine-tuned to assist in diagnosing conditions might initially be trained on data from predominantly white patients. Such a model could produce less accurate diagnoses for patients from other racial backgrounds. By using fairness-aware fine-tuning techniques, healthcare providers can develop models that perform more equitably across diverse patient populations.
\end{itemize}

\subsection{Privacy Concerns}
Fine-tuning often involves using sensitive or proprietary datasets, which poses significant privacy risks. If not properly managed, fine-tuned models can inadvertently leak private information from their training data. This issue is especially critical in domains like healthcare or finance, where data confidentiality is paramount.

\subsubsection{Ensuring Privacy During Fine-Tuning}
\begin{itemize}
    \item \textbf{Differential Privacy\footnote{\url{https://privacytools.seas.harvard.edu/differential-privacy}}:} Implementing differential privacy techniques during fine-tuning can prevent models from leaking sensitive information.
    \item \textbf{Federated Learning\footnote{\url{https://research.ibm.com/blog/what-is-federated-learning}}:} Utilising federated learning frameworks allows models to be fine-tuned across decentralised data sources, which enhances privacy by keeping data localised.
    \item \textbf{Example Application:} In customer service applications, companies might fine-tune LLMs using customer interaction data. Employing differential privacy ensures that the model learns from these interactions without memorising and potentially leaking personal information, thus maintaining customer confidentiality.
\end{itemize}

\subsection{Security Risks}
\begin{itemize}
    \item \textbf{Security Vulnerabilities in Fine-Tuned Models:} Fine-tuned LLMs are susceptible to security vulnerabilities, particularly from adversarial attacks. These attacks involve inputs designed to exploit model weaknesses, causing them to produce erroneous or harmful outputs. Such vulnerabilities can be more pronounced in fine-tuned models due to their specialised training data, which may not cover all possible input scenarios.

    \item \textbf{Recent Research and Industry Practices:} Microsoft’s Adversarial ML Threat Matrix provides a comprehensive framework for identifying and mitigating adversarial threats during model development and fine-tuning. This matrix helps developers understand the potential attack vectors and implement defensive strategies accordingly.

    \item \textbf{Enhancing Security in Fine-Tuning:}
    \begin{itemize}
        \item \textbf{Adversarial Training:} Exposing models to adversarial examples during fine-tuning can enhance their robustness against attacks.
        \item \textbf{Security Audits:} Regularly conducting security audits on fine-tuned models can help identify and address potential vulnerabilities.
    \end{itemize}
\end{itemize}

\section{Accountability and Transparency}

\subsection{The Need for Accountability and Transparency}
Fine-tuning can significantly alter an LLM’s behaviour, making it crucial to document and understand the changes and their impacts. This transparency is essential for stakeholders to trust the model’s outputs and for developers to be accountable for its performance and ethical implications.

\subsection{Recent Research and Industry Practices}
Meta’s Responsible AI framework\footnote{\url{https://ai.meta.com/responsible-ai/}} underscores the importance of documenting the fine-tuning process and its effects on model behaviour. This includes maintaining detailed records of the data used, the changes made during fine-tuning, and the evaluation metrics applied.

\subsection{Promoting Accountability and Transparency}
\begin{itemize}
    \item \textbf{Comprehensive Documentation:} Creating detailed documentation of the fine-tuning process and its impact on model performance and behaviour.
    \item \textbf{Transparent Reporting:} Utilising frameworks like Model Cards\footnote{\url{https://huggingface.co/docs/hub/en/model-cards}} to report on the ethical and operational characteristics of fine-tuned models.
    \item \textbf{Example Application:} In content moderation systems, LLMs fine-tuned to identify and filter harmful content need clear documentation and reporting. This ensures that platform users and regulators understand how the model operates and can trust its moderation decisions.
\end{itemize}

\subsection{Proposed frameworks/techniques for Ethical Fine-Tuning}

\subsubsection{Frameworks for Mitigating Bias}
Bias-aware fine-tuning frameworks aim to incorporate fairness into the model training process. FairBERTa, introduced by Facebook, is an example of such a framework that integrates fairness constraints directly into the model’s objective function during fine-tuning. This approach ensures that the model’s performance is balanced across different demographic groups.\\

\noindent Organisations can adopt fairness-aware frameworks to develop more equitable AI systems. For instance, social media platforms can use these frameworks to fine-tune models that detect and mitigate hate speech while ensuring fair treatment across various user demographics.

\subsubsection{Techniques for Privacy Preservation}
Differential privacy and federated learning are key techniques for preserving privacy during fine-tuning. TensorFlow Privacy\footnote{\url{https://www.tensorflow.org/responsible_ai/privacy/guide}}, developed by Google, provides built-in support for differential privacy, allowing developers to fine-tune models securely without compromising data confidentiality.

\noindent LLMs are highly effective but face challenges when applied in sensitive areas where data privacy is crucial. To address this, researchers focus on enhancing Small Language Models (SLMs) tailored to specific domains. Existing methods often use LLMs to generate additional data or transfer knowledge to SLMs, but these approaches struggle due to differences between LLM-generated data and private client data. In response, a new Federated Domain-specific Knowledge Transfer (FDKT)\cite{li2024federateddomainspecificknowledgetransfer} framework is introduced. FDKT leverages LLMs to create synthetic samples that mimic clients' private data distribution using differential privacy. This approach significantly boosts SLMs' performance by approximately 5\% while maintaining data privacy with a minimal privacy budget, outperforming traditional methods relying solely on local private data.\\

\noindent In healthcare, federated fine-tuning can allow hospitals to collaboratively train models on patient data without transferring sensitive information. This approach ensures data privacy while enabling the development of robust, generalisable AI systems.

\subsubsection{Frameworks for Enhancing Security}
Adversarial training and robust security measures\cite{madry2019deeplearningmodelsresistant} are essential for protecting fine-tuned models against attacks. The adversarial training approach involves training models with adversarial examples to improve their resilience against malicious inputs. Microsoft Azure’s adversarial training tools provide practical solutions for integrating these techniques into the fine-tuning process, helping developers create more secure and reliable models.\\

\noindent In cybersecurity, fine-tuned LLMs used for threat detection can benefit from adversarial training to enhance their ability to identify and respond to sophisticated attacks, thereby improving organisational security.

\subsubsection{Frameworks for Ensuring Transparency}
Transparency and accountability frameworks, such as Model Cards and AI FactSheets\footnote{\url{https://aifs360.res.ibm.com/}}, provide structured ways to document and report on the fine-tuning process and the resulting model behaviours. These frameworks promote understanding and trust among stakeholders by clearly outlining the model’s capabilities, limitations, and ethical considerations.\\

\noindent In government applications, where AI systems might be used for decision-making or public services, maintaining transparent documentation through frameworks like AI FactSheets ensures that these systems are accountable and their decisions can be audited and trusted by the public.\\

\noindent Fine-tuning LLMs introduces several ethical challenges, including bias, privacy risks, security vulnerabilities, and accountability concerns. Addressing these requires a multifaceted approach that integrates fairness-aware frameworks, privacy-preserving techniques, robust security measures, and transparency and accountability mechanisms. By leveraging recent advancements in these areas, researchers and practitioners can develop and deploy LLMs that are not only powerful but also ethically sound and trustworthy.

\section{Integration with Emerging Technologies}

Integrating LLMs with emerging technologies such as IoT (Internet of Things) and edge computing presents numerous opportunities and challenges, reflecting advancements and insights from recent research and industry developments.

\subsection{Opportunities}

\begin{itemize}
    \item \textbf{Enhanced Decision-Making and Automation:} LLMs have the capability to analyse and derive insights from vast amounts of unstructured data generated by IoT devices. This data can range from sensor readings in manufacturing plants to environmental data in smart cities. By processing this data in real-time, LLMs can optimise decision-making processes and automate tasks that traditionally required human intervention. For example:
    
    \begin{itemize}
        \item \textbf{Industrial Applications:} Predictive maintenance can be enhanced by LLMs analysing sensor data to predict equipment failures before they occur, thereby reducing downtime and maintenance costs.
        \item \textbf{Smart Cities:} LLMs can analyse traffic patterns and environmental data from IoT sensors to optimise city infrastructure and improve urban planning decisions.
    \end{itemize}

    \item \textbf{Personalised User Experiences:} Integration with edge computing allows LLMs to process data locally on devices rather than relying solely on cloud-based servers. This enables LLMs to deliver highly personalised services based on real-time data and user preferences, enhancing user experiences across various domains:
    
    \begin{itemize}
        \item \textbf{Healthcare:} LLMs can provide personalised healthcare recommendations by analysing data from wearable devices and integrating it with medical records securely stored on edge devices.
    \end{itemize}

    \item \textbf{Improved Natural Language Understanding:} IoT data integration enriches LLMs' ability to understand context and respond more intelligently to natural language queries. This can significantly improve user interactions with smart environments:
    
    \begin{itemize}
        \item \textbf{Smart Homes:} LLMs integrated with IoT devices can understand and respond to voice commands more accurately, adjusting smart home settings based on real-time sensor data (e.g., adjusting lighting and temperature based on occupancy and environmental conditions).
    \end{itemize}

\end{itemize}

\subsection{Challenges}

\begin{itemize}
    \item \textbf{Data Complexity and Integration:} Integrating data from diverse IoT devices poses challenges related to data quality, interoperability, and scalability. LLMs need to effectively process and interpret this heterogeneous data to derive meaningful insights:
    
    \begin{itemize}
        \item \textbf{Data Integration:} Ensuring seamless integration of data streams from different IoT platforms and devices without compromising data integrity or performance.
        \item \textbf{Data Preprocessing:} Cleaning and preprocessing IoT data to ensure consistency and reliability before feeding it into LLMs for analysis.
    \end{itemize}

    \item \textbf{Privacy and Security:} Edge computing involves processing sensitive data locally on devices, raising concerns about data privacy and security:
    
    \begin{itemize}
        \item \textbf{Data Privacy:} Implementing robust encryption techniques and access control mechanisms to protect sensitive data processed by LLMs on edge devices.
        \item \textbf{Secure Communication:} Ensuring secure communication channels between IoT devices and LLMs to prevent data breaches or unauthorised access.
    \end{itemize}

    \item \textbf{Real-Time Processing and Reliability:} LLMs deployed in edge computing environments must operate with low latency and high reliability to support real-time applications:
    
    \begin{itemize}
        \item \textbf{Latency:} Optimising algorithms and processing capabilities of LLMs to handle real-time data streams efficiently without delays.
        \item \textbf{Reliability:} Ensuring the accuracy and consistency of insights generated by LLMs in dynamic and unpredictable IoT environments.
    \end{itemize}
\end{itemize}

\section{Future Research Areas}

\begin{itemize}
    \item \textbf{Federated Learning and Edge Computing:} Exploring federated learning techniques where LLMs can be trained collaboratively across edge devices without centralised data aggregation. This approach addresses privacy concerns and reduces communication overhead.

    \item \textbf{Real-Time Decision Support Systems:} Developing LLM-based systems capable of real-time decision-making by integrating with edge computing infrastructure. This includes optimising algorithms for low-latency processing and ensuring reliability under dynamic environmental conditions.

    \item \textbf{Ethical and Regulatory Implications:} Investigating the ethical implications of integrating LLMs with IoT and edge computing, particularly regarding data ownership, transparency, and fairness. This area requires frameworks for ethical AI deployment and governance.
\end{itemize}

\chapter*{Glossary}
\addcontentsline{toc}{chapter}{Glossary}

\begin{description}
    \item[LLM] Large Language Model – A type of AI model, typically with billions of parameters, trained on vast amounts of text data to understand and generate human-like text. They are primarily designed for tasks in natural language processing (NLP).

    \item[NLP] Natural Language Processing – A field of artificial intelligence that focuses on the interaction between computers and humans through natural language, including tasks like language generation, translation, and sentiment analysis.

    \item[LoRA] Low-Rank Adaptation – A parameter-efficient fine-tuning technique that adjusts only small low-rank matrices to adapt pre-trained models to specific tasks, thus preserving most of the original model's parameters.

    \item[DoRA] Weight-Decomposed Low-Rank Adaptation – A technique that decomposes model weights into magnitude and direction components, facilitating fine-tuning while maintaining inference efficiency.

    \item[QLoRA] Quantised Low-Rank Adaptation – A variation of LoRA, specifically designed for quantised models, allowing for efficient fine-tuning in resource-constrained environments.

    \item[PPO] Proximal Policy Optimisation – A reinforcement learning algorithm that adjusts policies by balancing the exploration of new actions and exploitation of known rewards, designed for stability and efficiency in training.

    \item[DPO] Direct Preference Optimisation – A method that directly aligns language models with human preferences through preference optimisation, bypassing reinforcement learning models like PPO.

    \item[MoE] Mixture of Experts – A model architecture that employs multiple specialised subnetworks, called experts, which are selectively activated based on the input to improve model performance and efficiency.

    \item[MoA] Mixture of Agents – A multi-agent framework where several agents collaborate during training and inference, leveraging the strengths of each agent to improve overall model performance.

    \item[PEFT] Parameter-Efficient Fine-Tuning – A fine-tuning approach for large models that involves adjusting only a subset of model parameters, improving efficiency in scenarios with limited computational resources. This includes techniques like LoRA, QLoRA, and adapters.

    \item[Adapters] Small, trainable modules introduced into the layers of pre-trained language models, allowing efficient task-specific fine-tuning without modifying the core parameters of the original model. Techniques such as **AdapterFusion** and **AdapterSoup** fall under this category, facilitating the combination of multiple adapters for complex multitasking.

    \item[Soft Prompt Tuning (SPT)] A fine-tuning technique where a set of trainable prompt tokens are added to the input sequence to guide a pre-trained model towards task-specific performance without modifying internal model weights.

    \item[Prefix-Tuning] A variation of soft prompt tuning where a fixed sequence of trainable vectors is prepended to the input layer at every layer of the model, enhancing task-specific adaptation.

    \item[Quantisation] The process of reducing the precision of model weights and activations, often from 32-bit to lower-bit representations like 8-bit or 4-bit, to reduce memory usage and improve computational efficiency.

    \item[Quantised LLMs] Large Language Models that have undergone quantisation, a process that reduces the precision of model weights and activations, often from 32-bit to 8-bit or lower, to enhance memory and computational efficiency.

    \item[Pruning] A model optimisation technique that reduces the complexity of large language models by removing less significant parameters, enabling faster inference and lower memory usage.

    \item[Half Fine-Tuning (HFT)] A fine-tuning method where half of the model's parameters are kept frozen while the other half are updated, helping to maintain pre-trained knowledge while adapting the model to new tasks.

    \item[Structured Masking] A technique that masks entire layers, heads, or other structural components of a model to reduce complexity while fine-tuning for specific tasks.

    \item[Unstructured Masking] A technique where certain parameters of the model are masked out randomly or based on a pattern during fine-tuning, allowing for the identification of the most important model weights.

    \item[GLUE] General Language Understanding Evaluation – A benchmark used to evaluate the performance of NLP models across a variety of language understanding tasks, such as sentiment analysis and natural language inference.

    \item[SuperGLUE] Super General Language Understanding Evaluation – A more challenging extension of GLUE, consisting of harder tasks designed to test the robustness and adaptability of NLP models.

    \item[TruthfulQA] A benchmark designed to measure the truthfulness of a language model's output, focusing on factual accuracy and resistance to hallucination.

    \item[IFEval] Instruction Following Evaluation – A benchmark that assesses a model’s ability to follow explicit instructions across tasks, usually in the context of fine-tuning large models for adherence to specific instructions.

    \item[BBH] Big Bench Hard – A subset of the Big Bench dataset, which consists of particularly difficult tasks aimed at evaluating the advanced reasoning abilities of large language models.

    \item[MATH] A dataset created to evaluate a model’s ability to solve high-school level mathematical problems, presented in formal formats like LaTeX.

    \item[GPQA] General-Purpose Question Answering – A challenging dataset that features knowledge-based questions crafted by experts to assess deep reasoning and factual recall.

    \item[MuSR] Multimodal Structured Reasoning – A dataset that involves complex problems requiring language models to integrate reasoning across modalities, often combining text with other forms of data such as images or graphs.

    \item[MMLU] Massive Multitask Language Understanding – A benchmark that evaluates a language model’s ability to perform various tasks across diverse domains, such as humanities, STEM, social sciences, and others, typically requiring high-level reasoning.

    \item[MMLU-PRO] A refined version of the MMLU dataset with a focus on more challenging, multi-choice problems, typically requiring the model to parse long-range context.

    \item[ARC] AI2 Reasoning Challenge – A benchmark for evaluating a language model’s reasoning capabilities using a dataset of multiple-choice science questions.

    \item[COQA] Conversational Question Answering – A benchmark that evaluates how well a language model can understand and engage in back-and-forth conversation, especially in a question-answer format.

    \item[DROP] Discrete Reasoning Over Paragraphs – A benchmark that tests a model's ability to perform discrete reasoning over text, especially in scenarios requiring arithmetic, comparison, or logical reasoning.

    \item[SQuAD] Stanford Question Answering Dataset – A popular dataset for evaluating a model's ability to understand and answer questions based on passages of text.

    \item[TREC] Text REtrieval Conference – A benchmark that evaluates models on various text retrieval tasks, often focusing on information retrieval and document search.

    \item[WMT] Workshop on Machine Translation – A dataset and benchmark for evaluating the performance of machine translation systems across different language pairs.

    \item[XNLI] Cross-lingual Natural Language Inference – A dataset designed to evaluate a model’s ability to understand and infer meaning across multiple languages.

    \item[PiQA] Physical Interaction Question Answering – A dataset that measures a model’s understanding of physical interactions and everyday tasks.

    \item[Winogrande] A large-scale dataset aimed at evaluating a language model’s ability to handle commonsense reasoning, typically through tasks that involve resolving ambiguous pronouns in sentences.

    \item[RLHF] Reinforcement Learning from Human Feedback – A method where language models are fine-tuned based on human-provided feedback, often used to guide models towards preferred behaviours or outputs.

    \item[RAFT] Retrieval-Augmented Fine-Tuning – A method combining retrieval techniques with fine-tuning to enhance the performance of language models by allowing them to access external information during training or inference.

\end{description}

\bibliographystyle{unsrt}
\bibliography{main} 

\end{document}